\documentclass[10pt,journal,compsoc]{IEEEtran}

\usepackage{cite}
\usepackage{times}
\usepackage{epsfig}
\usepackage{graphicx}
\usepackage{amsmath}
\usepackage{amssymb}
\usepackage{multirow}
\usepackage{amsthm}
\usepackage{subfigure}
\usepackage{color}
\usepackage{csquotes}
\usepackage{soul}
\usepackage{enumerate}
\usepackage[normalem]{ulem}
\usepackage{algorithm}
\usepackage{algorithmicx}
\usepackage{algpseudocode}
\usepackage{booktabs}
\usepackage{array}
\usepackage{rotating}
\usepackage{arydshln}
\usepackage{ragged2e}
\usepackage{wrapfig}
\usepackage{hyperref}
\hypersetup{breaklinks=true,colorlinks}

\newcommand\wrt{w.r.t.}

\def\Rev#1{\textcolor{black}{#1}}

\begin{document}
%
\title{Learning Generative Vision Transformer with Energy-Based Prior for Saliency Prediction}
\title{Generative Vision Transformer for Saliency Prediction with Energy-Based Prior}
\title{An Energy-Based Prior for Saliency Detection}
\title{An Energy-Based Prior for Generative Saliency}
%
%
%
%

       
\author{Jing~Zhang,
       Jianwen Xie,
       Nick Barnes and
       Ping Li
\IEEEcompsocitemizethanks{\IEEEcompsocthanksitem Jing Zhang and Nick Barnes are with School of Computing, the Australian National University. (Email: zjnwpu@gmail.com, nick.barnes@anu.edu.au)
\IEEEcompsocthanksitem Jianwen Xie is with Akool Research. (Email: jianwen@ucla.edu)
\IEEEcompsocthanksitem Ping Li is with LinkedIn Ads. (Email: pingli98@gmail.com)

\IEEEcompsocthanksitem A preliminary version of this work appeared at \cite{jing2021_nips}.

}
}

%
%

\markboth{Journal of \LaTeX\ Class Files,~Vol.~14, No.~8, August~2015}%
{Shell \MakeLowercase{\textit{et al.}}: Bare Demo of IEEEtran.cls for Computer Society Journals}
%




\IEEEtitleabstractindextext{%
\begin{abstract}
\justifying
We propose a novel generative saliency prediction framework that adopts an informative energy-based model as a prior distribution. The
energy-based prior model is defined on the latent space of a saliency generator network that generates the saliency map based on a continuous latent variables and an observed image. Both the parameters of saliency generator and the energy-based prior are jointly trained via Markov chain Monte Carlo-based maximum likelihood estimation, in which the sampling from the intractable posterior and prior distributions of the latent variables are performed by Langevin dynamics. With the generative saliency model, we can obtain a pixel-wise uncertainty map from an image, indicating model confidence in the saliency prediction. Different from existing generative models, which define the prior distribution of the latent variables as a simple isotropic Gaussian distribution, our model uses an energy-based informative prior which can be more expressive in capturing the latent space of the data. With the informative energy-based prior, we extend the Gaussian distribution assumption of generative models to achieve a more representative distribution of the latent space, leading to more reliable uncertainty estimation. We apply the proposed frameworks to both RGB and RGB-D salient object detection tasks with both transformer and convolutional neural network backbones. We further propose an adversarial learning algorithm and a variational inference algorithm as alternatives to train the proposed generative framework. Experimental results show that our generative saliency model with an energy-based prior can achieve not only accurate saliency predictions but also reliable uncertainty maps that are consistent with human perception. Results and code are available at \url{https://github.com/JingZhang617/EBMGSOD}.
\end{abstract}

\begin{IEEEkeywords}
Salient Object Detection, Generative Models, Energy-based Models, Vision Transformer, Langevin Dynamics, Variational Inference, Adversarial Learning.
\end{IEEEkeywords}
}

\maketitle

\IEEEdisplaynontitleabstractindextext
\IEEEpeerreviewmaketitle

\IEEEraisesectionheading{\section{Introduction}}
\label{intro_sec}
\IEEEPARstart{S}{alient}
object detection~\cite{wei2020f3net,wei2020label,fan2020bbs,Fu2020JLDCF,chen2018progressively,jing2020weakly} (SOD), also known as visual saliency prediction, aims
to highlight objects in images that attract more attention than their surrounding areas. With the emergence of deep convolutional neural networks, significant performance improvements have been achieved in the field of SOD \cite{scrn_sal,wei2020f3net,ucnet_sal,fan2020bbs}.
Traditional SOD models aim to learn a deterministic one-to-one mapping function from the image domain to the saliency domain using a set of training images along with their saliency annotations. These approaches often lead to point estimation techniques.
The one-to-one deterministic mapping mechanism employed in the current SOD frameworks makes them
unsuitable for exploring the pixel-wise uncertainty in predicting salient objects~\cite{kendall2017uncertainties}. Uncertainty measures the level of confidence the model has in its predictions. Without an uncertainty indicator, a one-to-one deterministic mapping model is prone to being over-confident. Further,
the saliency output of an image is subjective and heavily relies on human preference~\cite{ucnet_sal}. Therefore, compared with the deterministic model, a stochastic generative model is a more natural approach to model the stochastic nature of labeling by individuals and represent image saliency.


To account for the subjective nature of saliency, we propose a novel generative model for salient object detection. This model incorporates latent variables to capture randomness and uncertainty in the mapping from the image domain to the saliency domain. By doing so, it allows the model to generate stochastic saliency predictions, which can be useful for uncertainty estimation.
There are mainly two types of generative models that have been widely used, namely the variational auto-encoder (VAE)~\cite{vae_bayes_kumar} and the generative adversarial net (GAN)~\cite{GAN_nips},  which correspond to two different generative learning strategies to train latent variable models. To train a top-down latent variable generator, VAE~\cite{vae_bayes_kumar} introduces an extra encoder to approximate the intractable posterior distribution of the latent variables, and trains the generator via a perturbation of maximum likelihood. Based on the VAE, \cite{ucnet_sal} introduces a conditional variational autoencoder (CVAE)~\cite{structure_output} for RGB-D salient object detection. 
However, the potential issue of posterior collapse~\cite{Lagging_Inference_Networks}, which is inherent within the VAE framework, can make stochastic predictions less effective in generating reliable uncertainty estimates. In contrast, GAN introduces a discriminator to distinguish the generated examples from the real examples, and trains the generator to fool the discriminator without an inference step. However, due to the absence of an explicit inference process in GANs, it can be difficult to handle inverse problems and may lead to the potential issue of mode collapse.

\begin{figure*}[!htp]
   \begin{center}
   \begin{tabular}{c@{ } c@{ } c@{ } c@{ } c@{ } c@{ } c@{ }}
   {\includegraphics[width=0.135\linewidth]{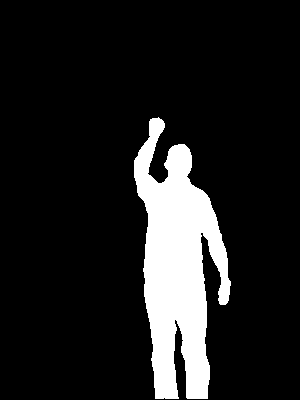}}&
   {\includegraphics[width=0.135\linewidth]{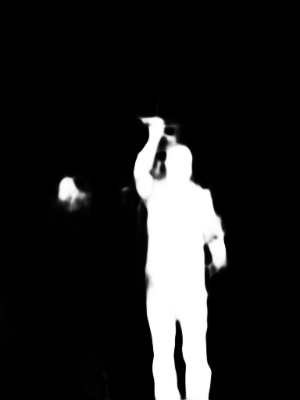}}&
   {\includegraphics[width=0.135\linewidth]{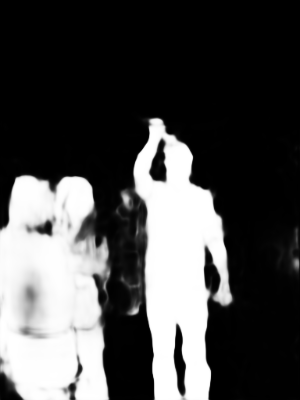}}&
   {\includegraphics[width=0.135\linewidth]{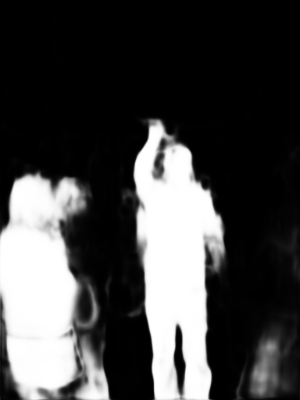}}&
   {\includegraphics[width=0.135\linewidth]{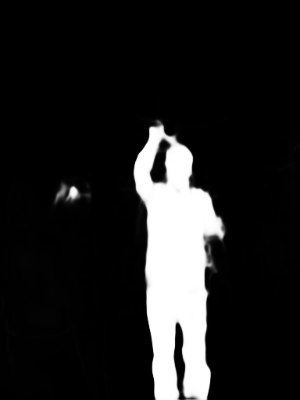}}&
   {\includegraphics[width=0.135\linewidth]{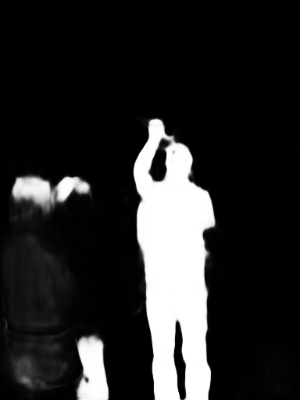}}&
   {\includegraphics[width=0.135\linewidth]{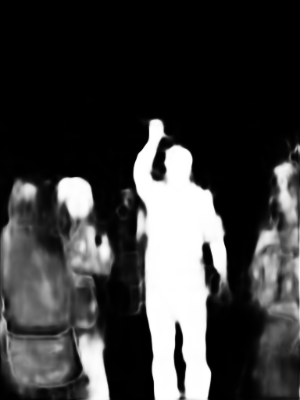}}\\
   {\includegraphics[width=0.135\linewidth]{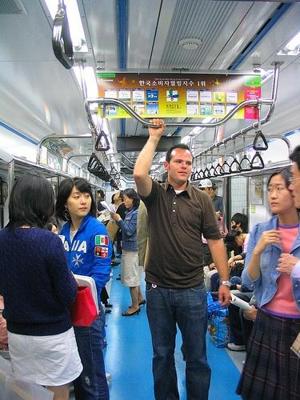}}&
   {\includegraphics[width=0.135\linewidth]{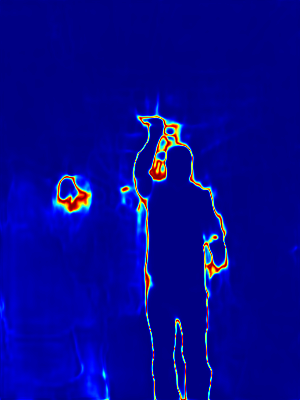}}&
   {\includegraphics[width=0.135\linewidth]{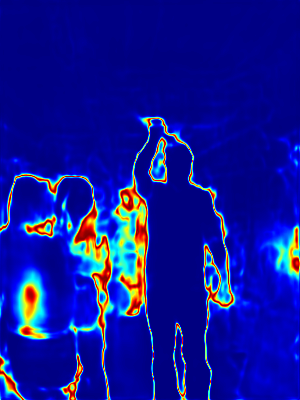}}&
   {\includegraphics[width=0.135\linewidth]{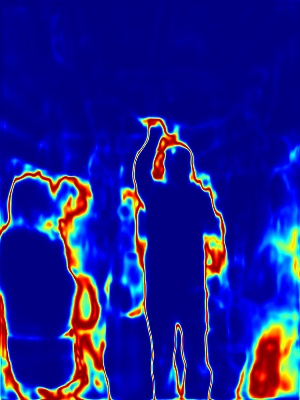}}&
   {\includegraphics[width=0.135\linewidth]{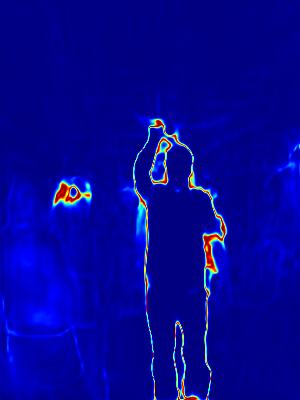}}&
   {\includegraphics[width=0.135\linewidth]{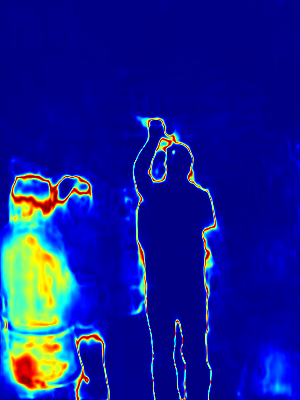}}&
   {\includegraphics[width=0.135\linewidth]{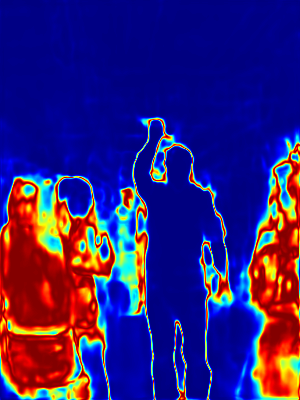}}\\
    \footnotesize{GT/Image}& \footnotesize{GAN \cite{gan_raw}} & \footnotesize{VAE \cite{vae_bayes_kumar}} & \footnotesize{ABP \cite{abp}} & \footnotesize{EGAN} & \footnotesize{EVAE} & \footnotesize{EABP} \\
   \end{tabular}
   \end{center}
   \caption{Predictions ($1^{st}$ row) and the corresponding uncertainty maps ($2^{nd}$ row) of the conventional isotropic Gaussian prior based generative models (GAN \cite{gan_raw}, VAE \cite{vae_bayes_kumar} and ABP \cite{abp}) and the proposed EBM prior based models (EGAN, EVAE and EABP).}
\label{fig:pred_visualization}
\end{figure*}

To address the potential issues like posterior collapse in VAEs and mode collapse in GANs,
\cite{abp,xie2019learning} present the third learning strategy called alternating back-propagation (ABP). This method trains the generator with an inference process, which directly samples latent variables from the true posterior distribution using gradient-based Markov chain Monte Carlo (MCMC)~\cite{liu2008monte} techniques such as Langevin dynamics~\cite{neal2011mcmc,WellingT11,DubeyRWPSX16}. It is worth noting that all the three generative models (i.e.,  VAE, GAN and ABP) assume that the latent variables follow a fixed and simple isotropic Gaussian distribution, resulting in an image-independent and uninformative latent space.



We aim to investigate generative SOD
with an expressive and meaningful latent representation.
To achieve this, \Rev{instead of defining the prior distribution of the latent variables as an isotropic Gaussian distribution as in the conventional generative models \cite{gan_raw,vae_bayes_kumar,abp}, we
assume that the latent variables follow an informative trainable energy-based prior distribution~\cite{ebm_prior,PangW21},
which is parameterized
by a deep neural network. Such informative latent space modeling strategy with an energy-based model (EBM) as a prior can be plugged into existing latent variable models to achieve reliable latent space modeling, leading to EBM based GAN~\cite{GAN_nips} (EGAN), EBM based VAE~\cite{vae_bayes_kumar} (EVAE) and EBM based ABP~\cite{abp} (EABP). For salient object detection, the informative latent space is the necessity for reliable saliency subjective nature modeling.}

For EGAN, we introduce an extra inference model,
where the prior of the latent variable is modeled with the EBM, and the posterior is updated via Langevin dynamics based MCMC \cite{neal2011mcmc}. For EVAE, the prior and posterior of the latent variable within the VAE framework \cite{vae_bayes_kumar} serve as the start point of our EBM prior and posterior models respectively, leading to an image-dependent warm-start of both models. The prior and posterior of EABP is the same as for EGAN. Differently, there exists extra discriminator in EGAN to achieve adversarial learning. Note that the generator of the three models can be
trained by maximum likelihood estimation (MLE).
The MLE algorithm relies on MCMC sampling to evaluate the intractable prior and posterior distributions of the latent variables. In this way, we do not
suffer the
posterior collapse issue as does
the original VAE model \cite{vae_bayes_kumar}, and the latent variable is more meaningful which is not an
isotropic Gaussian but modeled with a deep neural network, that can be any distribution.
We apply the proposed generative models to RGB and RGB-D salient object detection, and
show that the generative frameworks with the EBM prior
are powerful in representing the conditional distribution of object saliency given an image, leading to more reasonable uncertainty estimation~\cite{kendall2017uncertainties} compared with the isotropic Gaussian distribution based counterparts (see Figure~\ref{fig:pred_visualization}).

This paper is an extended version of our conference paper~\cite{jing2021_nips}. In particular, we extend~\cite{jing2021_nips} in the following ways: (i)
after we firstly
introduce \enquote{EABP} from
our original version, we then extend our EBM prior to GAN \cite{gan_raw} and VAE \cite{vae_bayes_kumar}, demonstrating that the proposed approach can be effective for a broad range of generative techniques;
(ii) we implement the three types of EBM prior based generative models with both transformer (which is our conference version in \cite{jing2021_nips}) and convolutional neural network (CNN) backbones (which is our extension)
to verify the superiority of the proposed EBM prior based generative models for uncertainty estimation; (iii) we also apply the proposed EBM prior based generative models to existing saliency detection models \cite{Liu_2021_ICCV_VST} to explain the generalization ability of the proposed strategies; (iv) we extensively analyse the transformer backbone and CNN backbones \wrt~both deterministic performance and uncertainty modeling, observing that although significant deterministic performance gaps exists for the two types of backbones, our EBM prior based generative models within both scenarios can
generate reliable uncertainty maps to explain the corresponding predictions.
\section{Related Work}

\textbf{Salient Object Detection:}
The main goal of the existing deep fully-supervised RGB image-based salient object detection models~\cite{cpd_sal,nldf_sal,scrn_sal,wei2020f3net,wang2020progressive,zhang2020learning_eccv,wei2020label,aixuan_cod_sod21,distributional_uncertainty_cvpr2023} is to achieve structure preserving saliency prediction, by either 
sophisticated feature aggregation~\cite{cpd_sal,wang2020progressive}, auxiliary edge detection~\cite{scrn_sal,qin2019basnet,wei2020label}, or resorting to structure-aware loss functions~\cite{nldf_sal,wei2020f3net}.
\cite{scrn_sal} introduces an extra edge module and uses the interaction between the edge module and the detection module to optimize the two tasks at the same time. \cite{wei2020f3net} adaptively selects complementary features while performing multi-scale feature aggregation with a new
structure-aware loss function for edge-aware saliency prediction.
Similarly,~\cite{qin2019basnet} uses multiple supervisions and high-quality boundaries to guide the network to gradually optimize saliency prediction.
~\cite{Pang_2020_CVPR} integrates the information of adjacent layers, as well as
multi-scale information to retain the internal consistency of each category, i.e., salient foreground and non-salient background.
With extra depth information, RGB-D salient object detection models \cite{qu2017rgbd,ucnet_sal,chen2018progressively,ji2021cal,zhao2019Contrast,ssf_rgbd,fan2020bbs,ji2020accurate,piao2019depth,zhang2020bilateral,Sun_2021_CVPR_DSA2F,Zhang_2021_ICCV_RGBD} mainly focus on effective multi-modal modeling, where both the appearance information from the RGB image and geometric information from depth are fused at various levels, for example, input level (early fusion models), output level (late fusion models), and feature level (cross-fusion models). 
Our paper solves the same problems, i.e., RGB and RGB-D salient object detection, by developing a new generative transformer-based saliency detection framework.  

\textbf{Vision Transformers:}
Transformer~\cite{transformer_nips} is a set-to-set method based on a self-attention mechanism, which 
has achieved great success in natural language processing (NLP), and
the breakthroughs
in the
NLP
domain have sparked the interest of the computer vision community in developing vision transformers for different computer vision tasks, such as image classification \cite{dosovitskiy_ViT_ICLR_2021,liu2021swin}, object detection~\cite{carion_DETR_ECCV_2020, wang_PVT_2021,mvt_multi_view_transformer,liu2021swin}, image segmentation~\cite{zheng_SETR_2020, dpt_transformer, wang_PVT_2021, liu2021swin}, object tracking~\cite{xu_TransCenterTracking_2021, yan_SpatialTemporalTransformerTrackingv2_2021}, pose estimation~\cite{mao_TFPose_2021,stoffl_InstancePose_2021}, etc.
Among vision transformers, DPT~\cite{dpt_transformer} adopts a U-shaped structure and uses ViT~\cite{dosovitskiy_ViT_ICLR_2021} as an encoder to perform semantic segmentation and monocular depth estimation.
Swin~\cite{liu2021swin} presents a hierarchical transformer with a shifted windowing scheme to achieve an efficient transformer network with high resolution images as input. Different from the above vision transformers that mainly focus on discriminative modeling and learning, our paper emphasizes generative modeling  and learning of the vision transformer by involving latent variables and MCMC inference.

\textbf{Dense Prediction with Generative Models:}
The goal of dense generative models is to produce both
deterministic
accurate model prediction and a
meaningful stochastic
uncertainty map representing model awareness of its prediction.
Two types of generative models have been widely studied, namely VAEs~\cite{structure_output,vae_bayes_kumar} and GANs~\cite{GAN_nips}. VAEs use an
extra encoder to constrain the distribution of the latent variable, and GANs design a discriminator to distinguish real samples and the generated samples.
VAEs have been successfully applied to image segmentation \cite{PHiSeg2019, probabilistic_unet}. For saliency prediction,~\cite{SuperVAE_AAAI19} adopts a VAE to model the image background, and separates salient objects from the background through reconstruction residuals.~\cite{ucnet_sal,jing2020uncertainty} design CVAEs to model the subjective nature of saliency. GAN-based methods can be classified into two categories, depending on whether they are designed for fully-supervised or semi-supervised settings. The fully-supervised models~\cite{groenendijk2020benefit,gan_maskerrcnn} use a discriminator to distinguish model predictions from the ground truth, while the semi-supervised models~\cite{gan_semi_seg,hung2018adversarial} use the GAN to explore the information contained in
unlabeled data. \cite{zhang2021energy} uses a cooperative learning framework \cite{XieLGW18,xie2018cooperative,xie2021cooperative} for generative saliency prediction. ~\cite{jing2020uncertainty} trains a single top-down generator in the ABP framework for RGB-D saliency prediction. Our model generalizes \cite{jing2020uncertainty} by replacing the simple Gaussian prior by a learnable EBM prior and adopting a vision transformer generator for SOD. 

\textbf{Energy-based Models:} Recent works have shown strong performance of data space EBMs \cite{xie2016theory,nijkamp2019learning} in modeling high-dimensional complex dependencies, such as images \cite{XieHZW15,xie2016inducing,ZhengXL21,ZhaoXL21,GaoLZZW18,DuM19,GaoSPWK21, XieZL21, XieZLL22}, videos \cite{XieZW17,XieZW21}, 3D shapes \cite{XieZGWZW18,xie2020generative}, unordered point clouds \cite{XieXZZW21} and trajectories \cite{xu2022energy}, and also the effectiveness of latent space EBMs \cite{ebm_prior} in improving model expressivity for text \cite{PangW21}, image \cite{ebm_prior,aistats_lebm}, and trajectory \cite{PangZ0W21} generation. Our paper also learns a latent space EBM as the prior model but builds the EBM on top of a transformer generator for image-conditioned saliency map prediction. This leads to three EBM prior based generative saliency detection models to relax the \enquote{Gaussian assumption} of conventional latent variable models \cite{gan_raw,vae_bayes_kumar,abp,xie2019learning}.


\section{Generative Saliency Detection}

\subsection{The Formulation}

Generative saliency detection can be formulated as a conditional generative learning problem. We begin by introducing some notation. Let $x \in \mathbb{R}^{h \times w \times 3}$ be an observed RGB image, $y \in \mathbb{R}^{h \times w \times 1}$ be the corresponding ground truth saliency map, and $z \in \mathbb{R}^{1 \times 1 \times d}$ be the $d$-dimensional vector of latent variables, where $h \times w  \gg d$. In this context, we consider the following generative model to predict a saliency map $y$ from an image $x$:
\begin{equation}
     y=T_{\theta}(x, z)+ \epsilon, \hspace{3mm} z \sim p_\alpha(z), \hspace{3mm} \epsilon \sim \mathcal{N}(0,\sigma_\epsilon^2 I),   \label{eq:generative_sod_model}
\end{equation}
where $T_\theta$ is the non-linear mapping process from $[z,x]$ to $y$ with parameters $\theta$, $p_\alpha(z)$ is the prior distribution with parameters $\alpha$, and $\epsilon \sim \mathcal{N}(0,\sigma_\epsilon^2 I)$ is the observation residual of saliency with $\sigma_\epsilon$ being given. Because of the stochasticity of the latent variables $z$, the saliency map for a given image $x$ is also probabilistic. This probabilistic model aligns with the subjective nature of image saliency \cite{jing2020uncertainty}. To create a generative saliency detection model, one can follow conventional generative models such as GAN \cite{gan_raw}, VAE \cite{vae_bayes_kumar}, and ABP \cite{abp} by defining the prior distribution $p_\alpha(z)$ of the latent variables $z$ as an simple isotropic Gaussian distribution, resulting in $p(z)=\mathcal{N}(0,I)$, where we drop $\alpha$ because the prior distribution does not have trainable parameters. The above three generative models differ in the inference process during training. Specifically, VAEs \cite{vae_bayes_kumar} use an additional deep neural network to approximate the posterior distribution of $z$ for inference, while ABPs \cite{abp} directly sample from the true posterior distribution of $z$ using gradient-based Markov Chain Monte Carlo (MCMC) \cite{neal2011mcmc}. In contrast, GANs \cite{gan_raw} do not have an explicit inference model but an extra classifier for adversarial learning.


\subsection{Generative Saliency v.s. Deterministic Saliency}
Different from a stochastic model, which produces different outputs given the same input due to random factors, a deterministic model provides a direct mapping between inputs and the outputs. A deterministic saliency detection model directly achieves a mapping function from input image space to the output saliency space, given by the following regression model
\begin{equation}
     y=T_{\theta}(x)+ \epsilon, \hspace{3mm} \epsilon \sim \mathcal{N}(0,\sigma_\epsilon^2 I). \label{eq:deterministic_sod_model}
\end{equation}
Similarly, $T_{\theta}$ is the mapping function parameterized by a neural net, and $\epsilon$ represents the observation residual.
Compared with the generative saliency formulation in Eq. (\ref{eq:generative_sod_model}), the absence of latent variables $z$ in the deterministic model shown in Eq. (\ref{eq:deterministic_sod_model}) does not require dealing with the challenging inference of latent variables, and the estimation of parameters $\theta$ can be easily achieved via supervised learning. \Rev{The deterministic model corresponds to a point estimation framework, where a fixed prediction is associated with each input image, making it impossible to explore the subjective nature of human saliency.} 
We argue that the inclusion of the extra latent variables $z$ within the generative saliency detection formulation in Eq. (\ref{eq:generative_sod_model}) leads to a stochastic prediction framework. This makes it natural and effective to represent the distribution of saliency responses over different individuals given an image.


\subsection{Energy-based Prior v.s. Gaussian Prior}

\begin{figure}[t]
\centering
\includegraphics[width=0.95\linewidth]{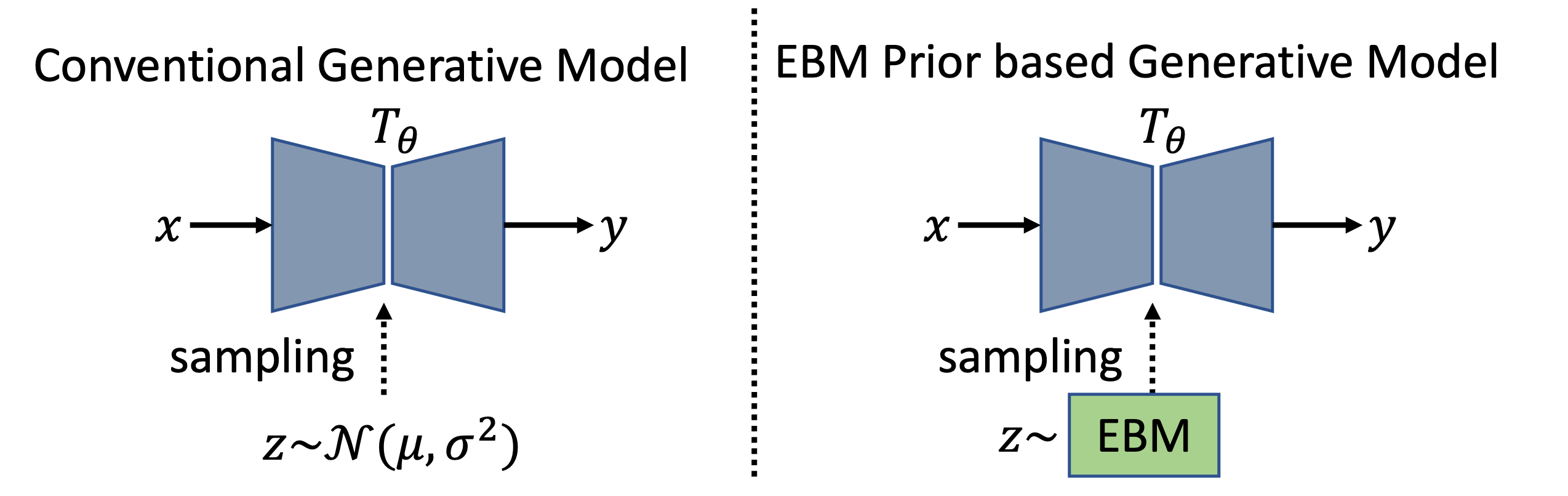}
\caption{An illustrative comparison between the conventional Gaussian prior-based generative models and the proposed EBM prior-based generative models. The key advantage of the EBM prior is its flexibility in representing arbitrary distributions, but sampling from an EBM prior relies on MCMC. ($\mu$ and $\sigma$ are the mean and standard deviation of the Gaussian prior distribution of the latent variables $z$, respectively. $T_{\theta}$ is a mapping function that takes an image $x$ as input and produces a saliency map $s$ as output.)
}
\label{fig:conventional_vs_ebm_prior}
\end{figure}

Although the generative model pipeline is beneficial for uncertainty estimation and representing the subjective nature of saliency, we believe that the effectiveness of the latent space is the core of the generative model. A less informative latent space, such as a simple Gaussian distribution, may fail to accurately reflect the true uncertainty of the saliency, resulting in a less reliable salient object prediction process. Secondly, an informative prior model, such as an energy-based model, in the latent space can further improve
the expressive power of the whole model. Thirdly, the energy function in the EBM prior can be viewed as an objective
function or a cost function, which captures regularities and rules in the latent space of the data. Finally, a trainable energy-based prior can be more advantageous than a fixed Gaussian prior as it has the capability to correct the mismatch between the prior and the aggregated posterior resulting from a biased inference process, such as an approximate inference network in VAE or a non-convergent short-run MCMC inference \cite{AnXL21} in ABP.
In this paper, we depart from the conventional generative approach of assuming a Gaussian distribution for the latent variable and propose the use of an EBM as a prior. The EBM is parameterized with a deep neural network to explore a more informative latent space, resulting in a more robust and reasonable generative saliency prediction framework. Refer to Figure~\ref{fig:conventional_vs_ebm_prior} for a comparison of diagrams between the Gaussian prior and the EBM prior, and Figure~\ref{fig:pred_visualization} for an qualitative comparison of the prediction results produced by different models.

\section{Learning An Energy-based Prior for Generative Saliency Detection}

\subsection{Model}
\label{subsec:model}


Given the generative saliency model in Eq.~\ref{eq:generative_sod_model}, following~\cite{ebm_prior}, the prior $p_{\alpha}(z)$ is not assumed to be a simple isotropic Gaussian distribution as it is GANs~\cite{GAN_nips}, VAEs~\cite{vae_bayes_kumar, structure_output} or ABP~\cite{abp}. 
Specifically, it is in the form of the energy-based correction or exponential tilting~\cite{xie2016theory} of an isotropic Gaussian reference distribution $p_0(z)=\mathcal{N}(0,\sigma_z^2I)$, i.e.:
\begin{equation}
\label{eq:ebm_prior}
\begin{aligned}
     p_\alpha(z)&= \frac{1}{Z(\alpha)} \exp \left[-U_{\alpha}(z) \right] p_{0}(z)\\
     &\propto \exp \left[-U_{\alpha}(z)-\frac{1}{2\sigma_z^2}||z||^2 \right],
\end{aligned}
\end{equation}
where $\mathcal{E}_{\alpha}(z)=U_{\alpha}(z)+\frac{1}{2\sigma_{z}^2}||z||^2$ is the energy function that maps latent variables $z$ to a scalar, and $U_{\alpha}(z)$ is parameterized by a multi-layer perceptron (MLP) with trainable parameters $\alpha$. The standard deviation $\sigma_z$ is a hyperparameter. $Z(\alpha)=\int \exp[-U_{\alpha}(z)]p_0(z)dz$ is the intractable normalizing constant that resolves the requirement for a probability distribution to have a total probability equal to one. $p_{\alpha}(z)$ is an informative prior distribution in our model and its parameters $\alpha$ need to be estimated along with the non-linear mapping function $T_{\theta}$ from training data.

The mapping function $T_{\theta}$
encodes the input image $x$ and then decodes it along with the vector of latent variables $z$ to the saliency map $y$, thus, $p_{\theta}(y|x,z)=\mathcal{N}(T_{\theta}(x,z),\sigma_{\epsilon}^2 I)$. The resulting generative model is a conditional directed graphical model that combines the EBM prior~\cite{ebm_prior} and the saliency generator $T_{\theta}$.

\subsection{Maximum Likelihood Learning}
\label{subsec:learning}

The generative saliency prediction model with an energy-based prior, which is presented in Eq.~(\ref{eq:generative_sod_model}), can be trained via maximum likelihood estimation. For notation simplicity, let $\beta=(\theta, \alpha)$. 
For the training examples $\{(x_i, y_i),i=1,...,n\}$, the observed-data log-likelihood function is defined as:
\begin{equation}
\label{eq:mle}
\begin{aligned}
     L(\beta)&=\sum_{i=1}^{n} \log p_{\beta}(y_i|x_i)=\sum_{i=1}^{n} \log \left[\int p_{\beta}(y_i,z_i|x_i)dz_i \right]\\
     &=\sum_{i=1}^{n} \log \left[\int p_{\alpha}(z_i)p_{\theta}(y_i|x_i,z_i)dz_i \right]. \nonumber %
\end{aligned}
\end{equation}
Maximizing $L(\beta)$ is equivalent to minimizing the Kullback-Leibler (KL) divergence between the model $p_{\beta}(y|x)$ and the data distribution $p_{\text{data}}(y|x)$. The gradient of $L(\beta)$ can be computed based on:
\begin{equation}
\label{eq:gradient}
\begin{aligned}
    \nabla_{\beta} \log p_{\beta}(y|x)&= \mathbb{E}_{p_{\beta}(z|y,x)} \left[\nabla_{\beta} \log p_{\beta}(y,z|x) \right]\\
    &=\mathbb{E}_{p_{\beta}(z|y,x)}[\nabla_{\beta} (\log p_{\alpha}(z)+  \log p_{\theta}(y|x,z))],
\end{aligned}
\end{equation}
where the posterior distribution is:
\begin{equation}
    \label{posterior_z}
    p_{\beta}(z|y,x)=p_{\beta}(y,z|x)/p_{\beta}(y|x)=p_{\alpha}(z)p_{\theta}(y|x,z)/p_{\beta}(y|x).
\end{equation}

The learning gradient in Eq.~(\ref{eq:gradient}) can be decomposed into two parts, i.e., the gradient for the energy-based model $\alpha$:
\begin{equation}
 \label{eq:prior_gradient}
\begin{aligned}
    &\mathbb{E}_{p_{\beta}(z|y,x)}[\nabla_{\alpha} \log p_{\alpha}(z)]\\
    &= \mathbb{E}_{p_{\alpha}(z)}[\nabla_{\alpha} U_{\alpha}(z)]-\mathbb{E}_{p_{\beta}(z|y,x)}[\nabla_{\alpha} U_{\alpha}(z)],
\end{aligned}
\end{equation}
and the gradient for the saliency generator $\theta$:
\begin{equation}
\label{eq:transformer_gradient}
\begin{aligned}
    &\mathbb{E}_{p_{\beta}(z|y,x)}[\nabla_{\theta} \log p_{\theta}(y|x,z)]\\
    &=\mathbb{E}_{p_{\beta}(z|y,x)}\left[\frac{1}{\sigma_{\epsilon}^2}(y-T_{\theta}(x,z)) \nabla_{\theta} T_{\theta}(x,z)\right].
\end{aligned}
\end{equation}
$\nabla_{\alpha} U_{\alpha}(z)$ in Eq.~(\ref{eq:prior_gradient}) and
$\nabla_{\theta} T_{\theta}(x,z)$ in Eq.~(\ref{eq:transformer_gradient}) can be efficiently computed via back-propagation. Both Eq.~(\ref{eq:prior_gradient}) and Eq.~(\ref{eq:transformer_gradient}) include intractable expectation terms $\mathbb{E}_{p}(\cdot)$, which can be approximated by MCMC samples. To be specific, we can use a gradient-based MCMC, e.g., Langevin dynamics, which is initialized with a Gaussian noise distribution $p_0$, to draw samples from the energy-based prior model $p_{\alpha}(z) \propto \exp \left[-\mathcal{E}_{\alpha}(z)\right]$ by iterating:
\begin{equation}
    z_{t+1} = z_{t} - \delta \nabla_z \mathcal{E}_{\alpha}(z_t) + \sqrt{2\delta} e_t, \hspace{2mm} z_0\sim p_{0}(z), \hspace{2mm} e_t \sim \mathcal{N}(0,I),
    \label{eq:MCMC_prior}
\end{equation}
and drawing samples from the posterior distribution $p_{\beta}(z|y, x)$ by iterating:
\begin{equation}
\label{eq:MCMC_post}
\begin{aligned}
     z_{t+1} &= z_{t} - \delta  \left[\nabla_{z} \mathcal{E}_{\alpha}(z_t)-\frac{1}{\sigma_{\epsilon}^2}(y-T_{\theta}(x,z_t)) \nabla_{z} T_{\theta}(x,z_t) \right] \\
     &+ \sqrt{2\delta} e_t, \hspace{2mm} z_0\sim p_{0}(z), \hspace{2mm} e_t \sim \mathcal{N}(0,I).
\end{aligned}
\end{equation}
$\delta$ is the Langevin step size and can be specified independently in Eq.~(\ref{eq:MCMC_prior}) and Eq.~(\ref{eq:MCMC_post}). We use $\{z_i^{+}\}$ and $\{z_i^{-}\}$ to denote, respectively, the samples from the posterior distribution $p_{\beta}(z|y,x)$ and the prior distribution $p_{\alpha}(z)$. The gradients of $\alpha$ and $\theta$ can be computed with $\{(x_i, y_i)\}$, $\{z_i^{+}\}$ and $\{z_i^{-}\}$ by:
\begin{align}
    \nabla{\alpha}&= \frac{1}{n}\sum_{i=1}^{n}[\nabla_{\alpha} U_{\alpha}(z_i^{-})]- \frac{1}{n}\sum_{i=1}^{n} \left[\nabla_{\alpha} U_{\alpha}(z_{i}^{+})\right],  \label{eq:prior_gradient_MCMC}
    \\
    \nabla{\theta}&= \frac{1}{n}\sum_{i=1}^{n}\left[\frac{1}{\sigma_{\epsilon}^2}(y_i-T_{\theta}(x_i,z_i^{+})) \nabla_{\theta} T_{\theta}(x_i,z_i^{+})\right] .\label{eq:posterior_gradient_MCMC} 
    \end{align}
We can then update the parameters with $\nabla{\alpha}$ and $\nabla{\theta}$ via the Adam optimizer~\cite{kingma2014adam}.
We present the full learning and sampling algorithm of our model in Algorithm~\ref{alg_eabp_train}, and the pipeline of saliency detection by a learned model in Algorithm~\ref{alg_eabp_test}.

\begin{algorithm}[H]
\small
\caption{Maximum likelihood learning of a generative saliency model with an energy-based prior model
}
\textbf{Input}: (1) Training images $\{x_i\}_{i}^{n}$ with associated saliency maps $\{y_i\}_{i}^{n}$;
(2) Number of learning iterations $M$; (3) Numbers of Langevin steps for prior and posterior $\{K^{-},K^{+}\}$; (4) Langevin step sizes for prior and posterior $\{\delta^{-},\delta^{+}\}$; (5) Learning rates for energy-based prior and saliency generator $\{\xi_\alpha,\xi_\theta\}$; (6) batch size $n'$.\\
\textbf{Output}: 
Parameters $\theta$ for the saliency generator
and $\alpha$ for the energy-based prior model
\begin{algorithmic}[1]
\State Initialize $\theta$ and $\alpha$ 
\For{$m \leftarrow  1$ to $M$}
\State Sample observed image-saliency pairs $\{(x_i,y_i)\}_i^{n'}$
\State For each $(x_i,y_i)$, sample the prior $z_i^{-} \sim p_{\alpha_m}(z)$ using $K^{-}$ Langevin steps in Eq.~(\ref{eq:MCMC_prior}) with a step size $\delta^{-}$. 
\State For each $(x_i,y_i)$, sample the posterior $z_i^{+} \sim p_{\beta_m}(z|y_i,x_i)$ using $K^{+}$ Langevin steps in  Eq.~(\ref{eq:MCMC_post}) with a step size $\delta^{+}$. 
\State Update energy-based prior using Adam with the gradient $\nabla{\alpha}$ computed in Eq.~(\ref{eq:prior_gradient_MCMC}) and a learning rate $\xi_\alpha$.
\State Update the saliency generator using Adam with the gradient $\nabla{\theta}$ computed in Eq.~(\ref{eq:posterior_gradient_MCMC}) and a learning rate $\xi_\theta$.
\EndFor
\end{algorithmic} \label{alg_eabp_train}
\end{algorithm}

\begin{algorithm}[H]
\small
\caption{Saliency detection by a generative saliency model with an energy-based prior model
}
\textbf{Input}: (1) A testing image $x^*$; (2) Parameters $\theta$ for the saliency generator and $\alpha$ for the energy-based prior model; (3) Number of Langevin steps for the prior model $K^{-}$; (4) Langevin step sizes for the prior $\delta^{-}$; (5) Number of trials $J$ for prediction.

\textbf{Output}: 
Saliency prediction $y^*$ for $x^*$ and the corresponding uncertainty map $u^*$
\begin{algorithmic}[1]
\For{$j \leftarrow  1$ to $J$}
\State Sample latent variables from the prior model $z^{-}_j \sim p_{\alpha}(z)$ using $K^{-}$ Langevin steps in Eq.~(\ref{eq:MCMC_prior}) with a step size $\delta^{-}$. 
\State Compute saliency prediction $y_j=T_\theta(x^*,z^{-}_j)$. 
\EndFor
\State Obtain final saliency prediction via  $y^*=\frac{1}{J}\sum_{j=1}^{J}y^*_j$.
\State Obtain uncertainty as variance of $J$ trials of predictions: 
$u^*=\frac{1}{J}\sum^{J}_{j=1}(y_j - y^*)^2$.
\end{algorithmic} \label{alg_eabp_test}
\end{algorithm}

\subsection{Analysis}

\subsubsection{Convergence} Theoretically, when the Adam optimizer of $\beta=(\theta,\alpha)$ in the learning algorithm converges to a local minimum, it solves the estimating equations $\nabla{\alpha}=0$ and $\nabla{\theta}=0$, which are:  
\begin{equation}
\label{eq:eq1} 
    \begin{aligned}
    \frac{1}{n}\sum_{i=1}^{n}\text{E}_{p_{\beta}(z_i|y_i,x_i)}[\nabla_{\alpha} U_{\alpha}(z_i)]-\text{E}_{p_{\alpha}(z)}[\nabla_{\alpha} U_{\alpha}(z)]=0,
    \end{aligned}
\end{equation}
\begin{equation}
\label{eq:eq2} 
    \begin{aligned}
    \frac{1}{n}\sum_{i=1}^{n}\text{E}_{p_{\beta}(z_i|y_i,x_i)}\left[\frac{1}{\sigma_{\epsilon}^2}(y_i-T_{\theta}(x_i,z_i)) \nabla_{\theta} T_{\theta}(x_i,z_i)\right]=0.
    \end{aligned}
\end{equation}
The above two equations are the maximum likelihood estimating equations. However, in practice, the Langevin dynamics shown in Eq.~(\ref{eq:prior_gradient_MCMC}) and Eq.~(\ref{eq:posterior_gradient_MCMC}) might not converge to the target distributions due to the use of a small number of Langevin steps (i.e., short-run non-convergent MCMC).  As a result, the estimating equations in Eq.~(\ref{eq:eq2}) and Eq.~(\ref{eq:eq1}) will correspond to a perturbation of the MLE estimating equation according to~\cite{nijkamp2019learning,nijkamp2020learning,ebm_prior}. The learning algorithm can be justified as a Robbins-Monro~\cite{robbins1951stochastic} algorithm, whose convergence is theoretically sound.

\subsubsection{Accuracy} Compared with the original GAN-based generative framework \cite{gan_raw}, our model is a likelihood-based generative framework that will not suffer from mode collapse~\cite{AroraRZ18}. In comparison with the conventional  VAE-based generative framework \cite{vae_bayes_kumar,structure_output}, whose training is also based on likelihood, our MCMC-based maximum likelihood learning algorithm will not encounter the posterior collapse issue that is caused by amortized inference. On the other hand, the variational inference typically relies on an extra inference network for efficient inference of the latent variables given an image and saliency pair, however, the approximate inference might not be able to take the full place of the posterior distribution in practice. To be specific, we use $q_{\theta}(z|x,y)$ to denote the tractable approximate inference network with parameters $\theta$. The variational inference seeks to optimize
$\min_{\beta} \min_{\theta} \text{KL}(p_{\text{data}}(y|x)q_{\theta}(z|x,y)||p_{\beta}(z,y|x))$, which can be further decomposed into $\min_{\beta} \min_{\theta} \text{KL}(p_{\text{data}}(y|x)|| p_{\beta}(y|x)) + \text{KL}( q_{\theta}(z|x,y)  || p_{\beta}(z|x,y))$. That is, the variational inference maximizes the conditional data likelihood plus a KL-divergence between the approximate inference network and the posterior distribution. Only when the $\text{KL}( q_{\theta}(z|x,y) || p_{\beta}(z|x,y)) \rightarrow 0$, the variational inference will lead to the MLE solution, which is exactly the objective of our model. However, there might exist a gap between them in practice due to the improper design of the inference network. Our learning algorithm is much simpler and more accurate than amortized inference. 

\subsubsection{Computational and Memory Costs Analysis}
From the learning perspective, due to the iterative Langevin sampling for the posterior and prior distributions of latent variables, our EBM prior based generative models are more time-consuming for training.
For example, the conventional generative models with amortized inference, such as VAE \cite{vae_bayes_kumar,structure_output}, is
roughly 1.2 times faster than ours on RGB salient object detection. However, for testing, the extra EBM prior Langevin does not lead to significantly longer inference time, which is comparable with the conventional generative models.

\subsection{Other Variants}
We have provided the energy-based prior model in Section \ref{subsec:model}, and introduce the principled maximum likelihood learning strategy in Section \ref{subsec:learning}, which can be called EABP, where we infer the latent variables $z$ by Langevin sampling from its true posterior distribution $p_\beta(z|y,x)$. In this section, we will utilize adversarial learning and variational inference techniques to develop two variants of our generative saliency model: the EGAN, which is a GAN-based model, and the EVAE, which is an VAE-based model. Both use an energy-based model as a prior distribution.

\subsubsection{An EBM Prior for GAN-based Generative Saliency}
We can train the generative saliency prediction model with an energy-based prior through adversarial learning. This can be achieved
by introducing an extra fully convolutional conditional discriminator \cite{hung2018adversarial}, denoted as $D_\gamma$, where $\gamma$ refers to the parameters of the discriminator.
Following the conventional conditional adversarial learning pipeline, the objective of our framework can be expressed as:
\begin{equation}
\label{gan_loss}
\begin{aligned}
    L_{\text{adv}} &(\gamma,\theta) = \mathbb{E}_{(x,y)\sim p_{\text{data}}(x,y)}[\log D_\gamma(y, x)]\\ &+ \mathbb{E}_{x\sim p_{\text{data}}(x),z\sim p_\alpha(z)}[\log(1-D_\gamma(T_\theta(x,z),x))],    
\end{aligned}
\end{equation}
where
$p_{\text{data}}(x,y)$ is the joint data distribution, $p_{\text{data}}(x)$ is the marginal data distribution, and $p_\alpha(z)$ is the prior distribution of the latent variables $z$, which is usually defined as a known Gaussian distribution in the conventional GAN framework but now is a trainable energy-based distribution in our model. 

Previous approaches have found it beneficial to combine the GAN objective with a more traditional reconstruction loss, such as L2 \cite{PathakKDDE16} or L1 \cite{IsolaZZE17} distance. While the discriminator's role remains unchanged, the generator aims to not only fool the discriminator but also produce outputs that are closer to the ground truth. In our case, where we are predicting saliency, the reconstruction loss can be calculated using a binary cross-entropy loss or a structure-aware loss \cite{wei2020f3net} between the generated outputs and the ground truth outputs, that is: 
\begin{equation}
\label{recon_loss}
\begin{aligned}
     L_{\text{rec}} (\theta) = \mathbb{E}_{(x,y)\sim p_{\text{data}}(x,y),z\sim p_{\alpha}(z)}[ \ell(T_{\theta}(x,z),y)],    
\end{aligned}
\end{equation}
where $\ell(y',y)$ calculates the cross-entropy between $y'$ and $y$. Our final objective to learn the saliency generator $\theta$ and the discriminator $\gamma$ is given by:
\begin{equation}
\label{gan_loss2}
\begin{aligned}
 \min_{\theta} \max_{\gamma}  L_{\text{rec}}(\theta) + \lambda L_{\text{adv}}(\gamma, \theta) , 
\end{aligned}
\end{equation}
where the hyper-parameter $\lambda$ is empirically set to be $0.1$ for stable training. Both Eq. (\ref{gan_loss}) and Eq. (\ref{recon_loss}) require sampling from the prior distribution to generate saliency outputs. We follow Langevin dynamics in Eq. (\ref{eq:MCMC_prior}) to draw samples from $p_{\alpha}(z)$. The update of the parameters $\alpha$ of the energy-based prior model follows Eq.~(\ref{eq:prior_gradient_MCMC}), in which we need to sample from the EBM prior and the posterior distribution $p_{\beta}(z|y,x)$.


We highlight a significant issue with conventional conditional GANs, namely that the distribution of the latent space is independent of training data. This makes it less effective in representing the latent space of the data distribution. Instead of assuming the prior to be an isotropic Gaussian distribution, we adopt an EBM prior model in EGAN and train it through empirical Bayes. Additionally, in contrast to the EABP, the EGAN recruits an additional assisting network $D$ and trains the parameters of the saliency generator $T$ by an adversarial loss without relying on inference. However, the inference step is still a necessary part of training the EBM prior.

At a testing stage, we can follow the same testing pipeline as in Algorithm \ref{alg_eabp_test}, where we first sample latent variables from the EBM prior through Langevin dynamics and then transform the latent variables and the input testing image by the saliency generator to produce a saliency prediction.


\subsubsection{An EBM Prior for VAE-based Generative Saliency}

We can also train the generative saliency prediction model with an energy-based prior through variational inference \cite{vae_bayes_kumar, structure_output,PangW21}. This can be achieved by introducing an extra tractable inference model $q_{\phi}(z|y,x)$, with parameters $\phi$, to approximate the intractable posterior distribution $p_{\beta}(z|y,x)$. Following the reparameterization trick \cite{vae_bayes_kumar}, we design the inference model $q_{\phi}(z|y,x)=\mathcal{N}(\mu_{\phi}(y,x), \text{diag}(\sigma_{\phi}(y,x)))$, where $\mu_{\phi}(y,x)$ and $\sigma_{\phi}(y,x)$ are $d$-dimensional outputs of encoding bottom-up networks that take $x$ and $y$ as input. 


Given an image $x$ with its saliency map $y$, the log-likelihood $\log p_{\beta}(y|x)$ is lower bounded by the evidence lower bound (ELBO), which is:
    \begin{align}    
     &\text{ELBO}(\beta, \phi|x,y) \notag\\
     =& \log p_{\beta}(y|x) -  {\mathbb{D}_{\text{KL}}}(q_{\phi}(z|y,x) || p_{\beta}(z|y,x)) \notag \\
     =& \mathbb{E}_{q_{\phi}(z|y,x)}[\log p_{\theta}(y|z,x)] -  {\mathbb{D}_{\text{KL}}}(q_{\phi}(z|y,x) || p_{\alpha}(z|x)), \label{CVAE_loss}
    \end{align}
where we assume $z$ and $x$ are independent, so that we can simply set the prior model $p_{\alpha}(z|x)=p_{\alpha}(z) \propto \exp[-U_{\alpha}(z)]p_0(z)$. As to Eq.~(\ref{CVAE_loss}), the first term is the reconstruction loss and the second one is the Kullback-Leibler divergence between the inference network and the EBM prior distribution. 

For the prior model $\alpha$, the learning gradient is:    
    \begin{align}
     & \nabla_{\alpha}\text{ELBO}(\beta, \phi|x,y) \notag\\
     =& \mathbb{E}_{p_{\alpha}(z)}[\nabla_{\alpha}U(z)]-\mathbb{E}_{q_{\phi}(z|y,x)}[\nabla_{\alpha}U(z)], \label{CVAE_prior_loss}   
    \end{align}
where $\mathbb{E}_{p_{\alpha}(z)}$ is approximated by MCMC samples from the EBM prior $p_{\alpha}(z)$, and $\mathbb{E}_{q_{\phi}(z|y,x)}$ is approximated by samples from the inference network.   

For the bottom-up inference model $\phi$ and the top-down saliency generator $\theta$, let $\psi=\{\theta, \phi\}$ and the learning gradients of $\psi$ are given by:    
    \begin{align}
      &\nabla_{\psi}\text{ELBO}(\beta, \phi|x,y) \notag = \nabla_{\psi} \mathbb{E}_{q_{\phi}(z|y,x)}[\log p_{\theta}(y|z,x)] \\ &-  \nabla_{\psi}{\mathbb{D}_{\text{KL}}}(q_{\phi}(z|y,x) || p_{0}(z)) -\nabla_{\psi}\mathbb{E}_{q_{\phi}(z|y,x)}[U_{\alpha}(z)] \label{CVAE_encoder_decoder_loss}
    \end{align}
where the first term is the derivative of the reconstruction loss and the second term is the derivative of the Kullback-Leibler divergence between the inference distribution and the Gausian reference distribution $p_0(z)$ in the EBM prior model. ${\mathbb{D}_{\text{KL}}}(q_{\phi}(z|y,x) || p_{0}(z))$ is tractable. The third term is approximated by the samples from the inference network. In contrast to the conventional VAE loss \cite{vae_bayes_kumar} using a Gaussian prior, Eq.~(\ref{CVAE_encoder_decoder_loss}) has an extra term, which is the third term involving the energy function $U_{\alpha}(z)$. 

\subsection{Network Architectures}\label{network_sec}

\subsubsection{Saliency Generator}

\begin{figure}[t]
\centering
\includegraphics[width=1.0\linewidth]{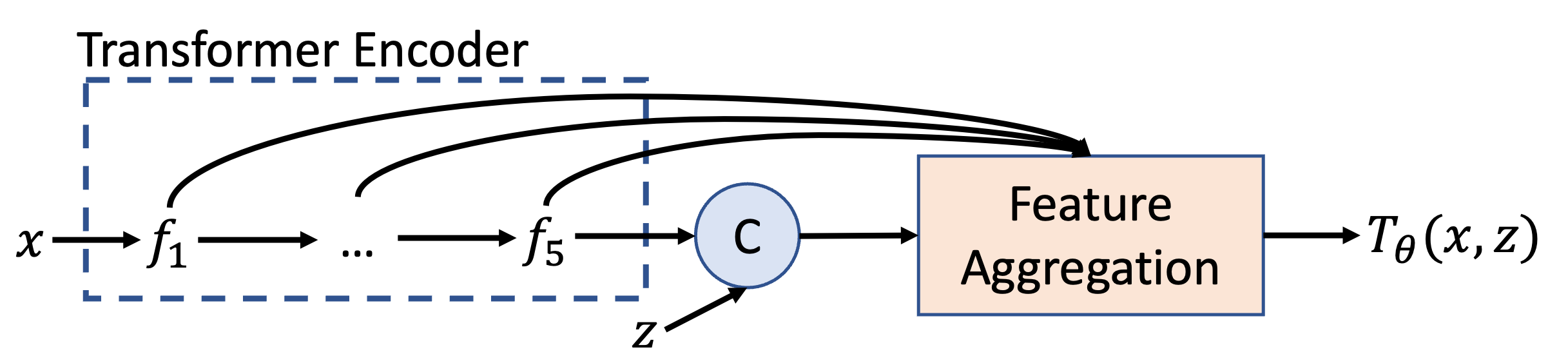}
\caption{\Rev{Architecture of the saliency generator for saliency generation.
}}
\label{fig:generative_transformer}
\end{figure}

We utilize the Swin transformer backbone~\cite{liu2021swin} to design our saliency generator and perform ablation studies with CNN backbones in Section \ref{sec:backbone_analysis}. As depicted in Figure~\ref{fig:generative_transformer}, our saliency generator takes a three-channel image $x$ and the latent variables $z$ as inputs and generates a one-channel saliency map $T_\theta(x,z)$. Our saliency generator includes two main modules: the \enquote{Transformer Encoder} module and the \enquote{Feature Aggregation} module. The former module takes $x$ as input and produces a set of feature maps $\{f_l\}_{l=1}^5$ of channel sizes 128, 256, 512, 1024, and 1024, respectively. The latter module takes $\{f_l\}_{l=1}^5$ and the vector of latent variables $z$ as inputs to generate the saliency prediction $y$. To achieve this, we first apply a multi-scale dilated convolutional layer~\cite{denseaspp} to each $f_l$ to reduce the channel dimension of each level feature $f_l$ to a channel size of $C=32$. This results in a new set of feature maps $\{f'_l\}_{l=1}^5$. Next, we spatially replicate the vector $z$ and concatenate it channel-wise with $f'_5$, followed by a $3\times3$ convolutional layer that produces a feature map $F_5$ with the same number of channels as $f'_5$. Finally, we sequentially concatenate feature maps from high to low levels via feature aggregation, i.e., from $l=4$ to 1, we compute $F_l=\text{Conv}_{3 \times 3}(\text{M}(\text{Concat}(f'_l,F_{l+1},...,F_5)))$, where $\text{Conv}_{3 \times 3}(\cdot)$ is a $3\times3$ convolutional layer that reduces the channel dimension to $32$, $\text{M}(\cdot)$ is the channel attention module~\cite{rca_eccv}, and $\text{Concat}(\cdot)$ is the channel-wise concatenation operation. It is important to note that we upsample the higher-level feature map to the same spatial size as that of the lower-level feature map before the concatenation operation. We then feed $F_1$ to a $3\times3$ convolutional layer to obtain the one-channel saliency map $T_\theta(x,z)$.

\subsubsection{Energy-based Prior Model}
We develop an energy-based model to represent the prior distribution of the latent variables $z$, utilizing a multilayer perceptron (MLP) to parameterize the energy function $U_{\alpha}(z)$. Our MLP comprises three fully connected layers that map the latent variables $z$ to a scalar. The feature map sizes for the MLP's different layers are $C_e$, $C_e$, and 1, respectively. For the size of the EBM prior, we will denote it as $C_e$ and set it to 60 for our experiment. We use GELU~\cite{hendrycks2016gaussian} activation after each layer except the last one.

\subsubsection{Discriminator in the  EGAN} 

The discriminator $D_\gamma$ in EGAN consists of five $3\times 3$ convolutional layers that take as input the concatenation of the image $x$ and the prediction $T_\theta(x,z)$ (or the ground truth map $y$), and output an all-zero feature map $\textbf{0}$ (or an all-one feature map $\textbf{1}$). The first four convolutional layers have a channel size of $C_d = 64$, and the last convolutional layer has a channel size of 1. To increase the receptive field, we apply a stride of 2 during the convolution operation in the $1^{st}$, $3^{rd}$, and $5^{th}$ layers. We also apply batch normalization and Leaky RELU after the first four convolutional~layers.


\subsubsection{Inference Model in the EVAE} 
The inference model for approximate posterior distribution $q_{\phi}(z|x,y)$ is composed of five convolutional layers and two fully connected layers. The network maps the concatenation of the RGB image $x$ and its saliency map $y$ to feature maps with channel sizes of $C_i$, $2 \times C_i$, $4 \times C_i$, $8 \times C_i$, and $8 \times C_i$ through five convolutional layers using a kernel size of $4 \times 4$ and a stride of $2$. We set $C_i=32$. The subsequent two fully connected layers map the $8 \times C$-channel feature map to the mean $\mu$ of the latent variables $z$ and its standard deviation $\sigma$. We perform batch normalization and Leaky RELU after each convolution operation. The inference distribution of the latent variables $z$ is then obtained via the reparameterization trick: $z=\mu+\epsilon\times\sigma$, where $\epsilon\sim\mathcal{N}(0,I)$.

\begin{table*}[ht!]
  \centering
  \scriptsize
  \renewcommand{\arraystretch}{1.2}
  \renewcommand{\tabcolsep}{0.55mm}
  \caption{Performance comparison with benchmark RGB salient object detection models.}
  \begin{tabular}{l|cccc|cccc|cccc|cccc|cccc|cccc}
\toprule
  &\multicolumn{4}{c|}{DUTS~\cite{imagesaliency}}&\multicolumn{4}{c|}{ECSSD~\cite{yan2013hierarchical}}&\multicolumn{4}{c|}{DUT~\cite{Manifold-Ranking:CVPR-2013}}&\multicolumn{4}{c|}{HKU-IS~\cite{li2015visual}}&\multicolumn{4}{c|}{PASCAL-S~\cite{pascal_s_dataset}}& \multicolumn{4}{c}{SOD \cite{sod_dataset}}\\
    Method & $S_{\alpha}\uparrow$&$F_{\beta}\uparrow$&$E_{\xi}\uparrow$&$\mathcal{M}\downarrow$& $S_{\alpha}\uparrow$&$F_{\beta}\uparrow$&$E_{\xi}\uparrow$&$\mathcal{M}\downarrow$& $S_{\alpha}\uparrow$&$F_{\beta}\uparrow$&$E_{\xi}\uparrow$&$\mathcal{M}\downarrow$& $S_{\alpha}\uparrow$&$F_{\beta}\uparrow$&$E_{\xi}\uparrow$&$\mathcal{M}\downarrow$& $S_{\alpha}\uparrow$&$F_{\beta}\uparrow$&$E_{\xi}\uparrow$&$\mathcal{M}\downarrow$& $S_{\alpha}\uparrow$&$F_{\beta}\uparrow$&$E_{\xi}\uparrow$&$\mathcal{M}\downarrow$ \\ \hline
   CPD~\cite{cpd_sal} & .869 & .821 & .898 & .043 & .913 & .909 & .937 & .040 & .825 & .742 & .847 & .056 & .906 & .892 & .938 & .034 & .848 & .819 & .882 & .071 & .799 & .779 & .811 & .088 \\
   SCRN~\cite{scrn_sal} & .885 & .833 & .900 & .040 & .920 & .910 & .933 & .041 & .837 & .749 & .847 & .056 & .916 & .894 & .935 & .034 & .869 & .833 & .892 & .063 & .817 & .790 & .829 & .087 \\ 
   PoolNet~\cite{Liu19PoolNet} & .887 & .840 & .910 & .037 & .919 & .913 & .938 & .038 & .831 & .748 & .848 & .054 & .919 & .903 & .945 & .030 & .865 & .835 & .896 & .065 & .820 & .804 & .834 & .084  \\ 
    BASNet~\cite{qin2019basnet} & .876 & .823 & .896 & .048 & .910 & .913 & .938 & .040 & .836 & .767 & .865 & .057 & .909 & .903 & .943 & .032 & .838 & .818 & .879 & .076 & .798 & .792 & .827 & .094 \\ 
   EGNet~\cite{zhao2019EGNet} & .878 & .824 & .898 & .043 & .914 & .906 & .933 & .043 & .840 & .755 & .855 & .054 & .917 & .900 & .943 & .031 & .852 & .823 & .881 & .074 & .824 & .811 & .843 & .081  \\
   F3Net~\cite{wei2020f3net} & .888 & .852 & .920 & .035 & .919 & .921 & .943 & .036 & .839 &  .766 & .864 & .053 & .917 & .910 & .952 & .028 & .861 & .835 & .898 & .062 & .824 & .814 & .850 & .077 \\
   ITSD~\cite{zhou2020interactive} & .886 & .841 & .917 & .039 & .920 & .916 & .943 & .037 & .842 & .767 & .867 & .056 & .921 & .906 & .950 & .030 & .860 & .830 & .894 & .066 & .836 & .829 & .867 & .076 \\
  SCNet \cite{zhang2021energy} & .902 & .870 & .936 & .032 & .928 & .930 & .955 & .030 & .847 & .778 & .879 & .053& .927 & .917 & .960 & .026 & .873 & .846 & .909 & .058 & .824 & .820 & .849 & .076 \\
   LDF~\cite{wei2020label} & .892 & .861 & .925 & .034 & .919 & .923& .943 & .036 & .839 & .770 & .865 & .052 & .920 & .913 & .953 & .028 & .860 & .856 & .901 & .063 & .826 & .822 & .852 & .075 \\ 
   UIS \cite{jing2020uncertainty}  & .888 & .860 & .927 & .034 & .921 & .926 & .947 & .035 & .839 & .773 & .869 & .051 & .921 & .919 & .957 & .026 & .848 & .836 & .899 & .063  & .808 & .808 & .847 & .079 \\ 
PAKRN \cite{xu2021locate}& .900 & .876 & .935 & .033  & .928 & .930 & .951 & .032 & .853 & .796 & .888 & .050 & .923 & .919 & .955 & .028 & .859 & .856 & .898 & .068  & .833 & .836 & .866 & .074 \\ 
MSFNet \cite{Miao_2021_ACM_MM} & .877 & .855 & .927 & .034  & .915 & .927 & .951 & .033 & .832 & .772 & .873 & .050 & .909 & .913 & .957 & .027 & .849 & .855 & .900 & .064  & .813 & .822 & .852 & .077  \\ 
   VST~\cite{Liu_2021_ICCV_VST} &.896 &.842 &.918 &.037 &.932 &.911 &.943 &.034 &.850 &.771 &.869 &.058 &.928 &.903 &.950 &.030 &.873 &.832 &.900 &.067  & .851 & .833 & .879 & .069 \\ \hline
EABP  &.908 &.875 &.942 &.029 &.935 &.935 &.962&.026 &.858 &.797 &.892 &.051 &.930 &.922 &.964 &.023 &.877 &.855 &.915 &.054  & .860 & \textbf{.860} & \textbf{.898} & \textbf{.061} \\
EGAN &.910 &.881 &\textbf{.948} &.028 &\textbf{.941} &.939 &\textbf{.967}&\textbf{.025} &.859 &.800 &\textbf{.896} &.049 &\textbf{.937} &\textbf{.926} &\textbf{.969} &\textbf{.022} &.879 &.858 &.918 &.053  & .862 & \textbf{.860} & \textbf{.898} & \textbf{.061} \\
EVAE &\textbf{.911} &\textbf{.884} &.947 &\textbf{.027} &.940 &\textbf{.941} &.965&\textbf{.025} &\textbf{.862} &\textbf{.805} &.892 &\textbf{.046} &.930 &.923 &.963 &.023 &\textbf{.881} &\textbf{.862} &\textbf{.920} &\textbf{.051}  & \textbf{.863} & .857 & .893 & .062 \\ 
\bottomrule
  \end{tabular}
  \label{tab:benchmark_rgb_sod}
\end{table*}

\begin{figure*}[!htp]
   \begin{center}
   \begin{tabular}{c@{ } c@{ } c@{ } c@{ } c@{ } c@{ } c@{ }}
   {\includegraphics[width=0.135\linewidth]{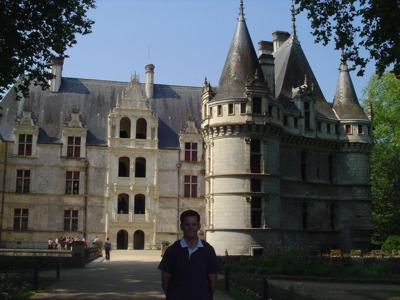}}&
   {\includegraphics[width=0.135\linewidth]{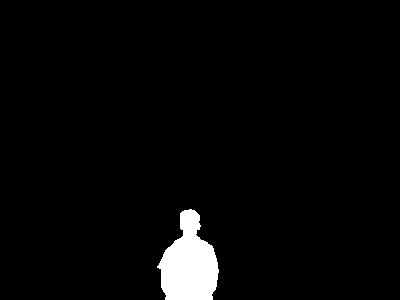}}&
   {\includegraphics[width=0.135\linewidth]{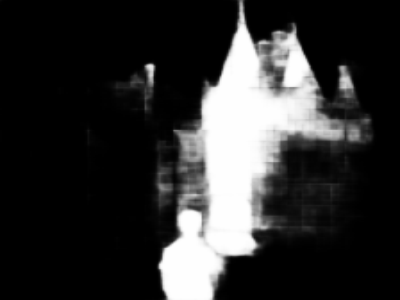}}&
   {\includegraphics[width=0.135\linewidth]{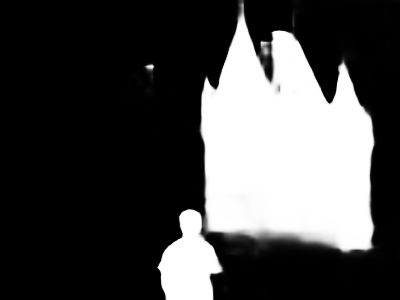}}&
   {\includegraphics[width=0.135\linewidth]{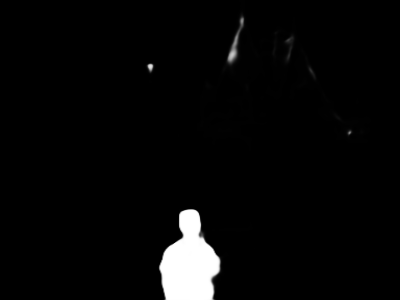}}&
   {\includegraphics[width=0.135\linewidth]{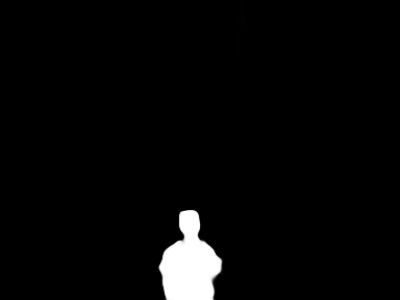}}&
   {\includegraphics[width=0.135\linewidth]{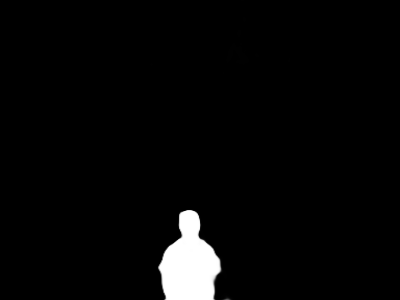}}\\
   {\includegraphics[width=0.135\linewidth]{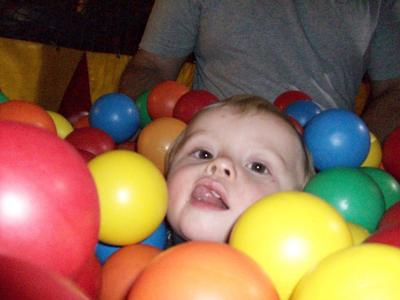}}&
   {\includegraphics[width=0.135\linewidth]{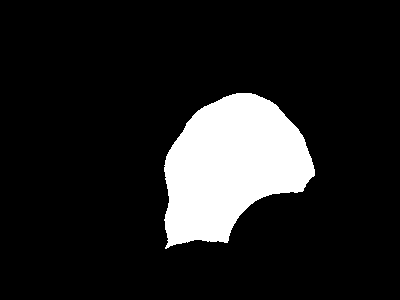}}&
   {\includegraphics[width=0.135\linewidth]{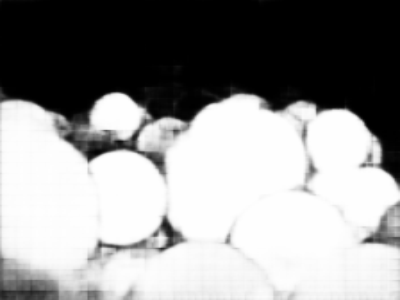}}&
   {\includegraphics[width=0.135\linewidth]{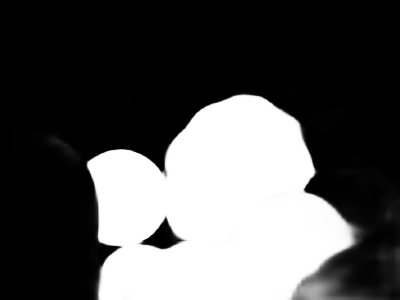}}&
   {\includegraphics[width=0.135\linewidth]{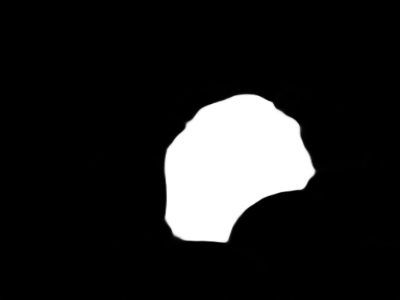}}&
   {\includegraphics[width=0.135\linewidth]{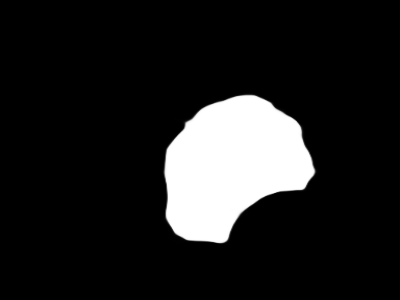}}&
   {\includegraphics[width=0.135\linewidth]{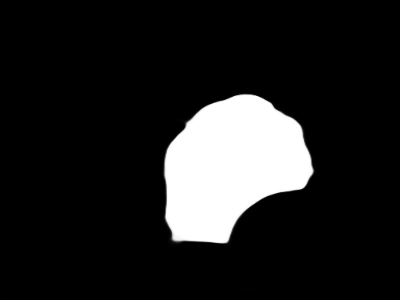}}\\
    \footnotesize{Image}& \footnotesize{GT} & \footnotesize{VST \cite{Liu_2021_ICCV_VST}} & \footnotesize{LDF \cite{wei2020label}} & \footnotesize{EGAN (ours)} & \footnotesize{EVAE (ours)} & \footnotesize{EABP (ours)} \\
   \end{tabular}
   \end{center}
   \caption{A qualitative comparison of our RGB SOD generative models with two state-of-the-art models, VST \cite{Liu_2021_ICCV_VST}, a transformer-based saliency detection model, and LDF \cite{wei2020label}, a CNN-based RGB saliency detection model.}
\label{fig:visual_comparison_rgb_sod}
\end{figure*}

\section{Experimental Results}

\subsection{Setup}
\label{setup_sec}

\subsubsection{Tasks, Datasets, and Evaluation Metrics} 
In this paper, we address two common static image-based saliency detection tasks: (i) RGB saliency detection and (ii) RGB-D saliency detection. For RGB saliency detection,
we train models on DUTS training  set~\cite{imagesaliency}, which consists of 10,553 images with the corresponding pixel-wise annotations.
We then test model performance on six benchmark RGB saliency detection testing datasets, including the DUTS testing set,  ECSSD~\cite{yan2013hierarchical}, DUT~\cite{Manifold-Ranking:CVPR-2013}, HKU-IS~\cite{li2015visual}, PASCAL-S~\cite{pascal_s_dataset}
and the
SOD testing dataset~\cite{sod_dataset}. 
For RGB-D saliency detection, our aim is to identify the most visually distinctive objects in a scene from the provided RGB and depth data. We follow the conventional training setting, in which the training set is a combination of 1,485 images from NJU2K
dataset~\cite{NJU2000} and 700 images from NLPR dataset~\cite{peng2014rgbd}. We test the trained models on NJU2K testing set, NLPR testing set,
LFSD~\cite{li2014saliency}, 
DES~\cite{cheng2014depth}, SSB~\cite{niu2012leveraging} and SIP~\cite{sip_dataset} testing set. We employ four evaluation metrics, including Mean Absolute Error $\mathcal{M}$, Mean F-measure ($F_{\beta}$), Mean E-measure ($E_{\xi}$)~\cite{fan2018enhanced} and S-measure ($S_{\alpha}$)~\cite{fan2017structure}, to measure the performance of both RGB saliency detection models and RGB-D saliency detection models.

\begin{figure}[!htp]
   \begin{center}
   \begin{tabular}{c@{ } c@{ } c@{ } c@{ } c@{ }}
   {\includegraphics[width=0.185\linewidth]{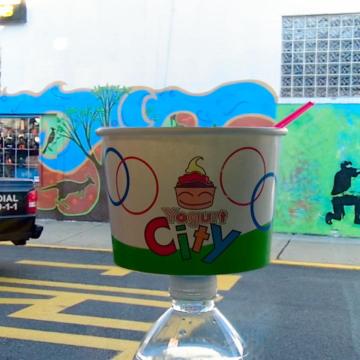}}&
   {\includegraphics[width=0.185\linewidth]{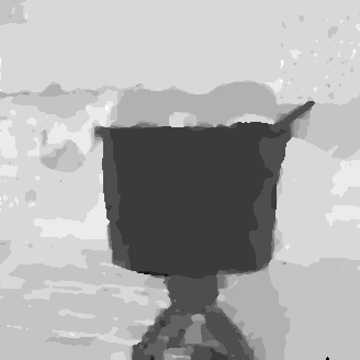}}&
   {\includegraphics[width=0.185\linewidth]{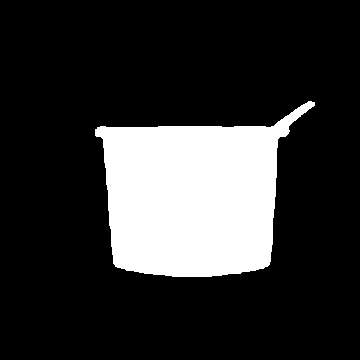}}&
   {\includegraphics[width=0.185\linewidth]{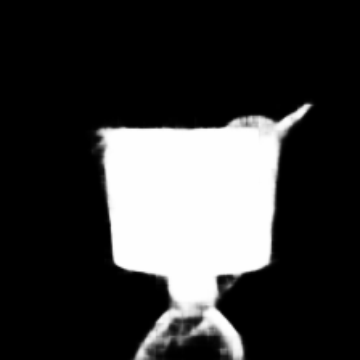}}&
   {\includegraphics[width=0.185\linewidth]{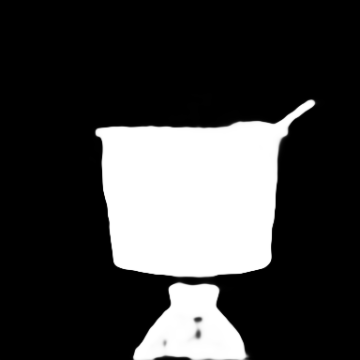}}\\
   {\includegraphics[width=0.185\linewidth]{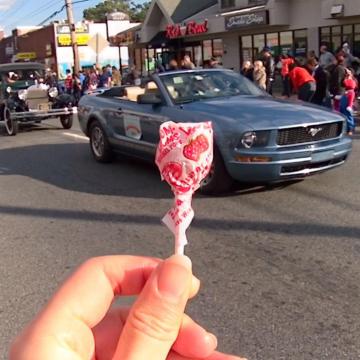}}&
   {\includegraphics[width=0.185\linewidth]{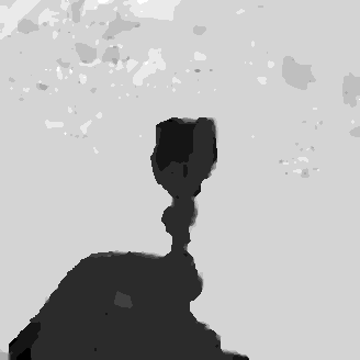}}&
   {\includegraphics[width=0.185\linewidth]{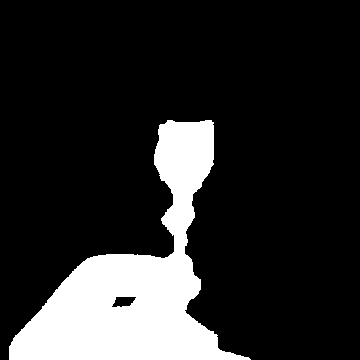}}&
   {\includegraphics[width=0.185\linewidth]{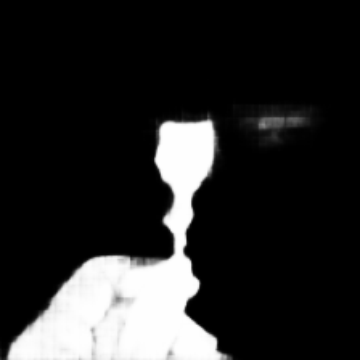}}&
   {\includegraphics[width=0.185\linewidth]{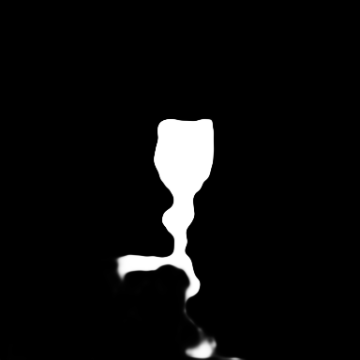}}\\
   {\includegraphics[width=0.185\linewidth]{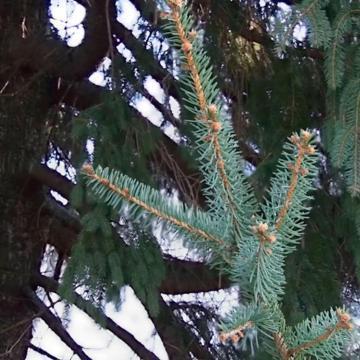}}&
   {\includegraphics[width=0.185\linewidth]{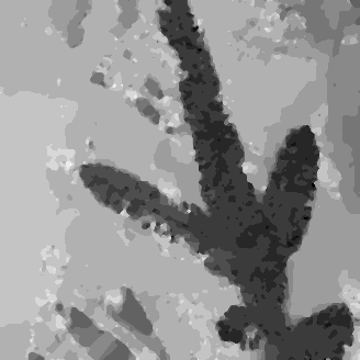}}&
   {\includegraphics[width=0.185\linewidth]{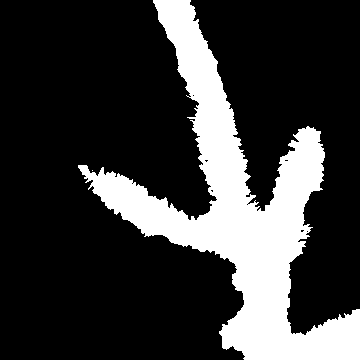}}&
   {\includegraphics[width=0.185\linewidth]{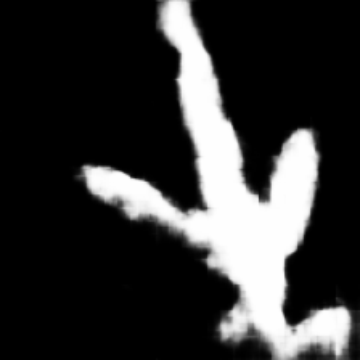}}&
   {\includegraphics[width=0.185\linewidth]{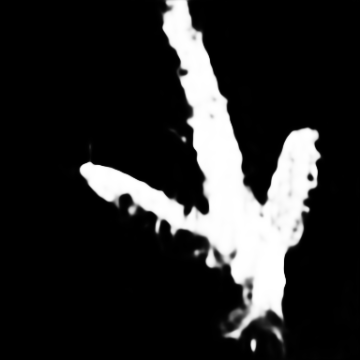}}\\
    \footnotesize{Image}& \footnotesize{Depth}& \footnotesize{GT} & \footnotesize{VST \cite{Liu_2021_ICCV_VST}} & \footnotesize{EGAN} \\
   \end{tabular}
   \end{center}
   \caption{An illustration of examples from LFSD dataset \cite{li2014saliency}, in which the state-of-the-art model, VST \cite{Liu_2021_ICCV_VST} that uses a cross-level-fusion network architecture, outperforms our EGAN framework that adopts an early-fusion network architecture.}
\label{fig:vst_outperform_egan}
\end{figure}

\subsubsection{Implementation Details}
Our generative model is built upon the Swin transformer~\cite{liu2021swin} and we use it as the backbone of the encoder in our framework. The Swin backbone is initialized with the parameters pretrained on the ImageNet-1K~\cite{imagenet_1k} dataset for image classification. The other parameters of the newly added components, including the decoder part, the MLP of the energy based prior model, the fully convolutional discriminator within EGAN and the prior along with the inference net within EVAE will be randomly initialized from the Gaussian distribution $\mathcal{N}(0,0.01)$.
Empirically we set the number of dimensions $d$ of the latent variables $z$ as $d=32$. We set $\sigma_{\epsilon}=1$ in Eq.~(\ref{eq:generative_sod_model})  and $\sigma_{z}=1$ in Eq.~(\ref{eq:ebm_prior}). 
We resize all the images and the saliency maps to the resolution of $384\times384$ pixels to fit the Swin transformer. The maximum epoch is 50 for RGB saliency detection and 100 for RGB-D saliency detection. The initial learning rates of the saliency generator, the discriminator and the EBM are $2.5 \times 10^{-5}$, $1 \times 10^{-5}$ and $1 \times 10^{-4}$ respectively. For both the prior and posterior Langevin processes, we set the numbers of Langevin steps $K^{-}=K^{+}=6$. The Langevin step sizes of the prior and posterior are $\delta^{-}=0.4$ and $\delta^{+}=0.1$. For RGB-D saliency detection, we perform early fusion by concatenating the RGB image and the depth data at the input layer. We then feed this concatenated data to a $3\times3$ convolutional layer before passing it to the Swin encoder \cite{liu2021swin}. As to the computational cost of EBM prior-based models, the complete training process requires roughly 17 hours, using a batch size of $n'=10$, on a single NVIDIA GTX 2080Ti GPU for an RGB saliency prediction model. For a model related to RGB-D saliency, the training time is about 12 hours. During testing, our model is capable of processing 15 images per second with one time of sampling from the EBM prior distribution.

\begin{table*}[ht!]
  \centering
  \scriptsize
  \renewcommand{\arraystretch}{1.2}
  \renewcommand{\tabcolsep}{0.55mm}
  \caption{Performance comparison with benchmark RGB-D salient object detection models.}
  \begin{tabular}{l|cccc|cccc|cccc|cccc|cccc|cccc}
\toprule
  &\multicolumn{4}{c|}{NJU2K~\cite{NJU2000}}&\multicolumn{4}{c|}{SSB~\cite{niu2012leveraging}}&\multicolumn{4}{c|}{DES~\cite{cheng2014depth}}&\multicolumn{4}{c|}{NLPR~\cite{peng2014rgbd}}&\multicolumn{4}{c|}{LFSD \cite{li2014saliency}}&\multicolumn{4}{c}{SIP~\cite{sip_dataset}} \\
    Method & $S_{\alpha}\uparrow$&$F_{\beta}\uparrow$&$E_{\xi}\uparrow$&$\mathcal{M}\downarrow$& $S_{\alpha}\uparrow$&$F_{\beta}\uparrow$&$E_{\xi}\uparrow$&$\mathcal{M}\downarrow$& $S_{\alpha}\uparrow$&$F_{\beta}\uparrow$&$E_{\xi}\uparrow$&$\mathcal{M}\downarrow$& $S_{\alpha}\uparrow$&$F_{\beta}\uparrow$&$E_{\xi}\uparrow$&$\mathcal{M}\downarrow$& $S_{\alpha}\uparrow$&$F_{\beta}\uparrow$&$E_{\xi}\uparrow$&$\mathcal{M}\downarrow$& $S_{\alpha}\uparrow$&$F_{\beta}\uparrow$&$E_{\xi}\uparrow$&$\mathcal{M}\downarrow$ \\ \hline
   BBSNet~\cite{fan2020bbs}  &.921 &.902 &.938 &.035  &.908 &.883 &.928 &.041 &.933 &.910 &.949 &.021 &.930 &.896 &.950 &.023 &.864 &.843 &.883 &.072 &.879 &.868 &.906 &.055 \\
   BiaNet~\cite{zhang2020bilateral}  &.915 &.903 &.934 &.039  &.904 &.879 &.926 &.043 &.931 &.910 &.948 &.021 &.925 &.894 &.948 &.024 &.845 &.834 &.871 &.085 &.883 &.873 &.913 &.052 \\
   CoNet~\cite{ji2020accurate}  &.911 &.903 &.944 &.036  &.896 &.877 &.939 &.040 &.906 &.880 &.939 &.026 &.900 &.859 &.937 &.030  &.842 &.834 &.886 &.077 &.868 &.855 &.915 &.054 \\
   UCNet~\cite{ucnet_sal} &.897 &.886 &.930 &.043 &.903 &.884 &.938 &.039 &.934 &.919 &.967 &.019 &.920 &.891 &.951 &.025 &.864 &.855 &.901 &.066 &.875 &.867 &.914 &.051 \\
   JLDCF~\cite{Fu2020JLDCF} &.902 &.885 &.935 &.041  &.903 &.873 &.936 &.040 &.931 &.907 &.959 &.021 &.925 &.894 &.955 &.022 &.862 &.848 &.894 &.070  &.880 &.873 &.918 &.049 \\ 
   DSA2F~\cite{Sun_2021_CVPR_DSA2F} &.903 &.901 &.923 &.039  &.904 &\textbf{.898} &.933 &.036 &.920 &.896 &.962 &.021 &.918 &.897 &.950 &.024 &.882 &\textbf{.878} &\textbf{.919} &.055  &- &- &- &- \\ 
   VST~\cite{Liu_2021_ICCV_VST} &.922 &.898 &.939 &.035 &.913 &.879 &.937 &.038 &.943 &.920 &.965 &.017 &.932 &.897 &.951 &.024 &\textbf{.890} &.871 &.917 &\textbf{.054}  &.904 &.894 &.933 &.040 \\ \hline
EABP &.929 &\textbf{.924} &.956 &.028 &.916 &\textbf{.898} &.950 &\textbf{.032}&.945 &.928 &.971 &\textbf{.016} &\textbf{.938} &.921 &.966 &\textbf{.018}&.872 &.862 &.901 &.066  &\textbf{.906} &\textbf{.908} &.940 &\textbf{.037}  \\
EGAN &\textbf{.931} &.923 &\textbf{.957} &\textbf{.027} &\textbf{.918} &.896 &\textbf{.954}&\textbf{.032} &\textbf{.948} &\textbf{.930} &\textbf{.977} &\textbf{.016} &.936 &\textbf{.928} &\textbf{.968} &\textbf{.018} &.877 &.869 &.910 &.060  & \textbf{.906} & .906 & \textbf{.947} & \textbf{.037} \\
EVAE &.926 &.919 &.955 &.028 &.911 &.892 &.946&.033 &.940 &.929 &.973 &\textbf{.016} &.933 &.914 &.963 &.019 &.875 &.870 &.908 &.060  & .896 & .901 & .935 & \textbf{.037} \\
   \bottomrule
  \end{tabular}
  \label{tab:benchmark_rgbd_sod}
\end{table*}

\subsection{Performance Comparison}

\subsubsection{RGB SOD Performance Comparison} We compare the proposed EBM prior-based generative saliency prediction models, including EABP, EVAE, and EGAN, with the state-of-the-art (SOTA) RGB SOD models, and show performance comparison in Table \ref{tab:benchmark_rgb_sod}.
All of these models utilize the Swin transformer backbone \cite{liu2021swin}. 
Notably, our generative models' performance is reported by averaging the results of ten prediction generation trials. As Table \ref{tab:benchmark_rgb_sod} illustrates, our proposed frameworks outperforms the state-of-the-art models, exhibiting the superiority of generative saliency prediction models.


Furthermore, we provide visualizations of the predictions generated by our models and two state-of-the-art (SOTA) models in Figure~\ref{fig:visual_comparison_rgb_sod} to analyze the effectiveness of our proposed framework.  Specifically, for the input RGB image shown in the first row, the salient foreground is dark with poor illumination. The SOTA models produce numerous false positives, while our models accurately localize the entire salient foreground. The salient foreground of the second RGB image in the second row is in a cluttered background, which is commonly considered a complex case for saliency detection. Existing techniques struggle to distinguish the salient region from the complex background. However, the precise predictions of all three of our generative models provide further evidence of the superiority of our generative frameworks.

\subsubsection{RGB-D SOD Performance Comparison} 

We compare the performance of our RGB-D saliency detection models, including EABP, EVAE, and EGAN, with state-of-the-art (SOTA) RGB-D SOD models in Table~\ref{tab:benchmark_rgbd_sod}. 
Our \enquote{early-fusion network architecture} show a clear advantage compared to existing techniques on five out of six benchmark testing datasets. However, we observe that our generative models perform worse on the LFSD dataset \cite{li2014saliency} than the state-of-the-art RGB-D saliency prediction models that use a cross-level-fusion network architecture, namely DSA2F \cite{Sun_2021_CVPR_DSA2F} and VST \cite{Liu_2021_ICCV_VST}. LFSD \cite{li2014saliency} is a small RGB-D saliency detection dataset consisting of 100 RGB-D images. To better understand the limitations of our models on this dataset, we visualize some examples where the VST \cite{Liu_2021_ICCV_VST} outperforms our model (EGAN in this case) in Figure~\ref{fig:vst_outperform_egan}.

In our analysis, we observe that the appearance information from the RGB image in the first row contributes more to saliency detection, whereas equal contribution of appearance and geometric information are observed for the second example. However, the geometric information dominates RGB-D saliency for the third example. Our failure cases in Figure~\ref{fig:vst_outperform_egan} further verify the necessity of effective cross-level feature fusion. Although our model performs early-fusion, it may work inferiorly compared to state-of-the-art cross-level fusion models \cite{Sun_2021_CVPR_DSA2F,Liu_2021_ICCV_VST}. To address this issue, we 
plan to introduce an effective cross-level fusion strategy within our generative saliency detection framework.



\begin{table*}[ht!]
  \centering
  \scriptsize
  \renewcommand{\arraystretch}{1.2}
  \renewcommand{\tabcolsep}{0.5mm}
  \caption{\Rev{Backbone analysis of the proposed EBM prior-based generative saliency detection framework for RGB saliency detection, where the bold numbers represent the best performance within each backbone model.}}
  \begin{tabular}{c|c|cccc|cccc|cccc|cccc|cccc|cccc}
\toprule
  &&\multicolumn{4}{c|}{DUTS~\cite{imagesaliency}}&\multicolumn{4}{c|}{ECSSD~\cite{yan2013hierarchical}}&\multicolumn{4}{c|}{DUT~\cite{Manifold-Ranking:CVPR-2013}}&\multicolumn{4}{c|}{HKU-IS~\cite{li2015visual}}&\multicolumn{4}{c|}{PASCAL-S~\cite{pascal_s_dataset}}& \multicolumn{4}{c}{SOD \cite{sod_dataset}}\\
    Backbone& Method & $S_{\alpha}\uparrow$&$F_{\beta}\uparrow$&$E_{\xi}\uparrow$&$\mathcal{M}\downarrow$& $S_{\alpha}\uparrow$&$F_{\beta}\uparrow$&$E_{\xi}\uparrow$&$\mathcal{M}\downarrow$& $S_{\alpha}\uparrow$&$F_{\beta}\uparrow$&$E_{\xi}\uparrow$&$\mathcal{M}\downarrow$& $S_{\alpha}\uparrow$&$F_{\beta}\uparrow$&$E_{\xi}\uparrow$&$\mathcal{M}\downarrow$& $S_{\alpha}\uparrow$&$F_{\beta}\uparrow$&$E_{\xi}\uparrow$&$\mathcal{M}\downarrow$& $S_{\alpha}\uparrow$&$F_{\beta}\uparrow$&$E_{\xi}\uparrow$&$\mathcal{M}\downarrow$ \\ \hline
   \multirow{4}{*}{ResNet50} & EGAN&.881 &.838 &.915 &.039 &.915 &.915 &.942 &.037 &.829 &.752 &.862 &.057 &.913 &.903 &.949 &.030 &.852 &.825 &.894 &.066  & \textbf{.839} & \textbf{.835} & \textbf{.883} & \textbf{.070} \\ 
   & EVAE&.879 &.833 &.913 &.039 &.915 &.910 &.942 &.037 &.820 &.737 &.848 &.061 &.914 &.900 &.949 &.030 &.856 &.824 &.895 &\textbf{.065}  & .833 & .828 & .869 & .074 \\
   & EABP&.877 &.833 &.910 &.041 &.916 &.914 &.941 &.039 &.825 &.746 &.855 &.061 &.915 &.904 &.950 &.031 &.855 &.826 &.894 &.067  & .829 & .824 & .865 & .075 \\
   & EABP\_A&\textbf{.896} &\textbf{.859} &\textbf{.923} &\textbf{.035} &\textbf{.925} &\textbf{.927} &\textbf{.947} &\textbf{.034} &\textbf{.840} &\textbf{.784} &\textbf{.871} &\textbf{.053} &\textbf{.919} &\textbf{.908} &\textbf{.956} &\textbf{.028} &\textbf{.861} &\textbf{.848} &\textbf{.900} &\textbf{.065}  & .831 & .828 & .870 & .073 \\
   \hline
   \multirow{4}{*}{VGG16} & EGAN&.863 &.819 &.900 &.042 &.905 &.903 &.936 &.042 &.810 &.723 &.842 &.060 &.906 &.893 &.944 &.033 &.848 &\textbf{.828} &\textbf{.899} &.067  & .816 & .818 & .867 & .082 \\ 
   & EVAE&.863 &.817 &.901 &.044 &.898 &.898 &.925 &.045 &.807 &.722 &.839 &.062 &.902 &.893 &.939 &.034 &.846 &.821 &.887 &.070  & .812 & .808 & .851 & .080 \\
   & EABP&.870 &.827 &.909 &.042 &.905 &.905 &.933 &.042 &.814 &.734 &.847 &.061 &.907 &.900 &.945 &.032 &.847 &.822 &.889 &.069  & .813 & .810 & .848 & .080 \\
   & EABP\_A&\textbf{.876} &\textbf{.831} &\textbf{.912} &\textbf{.040} &\textbf{.911} &\textbf{.910} &\textbf{.940} &\textbf{.039} &\textbf{.825} &\textbf{.745} &\textbf{.856} &\textbf{.059} &\textbf{.913} &\textbf{.903} &\textbf{.951} &\textbf{.030} &\textbf{.853} &.826 &.892 &\textbf{.066}  & \textbf{.826} & \textbf{.821} & \textbf{.869} & \textbf{.078} \\
   \hline
  \multirow{3}{*}{Swin} & EGAN&.910 &.881 &\textbf{.948} &.028 &\textbf{.941} &.939 &\textbf{.967}&\textbf{.025} &.859 &.800 &\textbf{.896} &.049 &\textbf{.937} &\textbf{.926} &\textbf{.969} &\textbf{.022} &.879 &.858 &.918 &.053  & .862 & \textbf{.860} & \textbf{.898} & \textbf{.061} \\
  & EVAE&\textbf{.911} &\textbf{.884} &.947 &\textbf{.027} &.940 &\textbf{.941} &.965&\textbf{.025} &\textbf{.862} &\textbf{.805} &.892 &\textbf{.046} &.930 &.923 &.963 &.023 &\textbf{.881} &\textbf{.862} &\textbf{.920} &\textbf{.051}  & \textbf{.863} & .857 & .893 & .062 \\
  & EABP&.908 &.875 &.942 &.029 &.935 &.935 &.962&.026 &.858 &.797 &.892 &.051 &.930 &.922 &.964 &.023 &.877 &.855 &.915 &.054  & .860 & \textbf{.860} & \textbf{.898} & \textbf{.061} \\ 
  \bottomrule
  \end{tabular}
  \label{tab:backbone_analysis_rgb_sod}
\end{table*}
\begin{table*}[ht!]
  \centering
  \scriptsize
  \renewcommand{\arraystretch}{1.2}
  \renewcommand{\tabcolsep}{0.5mm}
  \caption{\Rev{Backbone analysis of the proposed EBM prior-based generative saliency detection framework for RGB-D saliency detection, where the bold numbers represent the best performance within each backbone model.}}
  \begin{tabular}{c|c|cccc|cccc|cccc|cccc|cccc|cccc}
\toprule
  &&\multicolumn{4}{c|}{NJU2K~\cite{NJU2000}}&\multicolumn{4}{c|}{SSB~\cite{niu2012leveraging}}&\multicolumn{4}{c|}{DES~\cite{cheng2014depth}}&\multicolumn{4}{c|}{NLPR~\cite{peng2014rgbd}}&\multicolumn{4}{c|}{LFSD \cite{li2014saliency}}&\multicolumn{4}{c}{SIP~\cite{sip_dataset}} \\
    Backbone& Method & $S_{\alpha}\uparrow$&$F_{\beta}\uparrow$&$E_{\xi}\uparrow$&$\mathcal{M}\downarrow$& $S_{\alpha}\uparrow$&$F_{\beta}\uparrow$&$E_{\xi}\uparrow$&$\mathcal{M}\downarrow$& $S_{\alpha}\uparrow$&$F_{\beta}\uparrow$&$E_{\xi}\uparrow$&$\mathcal{M}\downarrow$& $S_{\alpha}\uparrow$&$F_{\beta}\uparrow$&$E_{\xi}\uparrow$&$\mathcal{M}\downarrow$& $S_{\alpha}\uparrow$&$F_{\beta}\uparrow$&$E_{\xi}\uparrow$&$\mathcal{M}\downarrow$& $S_{\alpha}\uparrow$&$F_{\beta}\uparrow$&$E_{\xi}\uparrow$&$\mathcal{M}\downarrow$ \\ \hline
   \multirow{4}{*}{ResNet50} & EGAN&.910 &.898 &\textbf{.939} &\textbf{.031} &.895 &.868 &.928 &.044 &\textbf{.941} &\textbf{.926} &\textbf{.974} &.017 &.909 &.877 &.944 &.028 &.864 &.849 &.885 &.074  & \textbf{.883} & \textbf{.877} & \textbf{.921} & \textbf{.047} \\ 
   & EVAE&.909 &.898 &.938 &.032 &.899 &\textbf{.875} &\textbf{.933} &.042 &.939 &\textbf{.926} &.973 &.017 &.918 &\textbf{.892} &\textbf{.950} &.026 &.868 &.855 &.893 &.071  & .882 & .869 & \textbf{.921} & \textbf{.047} \\
   & EABP&.907 &.895 &.937 & .034 &.897 &.871 &.930 &.043 &.916 &.904 &.946 &.023 &.914 &.884 &.945 &.026 &.867 &.847 &.889 &.073 &.881  & .875 & .919 & .050 \\
   & EABP\_A&\textbf{.911} &\textbf{.900} &\textbf{.939} &.032 &\textbf{.900} &\textbf{.875} &\textbf{.933} &\textbf{.041} &.939 &.925 &.973 &\textbf{.016} &\textbf{.921} &\textbf{.892} &\textbf{.950} &\textbf{.024} &\textbf{.880} &\textbf{.868} &\textbf{.908} &\textbf{.065}  & .879 & .870 & .919 & .049 \\
   \hline
   \multirow{4}{*}{VGG16} & EGAN&.897 &.892 &.930 &.041 &.889 &.870 &.927 &.045 &.911 &.902 &.945 &.024 &.912 &.888 &\textbf{.948} &.027 &.866 &.856 &.891 &.073  & .869 & .865 & .910 & .056 \\ 
   & EVAE&.902 &.893 &.934 &.040 &.886 &.862 &.927 &.046 &\textbf{.923} &\textbf{.912} &\textbf{.958} &\textbf{.021} &.911 &.885 &\textbf{.948} &.028 &\textbf{.871} &\textbf{.861} &\textbf{.897} &\textbf{.067}  & .872 & .860 & .913 & .054 \\
   & EABP&.898 &.886 &.932 &.041 &.890 &.870 &.929 &.044 &.904 &.890 &.941 &.025 &.906 &.876 &.939 &.029 &.869 &.852 &.892 &.073  & .875 & .864 & .916 & .053 \\
   & EABP\_A&\textbf{.906} &\textbf{.898} &\textbf{.935} &\textbf{.038} &\textbf{.900} &\textbf{.885} &\textbf{.939} &\textbf{.040} &.906 &.892 &.946 &.023 &\textbf{.918} &\textbf{.894} &\textbf{.948} &\textbf{.026} &.867 &.856 &.896 &.072  & \textbf{.883} & \textbf{.879} & \textbf{.923} & \textbf{.048} \\
   \hline
   \multirow{3}{*}{Swin} & EGAN&\textbf{.931} &.923 &\textbf{.957} &\textbf{.027} &\textbf{.918} &.896 &\textbf{.954}&\textbf{.032} &\textbf{.948} &\textbf{.930} &\textbf{.977} &\textbf{.016} &.936 &\textbf{.928} &\textbf{.968} &\textbf{.018} &\textbf{.877} &.869 &\textbf{.910} &\textbf{.060}  & \textbf{.906} & .906 & \textbf{.947} & \textbf{.037} \\ 
   & EVAE&.926 &.919 &.955 &.028 &.911 &.892 &.946&.033 &.940 &.929 &.973 &\textbf{.016} &.933 &.914 &.963 &.019 &.875 &\textbf{.870} &.908 &\textbf{.060} & .896 & .901 & .935 & \textbf{.037} \\
   & EABP&.929 &\textbf{.924} &.956 &.028 &.916 &\textbf{.898} &.950 &\textbf{.032}&.945 &.928 &.971 &\textbf{.016} &\textbf{.938} &.921 &.966 &\textbf{.018}&.872 &.862 &.901 &.066  &\textbf{.906} &\textbf{.908} &.940 &\textbf{.037}  \\ 
   \bottomrule
  \end{tabular}
  \label{tab:backbone_analysis_rgbd_sod}
\end{table*}

\begin{figure}[!htp]
   \begin{center}
   \begin{tabular}{c@{ } c@{ } c@{ } c@{ } c@{ }}
   {\includegraphics[width=0.185\linewidth]{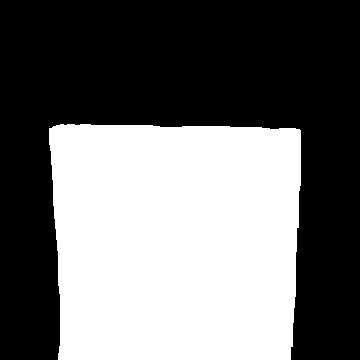}}&
   {\includegraphics[width=0.185\linewidth]{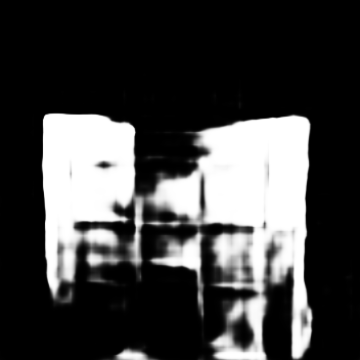}}&
   {\includegraphics[width=0.185\linewidth]{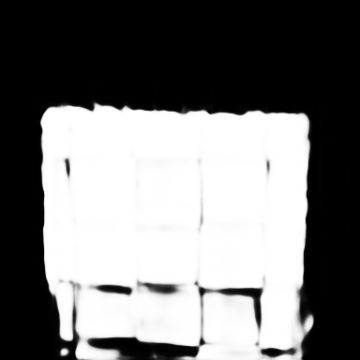}}&
   {\includegraphics[width=0.185\linewidth]{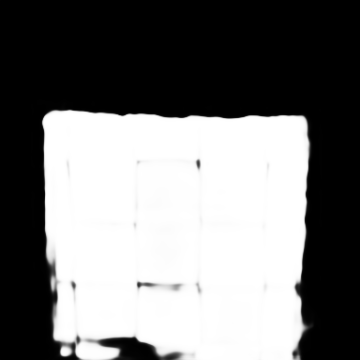}}&
   {\includegraphics[width=0.185\linewidth]{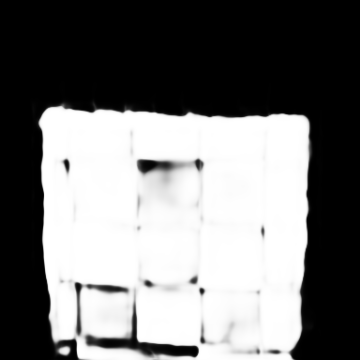}}\\
   {\includegraphics[width=0.185\linewidth]{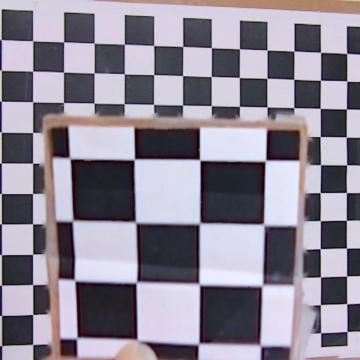}}&
   {\includegraphics[width=0.185\linewidth]{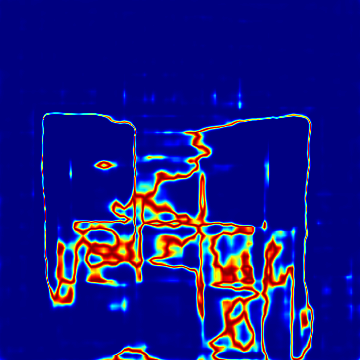}}&
   {\includegraphics[width=0.185\linewidth]{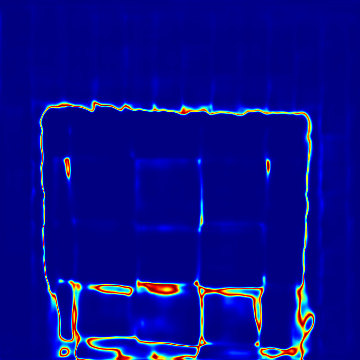}}&
   {\includegraphics[width=0.185\linewidth]{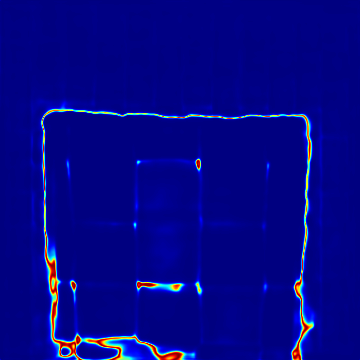}}&
   {\includegraphics[width=0.185\linewidth]{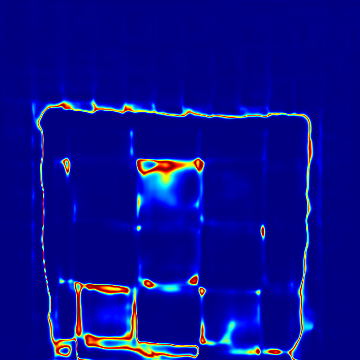}}\\
   {\includegraphics[width=0.185\linewidth]{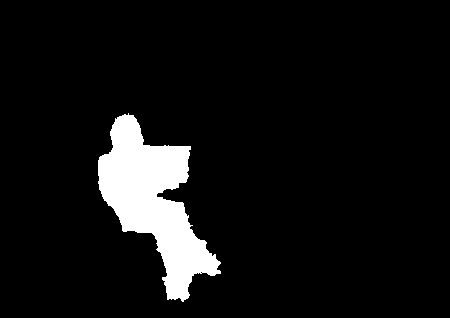}}&
   {\includegraphics[width=0.185\linewidth]{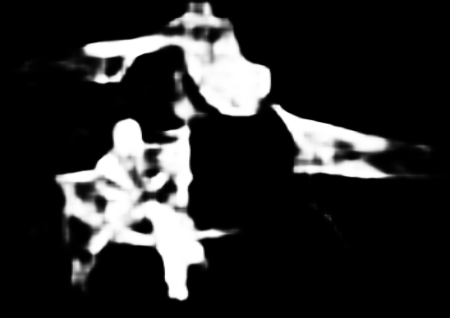}}&
   {\includegraphics[width=0.185\linewidth]{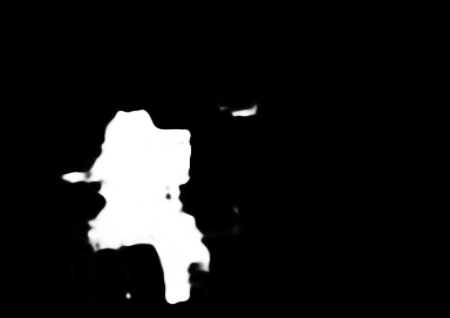}}&
   {\includegraphics[width=0.185\linewidth]{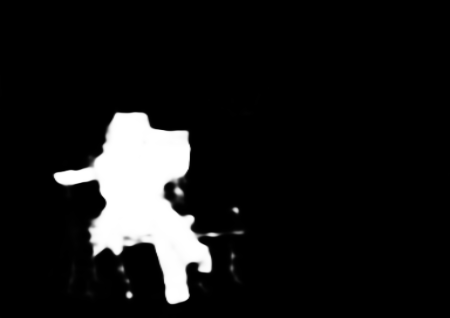}}&
   {\includegraphics[width=0.185\linewidth]{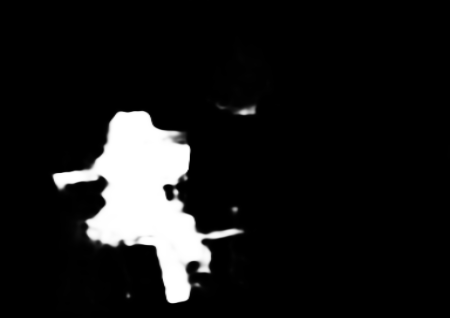}}\\
   {\includegraphics[width=0.185\linewidth]{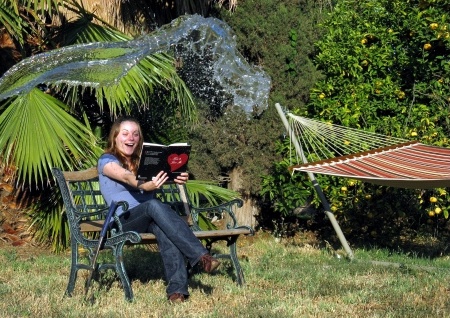}}&
   {\includegraphics[width=0.185\linewidth]{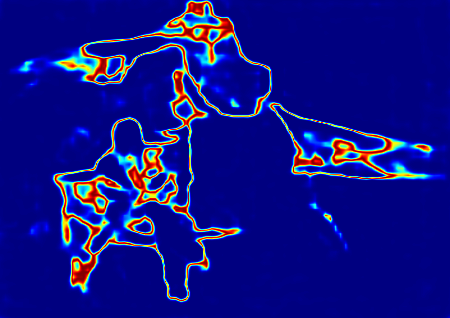}}&
   {\includegraphics[width=0.185\linewidth]{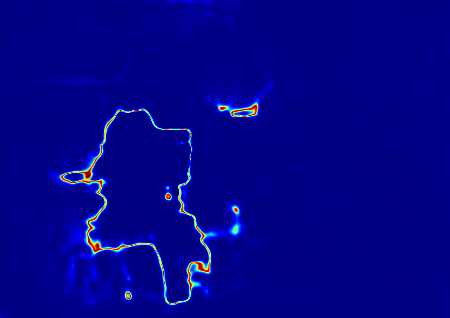}}&
   {\includegraphics[width=0.185\linewidth]{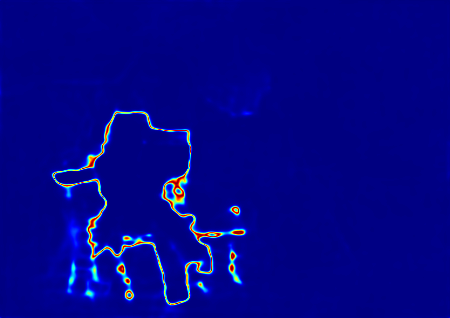}}&
   {\includegraphics[width=0.185\linewidth]{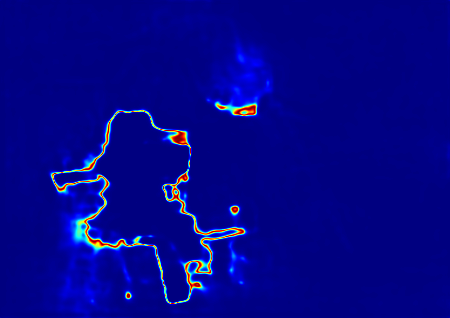}}\\
    \footnotesize{GT/Image}& \footnotesize{UCNet \cite{jing2020uncertainty}}& \footnotesize{EGAN} & \footnotesize{EVAE} & \footnotesize{EABP} \\
   \end{tabular}
   \end{center}
   \caption{Comparison of performance between state-of-the-art generative saliency detection model \cite{ucnet_sal} and the proposed three EBM prior-based generative saliency models, where the first and third rows show the ground truth saliency maps and model predictions, and the second and the fourth rows display the input images and the uncertainty maps.}
\label{fig:uncertainty_comparison_ucnet}
\end{figure}

\subsubsection{Uncertainty Comparison} As a generative learning framework, we further compare with existing generative saliency detection models \cite{ucnet_sal} with respect to the quality of uncertainty maps in Figure~\ref{fig:uncertainty_comparison_ucnet}. UCNet \cite{ucnet_sal} is a conditional variational auto-encoder (CVAE) \cite{structure_output} for RGB-D saliency detection. Different from UCNet \cite{ucnet_sal} which uses a Gaussian latent variable assumption, our three new generative models (i.e., EGAN, EVAE and EABP) use an EBM prior for the latent space.
The inference process of the latent variables is performed using gradient-based MCMC \cite{neal2011mcmc} in EABP, which helps to avoid the posterior collapse issue \cite{Lagging_Inference_Networks} within the conventional CVAE-based framework. Figure~\ref{fig:uncertainty_comparison_ucnet} shows that our three models can obtain more accurate saliency prediction performance and more reliable uncertainty maps than the baseline UCNet. 

\subsection{Backbone Analysis}
\label{sec:backbone_analysis}

As most saliency detection models use CNN backbones, we test our models using two widely used CNN backbones (ResNet50 and VGG16), and show performance
for both RGB and RGB-D saliency detection in Table \ref{tab:backbone_analysis_rgb_sod} and Table \ref{tab:backbone_analysis_rgbd_sod}, respectively.


\begin{table*}[ht!]
  \centering
  \scriptsize
  \renewcommand{\arraystretch}{1.2}
  \renewcommand{\tabcolsep}{0.5mm}
  \caption{\Rev{An ablation study that analyzes the deterministic and stochastic components of our proposed framework for RGB salient object detection.}}
  \begin{tabular}{c|c|cccc|cccc|cccc|cccc|cccc|cccc}
\toprule
  &&\multicolumn{4}{c|}{DUTS~\cite{imagesaliency}}&\multicolumn{4}{c|}{ECSSD~\cite{yan2013hierarchical}}&\multicolumn{4}{c|}{DUT~\cite{Manifold-Ranking:CVPR-2013}}&\multicolumn{4}{c|}{HKU-IS~\cite{li2015visual}}&\multicolumn{4}{c|}{PASCAL-S~\cite{pascal_s_dataset}}& \multicolumn{4}{c}{SOD \cite{sod_dataset}}\\
    Backbone& Method & $S_{\alpha}\uparrow$&$F_{\beta}\uparrow$&$E_{\xi}\uparrow$&$\mathcal{M}\downarrow$& $S_{\alpha}\uparrow$&$F_{\beta}\uparrow$&$E_{\xi}\uparrow$&$\mathcal{M}\downarrow$& $S_{\alpha}\uparrow$&$F_{\beta}\uparrow$&$E_{\xi}\uparrow$&$\mathcal{M}\downarrow$& $S_{\alpha}\uparrow$&$F_{\beta}\uparrow$&$E_{\xi}\uparrow$&$\mathcal{M}\downarrow$& $S_{\alpha}\uparrow$&$F_{\beta}\uparrow$&$E_{\xi}\uparrow$&$\mathcal{M}\downarrow$& $S_{\alpha}\uparrow$&$F_{\beta}\uparrow$&$E_{\xi}\uparrow$&$\mathcal{M}\downarrow$ \\ \hline
   \multirow{4}{*}{ResNet50} & Base&.884 &.839 &\textbf{.921} &.038 &.916 &.911 &\textbf{.943} &\textbf{.038} &.828 &.748 &\textbf{.861} &.058 &.914 &.899 &\textbf{.950} &.031 &.857 &.827 &\textbf{.897} &.065  & \textbf{.839} & \textbf{.831} & \textbf{.878} & \textbf{.071} \\ 
   & GAN&.883 &.835 &.916 &.039 &\textbf{.918} &.912 &\textbf{.943} &\textbf{.038} &.826 &.742 &.855 &.060 &.915 &.900 &.949 &.031 &.857 &.823 &.894 &.066  & .837 & .825 & .875 & .073 \\
   & VAE&\textbf{.885} &\textbf{.846} &.919 &\textbf{.037} &.916 &\textbf{.914} &.942 &\textbf{.038} &\textbf{.830} &\textbf{.754} &\textbf{.861} &\textbf{.054} &\textbf{.916} &\textbf{.906} &\textbf{.950} &\textbf{.030} &\textbf{.859} &\textbf{.832} &\textbf{.897} &\textbf{.064}  & .827 & .816 & .857 & .078 \\
   & ABP&.876 &.817 &.903 &.043 &.915 &.902 &.934 &.042 &.823 &.732 &.845 &.064 &.912 &.890 &.941 &.035 &.857 &.818 &.891 &.068  & .831 & .811 & .855 & .078 \\
   \hline
   \multirow{4}{*}{VGG16} & Base&\textbf{.867} &.824 &\textbf{.909} &\textbf{.041} &.901 &.902 &.931 &.045 &\textbf{.811} &\textbf{.733} &\textbf{.849} &\textbf{.061} &.904 &\textbf{.897} &.944 &\textbf{.033} &.848 &.825 &\textbf{.893} &\textbf{.068}  & .799 & .792 & .837 & .087 \\
   & GAN&\textbf{.867} &\textbf{.825} &.908 &.042 &.900 &\textbf{.903} &.930 &.045 &.810 &.728 &.844 &\textbf{.061} &.905 &\textbf{.897} &.944 &\textbf{.033} &\textbf{.849} &\textbf{.826} &.891 &.069  & .805 & .803 & .842 & .085 \\
   & VAE&.863 &.814 &.903 &.045 &\textbf{.904} &\textbf{.903} &\textbf{.934} &\textbf{.043} &.810 &.726 &.844 &.064 &\textbf{.906} &.895 &\textbf{.945} &\textbf{.033} &.847 &.820 &.890 &.069  & \textbf{.812} & \textbf{.806} & \textbf{.850} & \textbf{.080} \\
   & ABP&.859 &.801 &.886 &.048 &.897 &.891 &.918 &.050 &.803 &.708 &.826 &.066 &.900 &.884 &.929 &.038 &.847 &.816 &.880 &.072  & .806 & .790 & .829 & .088 \\
   \hline
  \multirow{4}{*}{Swin} & Base&.901 &.864 &.938 &.032 &.931 &.930 &.959 &.027 &.854 &.792 &.888 &.053 &.928 &.919 &.964 &.024 &.874 &.848 &.914 &.055  & .861 & .858 & .901 & .061 \\ 
   & GAN&\textbf{.911} &\textbf{.881} &\textbf{.948} &\textbf{.027} &.932 &.933 &.959 &.027 &.858 &.797 &\textbf{.893} &.049 &\textbf{.931} &.923 &\textbf{.965} &\textbf{.023} &.878 &.857 &.918 &.053  & .860 & .858 & .900 & .061 \\
   & VAE&.908 &.879 &.944 &.029 &\textbf{.934} &\textbf{.936} &\textbf{.961} &\textbf{.026} &\textbf{.859} &\textbf{.801} &\textbf{.893} &.050 &\textbf{.931} &\textbf{.925} &\textbf{.965} &\textbf{.023} &.880 &\textbf{.860} &\textbf{.920} &\textbf{.051}  & \textbf{.863} & \textbf{.862} & \textbf{.903} & \textbf{.060} \\
   & ABP&.909 &.870 &.941 &.029 &.931 &.925 &.954 &.030 &.857 &.788 &.885 &\textbf{.048} &.929 &.914 &.960 &.026 &\textbf{.883} &.856 &.919 &.052  & .855 & .846 & .886 & .064  \\ 
   \bottomrule
  \end{tabular}
  \label{tab:ablation_rgb_sod}
\end{table*}


We observe that our generative models using the Swin transformer backbone \cite{liu2021swin} perform better than those using conventional CNN backbones for both RGB and RGB-D saliency detection tasks, which indicates the superiority of the Swin transformer backbone for these tasks. Additionally, we find that the ResNet50 backbone perform slightly better than the VGG16 backbone. During the training of the two CNN backbone-based models, we use the same initial learning rates and maximum number of learning epochs. However, as the ResNet50 backbone converges faster than the VGG16 backbone, the relatively worse performance of the VGG16 backbone-based generative models may be due to less thorough training. We plan to investigate this in future~research.

In our generative models, we do not use widely adopted data augmentation techniques, which are common in existing literature. For a fair comparison, we also introduce multi-scale training to the proposed \enquote{EABP} models to achieve data augmentation, leading to \enquote{EABP\_A} models shown in both Table \ref{tab:backbone_analysis_rgb_sod} and Table \ref{tab:backbone_analysis_rgbd_sod}.  This involves performing training on three scales (i.e., 0.75, 1, 1.25) of the original training images. However, as the Swin transformer backbone \cite{liu2021swin} can only take a fixed input size, we can not perform multi-scale training for the Swin-related generative models. The improved performance of our models with the data augmentation technique highlights the effectiveness of data augmentation for saliency detection models. 

\begin{figure*}[!htp]
   \begin{center}
   \begin{tabular}{c@{ } c@{ } c@{ } c@{ } c@{ } c@{ } c@{ }}
   {\includegraphics[width=0.133\linewidth]{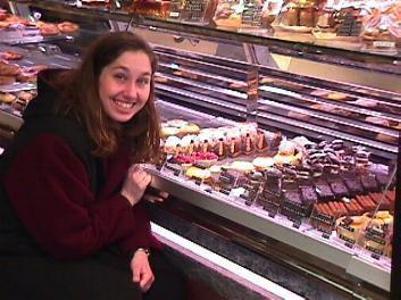}}&
   {\includegraphics[width=0.133\linewidth]{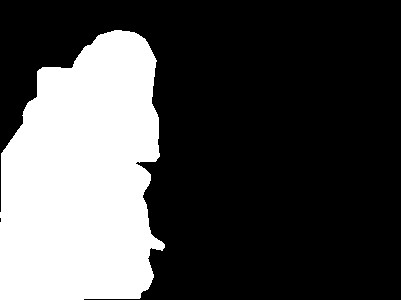}}&
   {\includegraphics[width=0.133\linewidth]{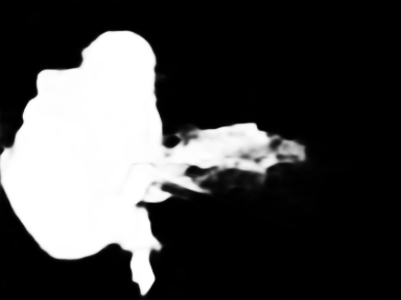}}&
   {\includegraphics[width=0.133\linewidth]{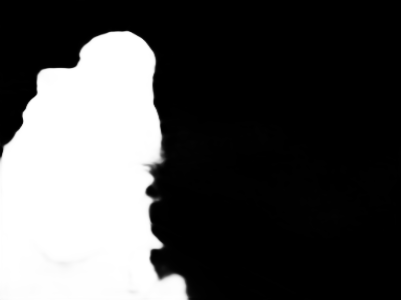}}&
   {\includegraphics[width=0.133\linewidth]{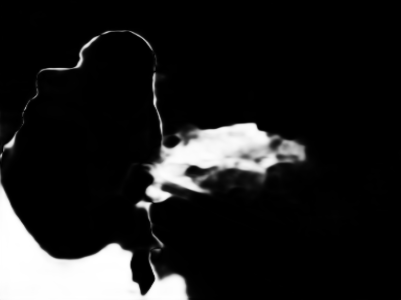}}&
   {\includegraphics[width=0.133\linewidth]{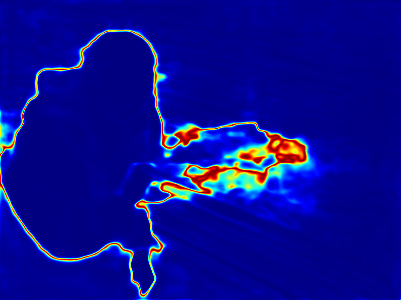}}&
   {\includegraphics[width=0.133\linewidth]{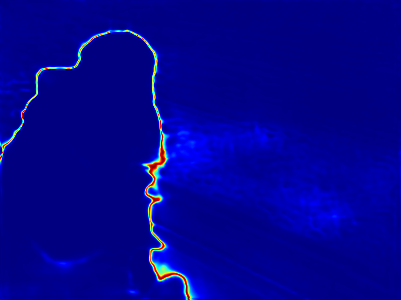}}
   \\
   {\includegraphics[width=0.133\linewidth]{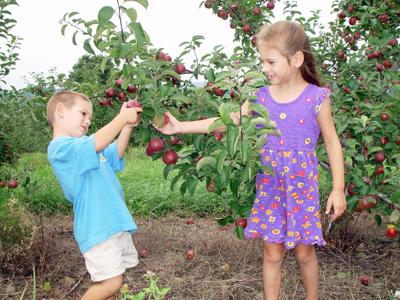}}&
   {\includegraphics[width=0.133\linewidth]{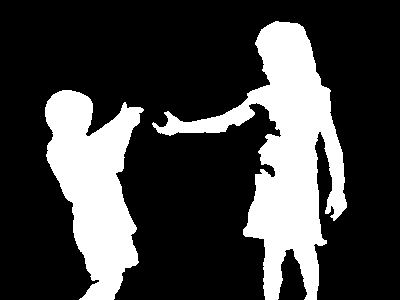}}&
   {\includegraphics[width=0.133\linewidth]{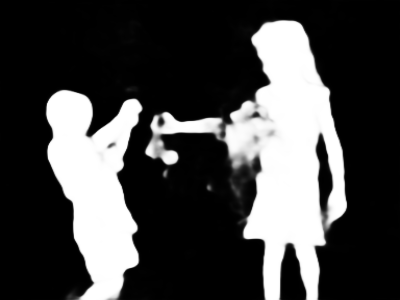}}&
   {\includegraphics[width=0.133\linewidth]{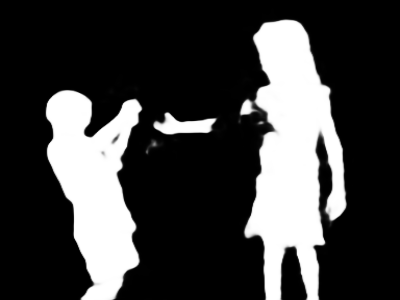}}&
   {\includegraphics[width=0.133\linewidth]{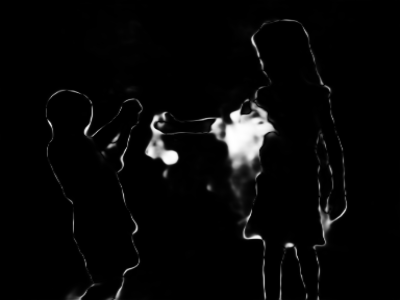}}&
   {\includegraphics[width=0.133\linewidth]{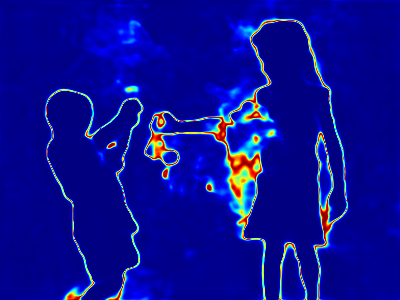}}&
   {\includegraphics[width=0.133\linewidth]{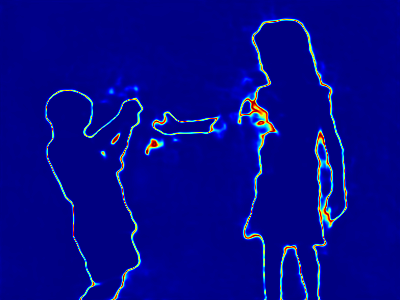}}
   \\
   {\includegraphics[width=0.133\linewidth]{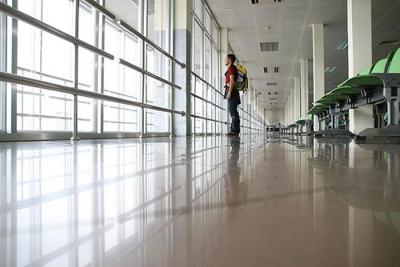}}&
   {\includegraphics[width=0.133\linewidth]{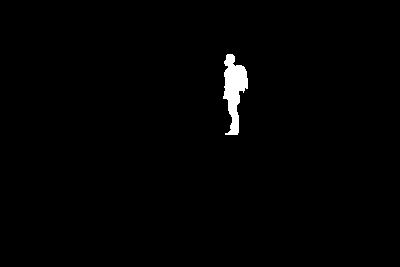}}&
   {\includegraphics[width=0.133\linewidth]{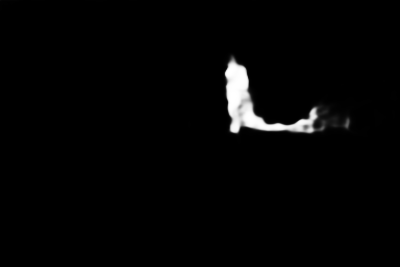}}&
   {\includegraphics[width=0.133\linewidth]{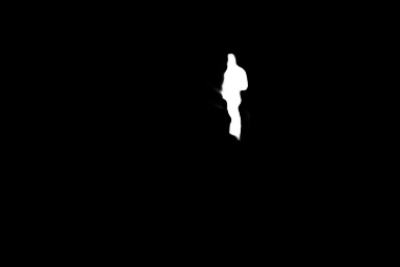}}&
   {\includegraphics[width=0.133\linewidth]{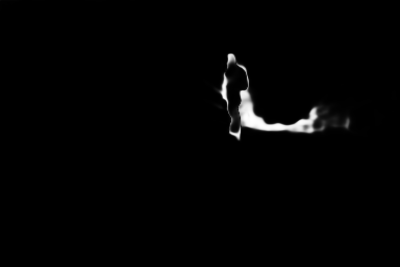}}&
   {\includegraphics[width=0.133\linewidth]{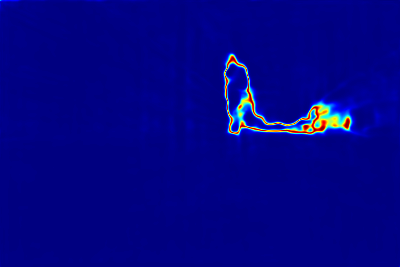}}&
   {\includegraphics[width=0.133\linewidth]{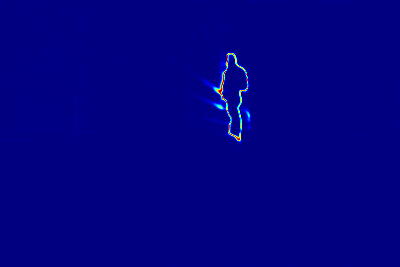}}\\
    \footnotesize{Image}& \footnotesize{GT}
    &\multicolumn{3}{c}{\footnotesize{Saliency Prediction}} & \multicolumn{2}{c}{\footnotesize{Uncertainty Map}} \\
   \end{tabular}
   \end{center}
   \caption{Comparison of saliency predictions and uncertainty maps.
   Within the \enquote{Saliency Prediction} panel, we show the saliency predictions of the ResNet50 backbone, the Swin transformer backbone, and
   their prediction difference.
   We show uncertainty maps of the ResNet50 backbone and the Swin transformer backbone in the \enquote{Uncertainty Map} panel.}
\label{fig:swin_vs_resnet50}
\end{figure*}

Table \ref{tab:backbone_analysis_rgb_sod} and Table \ref{tab:backbone_analysis_rgbd_sod} demonstrate the advantages of using the Swin transformer as a backbone for saliency detection. To further analyze the difference in uncertainty modeling abilities between transformer-based and CNN backbone-based models, we visualize the uncertainty maps and prediction gaps of Swin transformer-based model and ResNet50 backbone-based model in Figure~\ref{fig:swin_vs_resnet50}. We only show the ``EABP'' model of each backbone for simplicity. The qualitative prediction results shown in Figure~\ref{fig:swin_vs_resnet50} demonstrate the superior performance of the transformer backbone-based generative model, which can be attributed to the long-range dependency modeling ability of the vision transformer structure. Moreover, The difference map between the saliency prediction shown in the fifth column reveal the performance gap between the transformer and the CNN backbones, and the uncertainty maps of models based on both backbones are generally reasonable. Thus, we can conclude that although the performance of transformer and CNN backbones differ, their uncertainty modeling abilities within our EBM prior-based generative models are useful and reasonable.


\begin{figure*}[!htp]
   \begin{center}
   \begin{tabular}{c@{ } c@{ } c@{ }}
   {\includegraphics[width=0.320\linewidth]{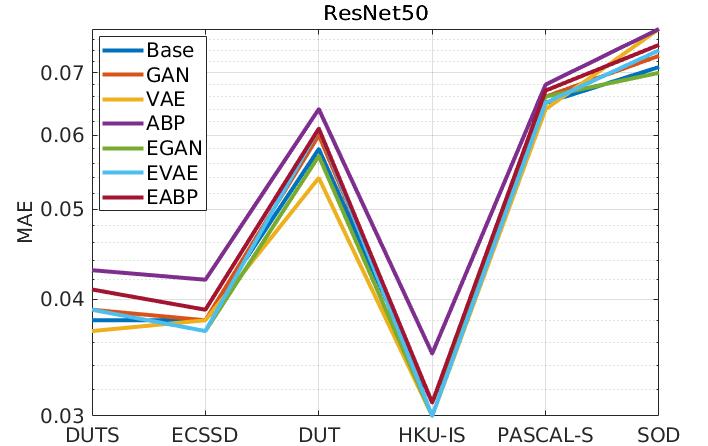}}&
   {\includegraphics[width=0.320\linewidth]{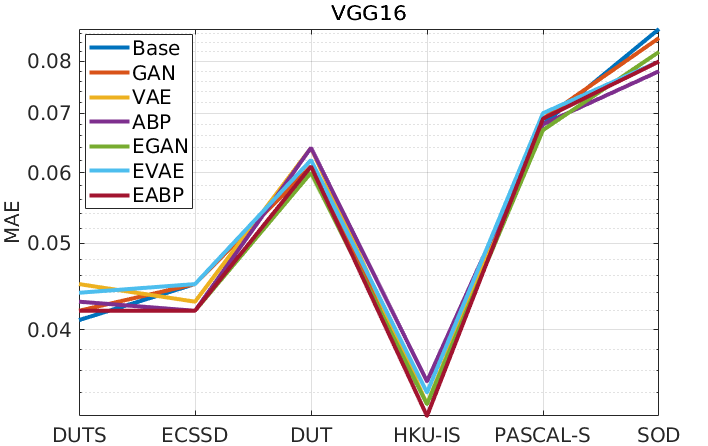}}&
   {\includegraphics[width=0.320\linewidth]{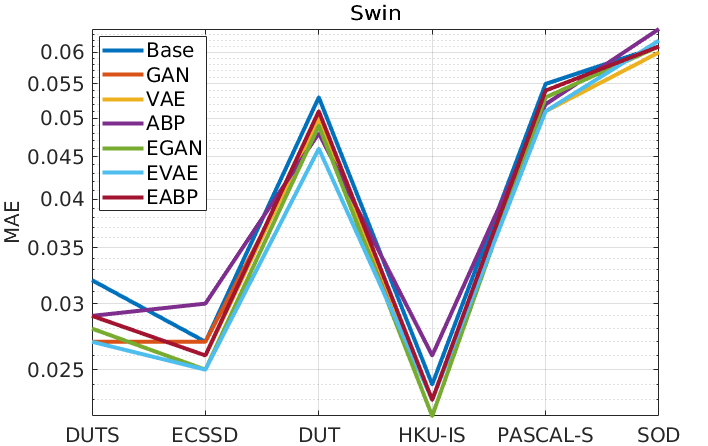}} \\
   \end{tabular}
   \end{center}
   \caption{\Rev{Comparison of RGB saliency detection performance (MAE) among the base models (\enquote{Base}),  the conventional generative models (\enquote{GAN}, \enquote{VAE} and \enquote{ABP}) and the proposed EBM prior-based generative models (\enquote{EGAN}, \enquote{EVAE} and \enquote{EABP}).}}
\label{fig:deter_per_conv_gen_ebm_gen}
\end{figure*}

\subsection{Ablation Study}
We've conducted additional experiments.
Unless otherwise specified, the experiments
pertain to RGB saliency detection.

\subsubsection{Analyzing Deterministic Component} We first investigate the deterministic component of our framework. We first design a deterministic base model for saliency detection by feeding the backbone features directly to the decoder to generate a one-channel saliency map. We design three base models with ResNet50, VGG16 and Swin transformer backbones respectively. The performance of these models is presented as \enquote{Base} in each \enquote{Backbone} category of Table \ref{tab:ablation_rgb_sod}, which clearly demonstrates the advantage of using transformer backbones for saliency detection. The superior performance of the Swin transformer backbone \cite{liu2021swin} compared to ResNet50 and VGG16 can be attributed to its massive number of parameters. Specifically, the Swin transformer contains 90M parameters for the \enquote{Base} model, which is almost double the number of parameters of ResNet50 (52M) and triple the number of parameters of VGG16 (29M). The increased number of parameters allows the Swin transformer to capture more complex and high-level features, resulting in improved performance.

\subsubsection{Analyzing Stochastic Component} Building upon the deterministic models, we introduce Gaussian latent variables and design the corresponding stochastic models, namely GAN \cite{gan_raw}, VAE \cite{vae_bayes_kumar}, and ABP \cite{abp}. Their performance is indicated as \enquote{GAN}, \enquote{VAE} and \enquote{ABP}, respectively, in Table \ref{tab:ablation_rgb_sod}. For stochastic models using VGG16 and ResNet50 backbones, we observe similar or inferior performance compared to the base models. However, for the Swin transformer-based models, we find comparable or improved performance of the stochastic models (especially \enquote{GAN} and \enquote{VAE}) compared to the base model. We argue that the reason for this may be the different capacity of each backbone model. The latent variables in the larger network can be more expressive to capture the underlying randomness of the saliency maps. In contrast, for a smaller network, the latent variables are more sensitive to the output, and the model requires more careful tuning to achieve a good performance. In this study, we use the same hyper-parameters for the three generative models with different backbones, including the number of dimensions of the latent space and the MCMC-related hyper-parameters. It is worth nothing that more effective models can be obtained with more careful tuning.

\begin{table*}[ht!]
  \centering
  \scriptsize
  \renewcommand{\arraystretch}{1.2}
  \renewcommand{\tabcolsep}{0.5mm}
  \caption{Applying the proposed generative frameworks (including EGAN, EVAE, and EABP) to state-of-the-art deterministic VST models for RGB and RGB-D salient object detection.}
  \begin{tabular}{c|c|cccc|cccc|cccc|cccc|cccc|cccc}
\toprule
  &&\multicolumn{4}{c|}{DUTS~\cite{imagesaliency}}&\multicolumn{4}{c|}{ECSSD~\cite{yan2013hierarchical}}&\multicolumn{4}{c|}{DUT~\cite{Manifold-Ranking:CVPR-2013}}&\multicolumn{4}{c|}{HKU-IS~\cite{li2015visual}}&\multicolumn{4}{c|}{PASCAL-S~\cite{pascal_s_dataset}}& \multicolumn{4}{c}{SOD \cite{sod_dataset}}\\
    Task& Method & $S_{\alpha}\uparrow$&$F_{\beta}\uparrow$&$E_{\xi}\uparrow$&$\mathcal{M}\downarrow$& $S_{\alpha}\uparrow$&$F_{\beta}\uparrow$&$E_{\xi}\uparrow$&$\mathcal{M}\downarrow$& $S_{\alpha}\uparrow$&$F_{\beta}\uparrow$&$E_{\xi}\uparrow$&$\mathcal{M}\downarrow$& $S_{\alpha}\uparrow$&$F_{\beta}\uparrow$&$E_{\xi}\uparrow$&$\mathcal{M}\downarrow$& $S_{\alpha}\uparrow$&$F_{\beta}\uparrow$&$E_{\xi}\uparrow$&$\mathcal{M}\downarrow$& $S_{\alpha}\uparrow$&$F_{\beta}\uparrow$&$E_{\xi}\uparrow$&$\mathcal{M}\downarrow$ \\ \hline
   \multirow{4}{*}{RGB} & VST \cite{Liu_2021_ICCV_VST}&\textbf{.896} &.842 &\textbf{.918} &\textbf{.037} &.932 &\textbf{.911} &.943 &\textbf{.034} &\textbf{.850} &\textbf{.771} &\textbf{.869} &\textbf{.058} &\textbf{.928} &\textbf{.903} &\textbf{.950} &\textbf{.030} &\textbf{.873} &.832 &\textbf{.900} &.067  & \textbf{.851} & \textbf{.833} & \textbf{.879} & \textbf{.069} \\
   & EGAN&.895 &\textbf{.843} &.916 &.038 &.928 &.908 &\textbf{.947} &.036 &\textbf{.850} &.767 &.864 &.060 &.923 &.893 &.946 &.031 &.870 &\textbf{.834} &.896 &.068  & .846 & .827 & \textbf{.879} & .070 \\
   & EVAE&.893 &.841 &.914 &.039 &.931 &.907 &.942 &.037 &.844 &.770 &.862 &.060 &.919 &.894 &.945 &.032 &.869 &.832 &.897 &.067  & .847 & .832 & .872 & .070 \\
   & EABP&.895 &.838 &.916 &.038 &\textbf{.935} &.909 &.944 &.035 &.847 &.770 &.868 &.059 &.923 &.899 &.944 &.031 &.869 &.827 &.895 &\textbf{.063}  & .847 & .829 & .876 & .070 \\
   \hline
   &&\multicolumn{4}{c|}{NJU2K~\cite{NJU2000}}&\multicolumn{4}{c|}{SSB~\cite{niu2012leveraging}}&\multicolumn{4}{c|}{DES~\cite{cheng2014depth}}&\multicolumn{4}{c|}{NLPR~\cite{peng2014rgbd}}&\multicolumn{4}{c|}{LFSD \cite{li2014saliency}}&\multicolumn{4}{c}{SIP~\cite{sip_dataset}} \\ \hline
   \multirow{4}{*}{RGB-D} & VST~\cite{Liu_2021_ICCV_VST} &\textbf{.922} &.898 &.939 &.035 &\textbf{.913} &\textbf{.879} &\textbf{.937} &\textbf{.038} &.943 &.920 &.965 &.017 &\textbf{.932} &\textbf{.897} &\textbf{.951} &\textbf{.024} &.890 &.871 &\textbf{.917} &\textbf{.054}  &.904 &\textbf{.894} &.933 &.040 \\
   & EGAN&\textbf{.922} &\textbf{.901} &\textbf{.942} &\textbf{.034} &.906 &.870 &.932 &.039 &\textbf{.945} &\textbf{.922} &\textbf{.969} &\textbf{.016} &.930 &.896 &\textbf{.951} &.025 &.882 &.864 &.908 &.057  & .905 & .891 & .936 & \textbf{.038} \\
   & EVAE&.919 &.892 &.934 &.036 &.908 &.870 &.929 &.039 &.942 &.920 &.966 &\textbf{.016} &.928 &.887 &.945 &.026 &.885 &.865 &.911 &.058  & .902 & .886 & .931 & .040 \\
   & EABP&\textbf{.922} &.897 &.939 &.036 &.908 &.869 &.929 &.039 &.945 &.915 &.962 &.018 &.926 &.896 &\textbf{.945} &.025 &\textbf{.898} &\textbf{.875} &.916 &.059  & \textbf{.907} & \textbf{.894} & \textbf{.937} & .039 \\ 
\bottomrule
  \end{tabular}
  \label{tab:applying_generative_sod}
\end{table*}

\subsubsection{Deterministic v.s. Stochastic using Gaussian Prior v.s. Stochastic using EBM Prior}
\Rev{We compare deterministic models (\enquote{Base}), stochastic models using a Gaussian prior (\enquote{GAN}, \enquote{VAE}, and \enquote{ABP}), and stochastic models using an EBM prior (\enquote{EGAN}, \enquote{EVAE}, and \enquote{EABP}) in terms of MAE, and show the results in Figure~\ref{fig:deter_per_conv_gen_ebm_gen}. We observe similar performance of the models on easier testing datasets, such as ECSSD \cite{yan2013hierarchical} and HKU-IS \cite{li2015visual}. However, for harder datasets, such as DUT \cite{imagesaliency} and SOD \cite{sod_dataset}, we observe obvious differences in terms of MAE among the models. This is consistent with the fact that observers tend to generate relatively consistent salient objects for easier samples, resulting in lower prediction variance. However, complex images have a higher degree of inherent uncertainty, thus leading to various uncertainty results generated by different generative models.} To qualitatively demonstrate the uncertainty modeling abilities of the those generative models, we generate uncertainty maps of both conventional generative models and our EBM prior-based generative models with the Swin transformer backbone, which are shown in Figure~\ref{fig:pred_visualization}. The uncertainty maps are obtained by computing the variance of multiple stochastic prediction outputs, which visualize the pixel-wise confidence of the model for saliency prediction. 


\subsubsection{Applying the Proposed Stochastic Framework to SOTA Saliency Detection Model} 
\begin{figure*}[!htp]
   \begin{center}
   \begin{tabular}{c@{ } c@{ } c@{ } c@{ } c@{ } c@{ } c@{ } c@{ } c@{ }}
   {\includegraphics[width=0.105\linewidth]{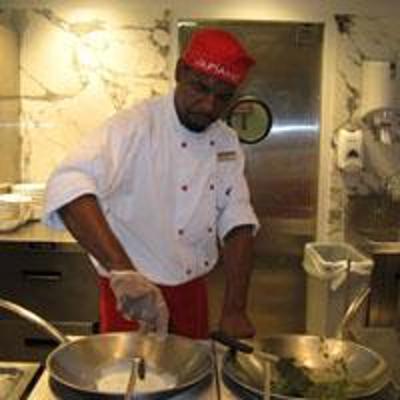}}&
   {\includegraphics[width=0.105\linewidth]{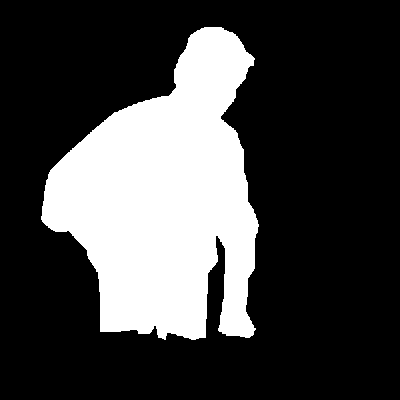}}&
   {\includegraphics[width=0.105\linewidth]{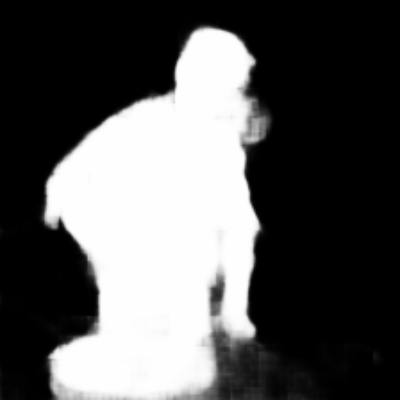}}&
   {\includegraphics[width=0.105\linewidth]{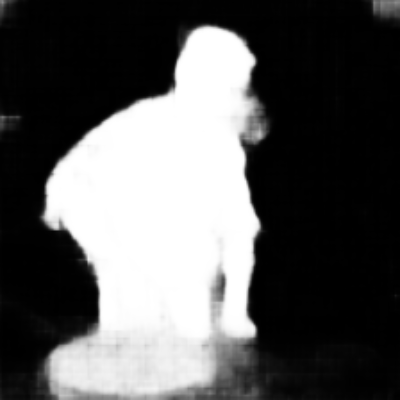}}&
   {\includegraphics[width=0.105\linewidth]{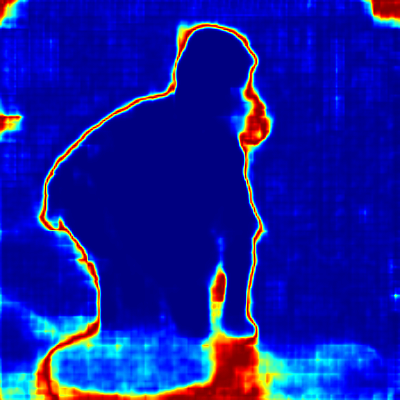}}&
   {\includegraphics[width=0.105\linewidth]{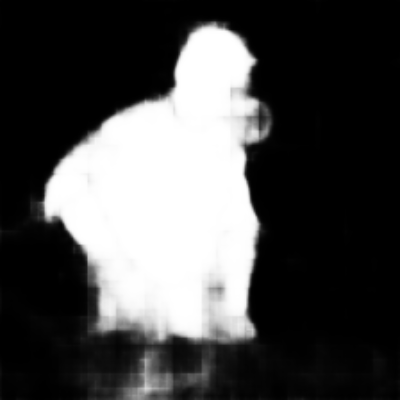}}&
   {\includegraphics[width=0.105\linewidth]{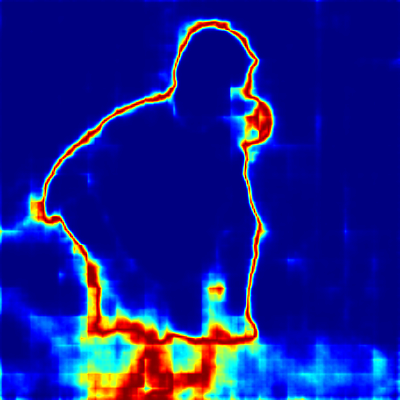}}&
   {\includegraphics[width=0.105\linewidth]{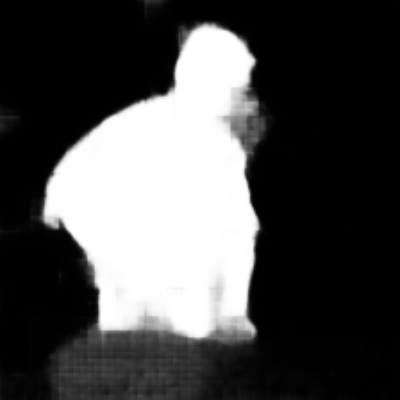}}&
   {\includegraphics[width=0.105\linewidth]{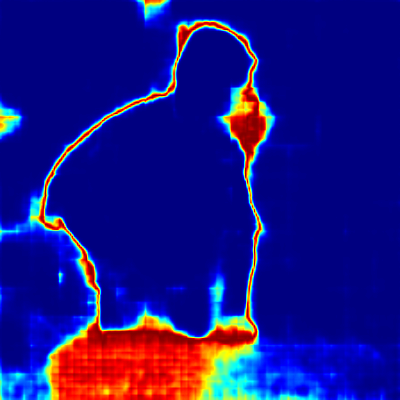}}\\
   {\includegraphics[width=0.105\linewidth]{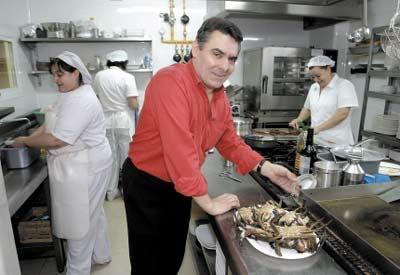}}&
   {\includegraphics[width=0.105\linewidth]{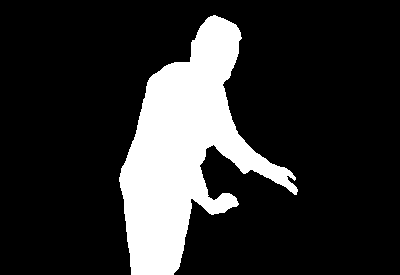}}&
   {\includegraphics[width=0.105\linewidth]{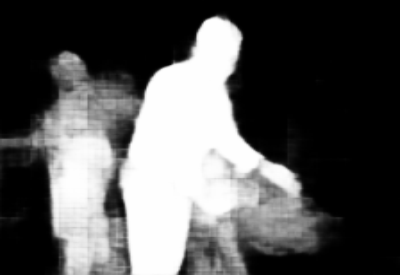}}&
   {\includegraphics[width=0.105\linewidth]{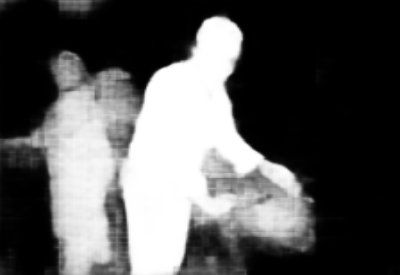}}&
   {\includegraphics[width=0.105\linewidth]{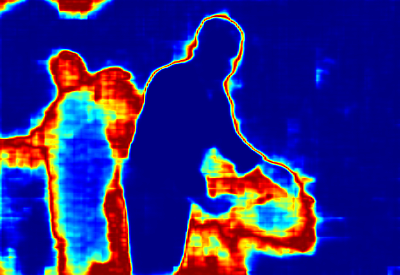}}&
   {\includegraphics[width=0.105\linewidth]{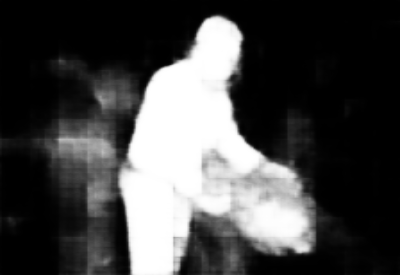}}&
   {\includegraphics[width=0.105\linewidth]{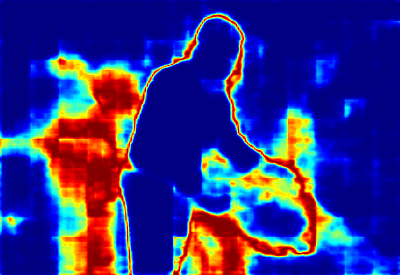}}&
   {\includegraphics[width=0.105\linewidth]{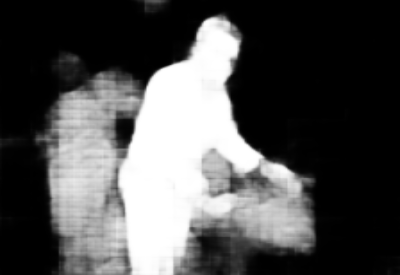}}&
   {\includegraphics[width=0.105\linewidth]{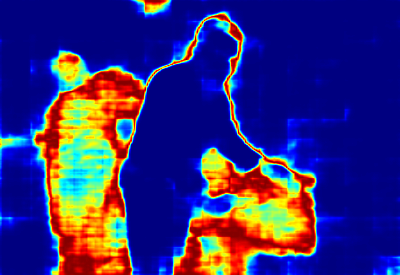}}\\
   {\includegraphics[width=0.105\linewidth]{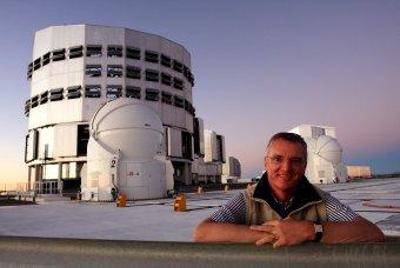}}&
   {\includegraphics[width=0.105\linewidth]{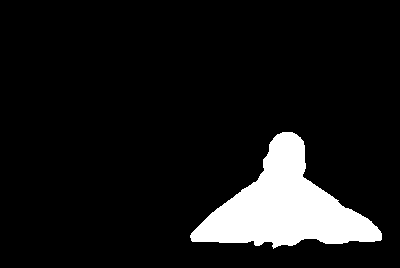}}&
   {\includegraphics[width=0.105\linewidth]{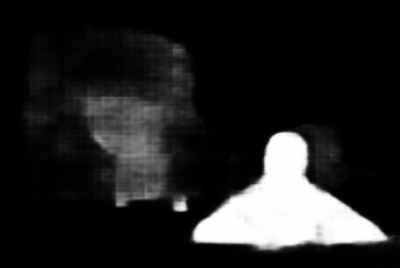}}&
   {\includegraphics[width=0.105\linewidth]{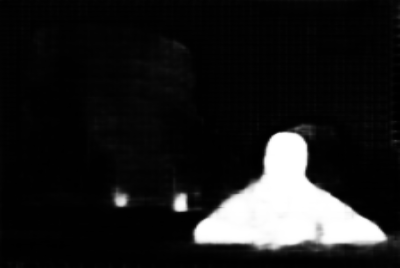}}&
   {\includegraphics[width=0.105\linewidth]{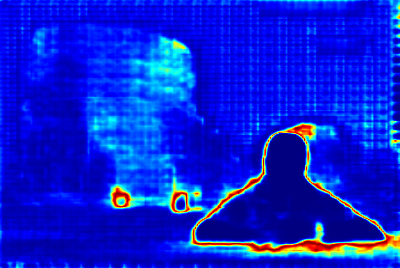}}&
   {\includegraphics[width=0.105\linewidth]{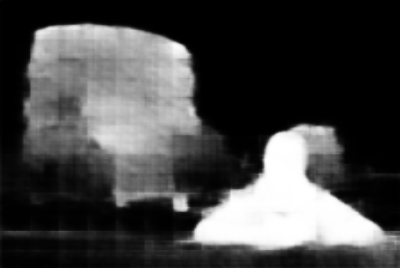}}&
   {\includegraphics[width=0.105\linewidth]{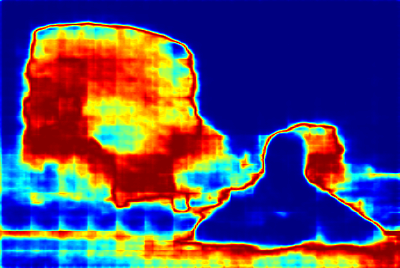}}&
   {\includegraphics[width=0.105\linewidth]{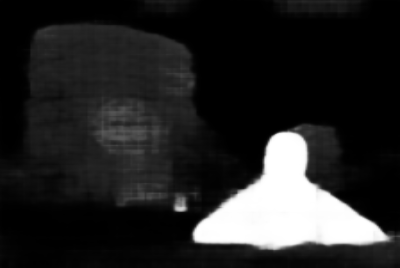}}&
   {\includegraphics[width=0.105\linewidth]{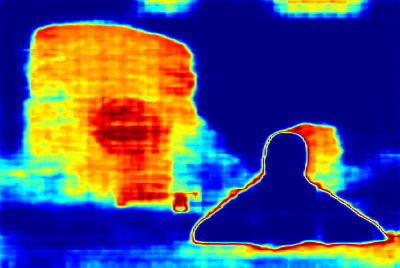}}\\
    \footnotesize{Image}& \footnotesize{GT}& \footnotesize{VST \cite{Liu_2021_ICCV_VST}}
    &\multicolumn{2}{c}{\footnotesize{EGAN}} & \multicolumn{2}{c}{\footnotesize{EVAE}} & \multicolumn{2}{c}{\footnotesize{EABP}} \\
   \end{tabular}
   \end{center}
   \caption{Visual comparison of VST model \cite{Liu_2021_ICCV_VST} and its generative variants, including EGAN, EVAE, and EABP. Each row presents one example of salient object detection. For each example, the first column shows a testing image and the second column shows the ground truth saliency map. The 3rd, 4th, 6th, and 8th columns, respectively, display the prediction outputs of VST, EGAN, EVAE and EABP. For each prediction output of the generative models, we also display the corresponding uncertainty map.}
\label{fig:visual_comparison_applying_sod}
\end{figure*}

As general generative models, we apply our proposed EBM prior-based generative frameworks to the existing SOTA RGB and RGB-D saliency detection models. Especially, we choose VST \cite{Liu_2021_ICCV_VST} as the base model as it provides models for both RGB and RGB-D saliency detection tasks. The performance is shown in Table \ref{tab:applying_generative_sod}, where we also include performance of the original deterministic VST model \cite{Liu_2021_ICCV_VST} for reference. Specifically, the generative VST models
are achieved by following the same manner as our generative models presented in Table \ref{tab:backbone_analysis_rgb_sod} and Table \ref{tab:backbone_analysis_rgbd_sod}. 
Table \ref{tab:applying_generative_sod} shows similar deterministic performance of the EBM prior-based generative VST \cite{Liu_2021_ICCV_VST} compared with the deterministic VST \cite{Liu_2021_ICCV_VST}, where the deterministic performance is achieved by sampling ten iterations from the EBM prior model, and performance of the mean predictions are reported. Table \ref{tab:applying_generative_sod} indicates that the EBM prior-based generative VST has comparable deterministic performance with the deterministic VST \cite{Liu_2021_ICCV_VST}. The deterministic performance is obtained by sampling ten latent vectors from the EBM prior model, and then computing the average performance over the ten trials of predictions. We further visualize the generated uncertainty maps of the stochastic RGB saliency detection models in Figure~\ref{fig:visual_comparison_applying_sod}. \Rev{The visualizations demonstrate that while our generative VST saliency models may not significantly improve the VST performance, they are able to generate uncertainty maps that help explain the prediction behavior of the models and reveal the stochastic nature of salient object prediction in humans.}

\subsubsection{Hyperparameter Analysis} The main hyperparameters in our framework include
the number of Langevin steps $K$, the Langevin step size $\delta$, the number of dimensions of the latent space $d$. We have two sets of hyperparameters $\{\delta^-, K^-\}$ and $\{\delta^+, K^+\}$ of the Langevin dynamics for sampling from the prior distribution and the posterior distribution, respectively.  
As to the Langevin step size, we find stable model performance with $\delta^{-}\in [0.2,0.6]$ and $\delta^{+}\in [0.05,0.3]$, and we set $\delta^{-}=0.4$ and $\delta^{+}=0.1$ in our paper. For the number of Langevin steps, we empirically set $K^-=K^+=5$ to achieve a trade-off between the training efficiency and the model performance, \Rev{as more Langevin steps will lead to longer training time but better convergence.}
Additionally, we investigate the influence of the number of latent dimensions by varying  $d=\{8,16,32,64\}$, and observe comparable performance among different choices of $d$. We set $d=32$ in our paper.

\subsubsection{Analyzing Model Performance over Percentage of Salient Foreground} 
We further analyze model performance with respect to the percentage of the salient foreground of the whole image, which can be one of the main factors that lead to various performance of different saliency detection models. We consider ReNet50, VGG16 and Swin backbones. We first select one backbone as a baseline method, and then compute the S-measure performance \cite{fan2017structure} and the salient foreground percentage for each image on the DUT testing dataset~\cite{Manifold-Ranking:CVPR-2013}. Then we keep those image examples with S-measure scores less than 0.5, indicating failure cases. We test other two backbones on these failure cases and plot curves of the S-measure performance over percentage of salient foreground for all three backbones. 

\begin{figure}[!htp]
   \begin{center}
   \begin{tabular}{c@{ }}
   {\includegraphics[width=0.96\linewidth]{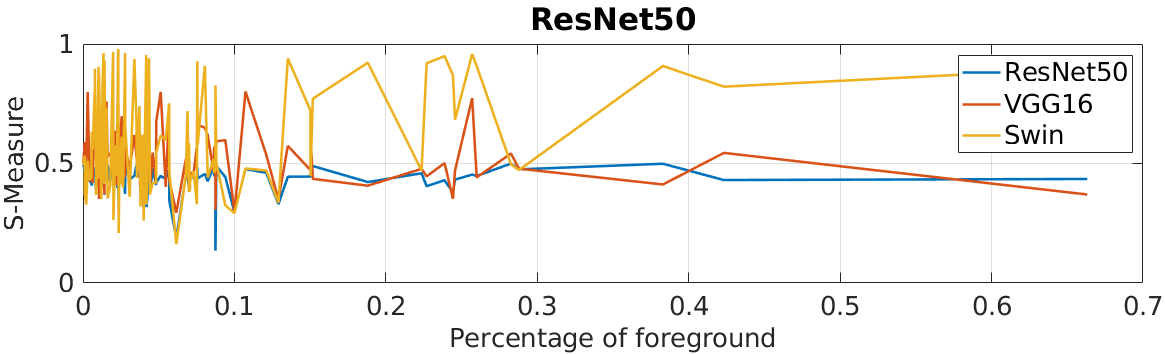}}\\
   {\includegraphics[width=0.96\linewidth]{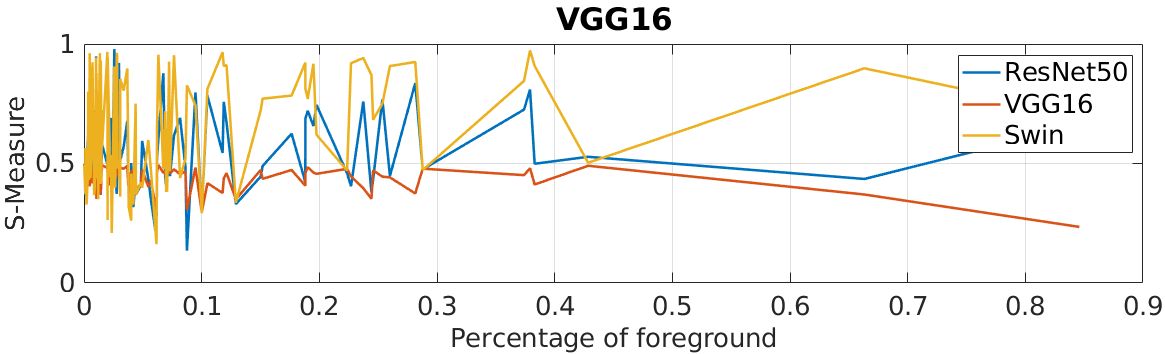}}\\
   {\includegraphics[width=0.96\linewidth]{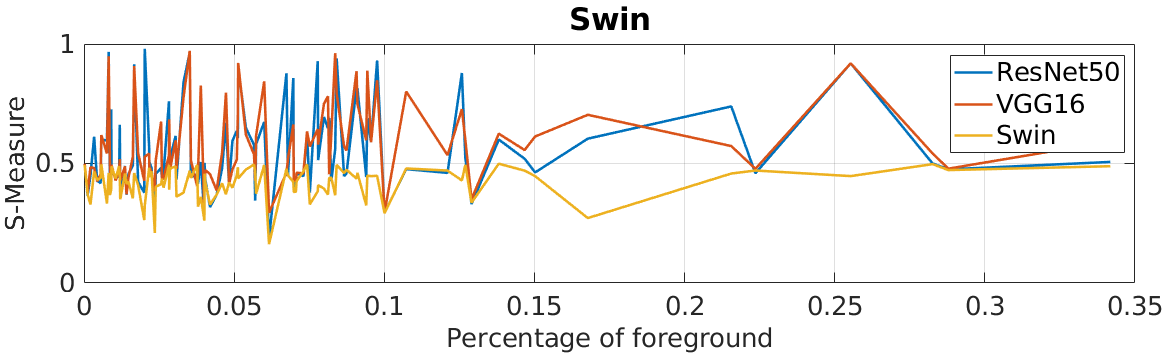}}\\
   \end{tabular}
   \end{center}
   \caption{Model performance \wrt percentage of the salient foreground.}
\label{fig:model_performance_to_percentage}
\end{figure}

In Figure~\ref{fig:model_performance_to_percentage}, each sub-figure corresponds to one base backbone we select and displays S-measure curves over the percentage of salient foreground for all the three backbones. For example, in the \enquote{ResNet50} sub-figure, we use testing examples where the S-measure performance scores of the \enquote{ResNet50} base model is less than 0.5, and plot the S-measure curves for all backbone candidates by using those selected samples. The \enquote{ResNet50} sub-figure shows how other backbones (i.e., \enquote{VGG16} and \enquote{Swin transformer}) perform on the failure examples of the \enquote{ResNet50}. 

Figure~\ref{fig:model_performance_to_percentage} shows that when the performance of the two CNN backbone-based models are unsatisfactory, the Swin transformer backbone-based model can still perform consistently well across the different foreground scales, which can be observed in the top and the middle sub-figures of Figure~\ref{fig:model_performance_to_percentage}. We can also observe CNN backbone based-models perform well on examples on which the Swin transformer backbone performs poorly, thus motivating us to combine both CNN backbone and Swin transformer backbone in the future
to effectively explore their individual advantages. 

It is worth noting that the range of foreground scales for the failure cases of the two CNN backbone-based models is larger than that of the Swin transformer backbone-based model. Specifically, the foreground scale ranges are [0, 0.7] for ResNet50, [0, 0.9] for VGG16, and [0, 0.33] for Swin transformer. This suggests that the Swin transformer-based model is more capable of handling smaller salient objects than the other two CNN backbone-based models, which is consistent with the Swin transformer's powerful long-range dependency modeling ability. However, we also observe that all three backbones still struggle with images containing tiny salient objects, as their performance in the range of [0, 0.1] is highly unstable. Our future work will be focused on addressing the challenging task of detecting tiny salient objects.

\section{Conclusion and Discussion}

In this paper, we study  generative modeling and learning of RGB and RGB-D salient object detection. We start by defining a conditional probability distribution of saliency maps given an input image by a top-down latent variable generative framework, in which the non-linear mapping from image domain to saliency domain is parameterized by a  vision transformer 
network and the prior distribution of the low-dimensional latent space is represented by a trainable energy-based model. Instead of using amortized inference and sampling strategies, we learn the model by the MCMC-based maximum likelihood, where the Langevin sampling is used to evaluate the intractable posterior and prior distributions of the latent variables for calculating the learning gradients of the model parameters. With the informative energy-based prior and the expressive top-down vision transformer network, our model can achieve both accurate predictions and meaningful uncertainty maps that are consistent with the human perception. 
We further propose two variants of the EBM prior-based generative framework: VAE-based and GAN-based generative saliency models using EBM as a prior. To extensively analyze the superiority of the proposed generative framework, we conduct a series of experiments for an ablation study. We also apply our frameworks to existing SOTA deterministic saliency detection models to further verify the generalization ability of our frameworks.

From the machine learning perspective, our model is a top-down deep conditional generative model using EBM as a prior.
The MLE learning algorithm derived from the proposed model is based on MCMC inference for the posterior and MCMC sampling for the prior, which makes our framework more natural, principled, and statistically rigorous than others. On the other hand, the traditional adversarial learning and  variational learning techniques can be easily adapted to the proposed EBM prior-based generative model, which can bring in benefits of good performance and fast inference respectively.   
The proposed frameworks are not only beneficial for saliency prediction but can also be applied to other conditional learning scenarios such as semantic segmentation, image-to-image translation, and more. \Rev{For instance, we can utilize the same modeling approach to construct a conditional distribution of the target image given the source image for stochastic image-to-image translation or a conditional distribution of semantic segmentation map given the RGB image for generative semantic segmentation. The only variation in these generalizations lies in the network architecture design to represent various modalities. Therefore, the proposed generative model and learning algorithm are generic and versatile.}

From the computer vision perspective, our model with a special network design to handle saliency prediction is a new member of the family of saliency prediction methods. In comparison with the traditional discriminative saliency prediction methods, our generative method is natural because it models the saliency prediction as a conditional probability distribution. Further, the EBM prior leads to more reliable uncertainty modeling compared with the conventional generative models.

\Rev{The use of the latent space energy-based prior model poses a challenging problem for MCMC sampling from an unnormalized probability distribution. Our model requires an iterative MCMC to evaluate the EBM, but MCMC via Langevin dynamics is time-consuming. To strike a balance between accurate evaluation and efficient sampling, we may use short-run MCMC chains in practice, which may lead to non-convergent MCMC sampling issues. Therefore, further investigation is expected to understand the learning behaviors of our proposed algorithms in the context of non-convergent short-run MCMC sampling.}

\bibliographystyle{ieeetr}
\bibliography{SOD_Reference}

\begin{thebibliography}{100}

\bibitem{jing2021_nips}
J.~Zhang, N.~B. Jianwen~Xie, and P.~Li, ``Learning generative vision
  transformer with energy-based latent space for saliency prediction,'' in {\em
  Conference on Neural Information Processing Systems (NeurIPS)}, 2021.

\bibitem{wei2020f3net}
J.~Wei, S.~Wang, and Q.~Huang, ``F$^3$net: Fusion, feedback and focus for
  salient object detection,'' in {\em AAAI Conference on Artificial
  Intelligence (AAAI)}, pp.~12321--12328, 2020.

\bibitem{wei2020label}
J.~Wei, S.~Wang, Z.~Wu, C.~Su, Q.~Huang, and Q.~Tian, ``Label decoupling
  framework for salient object detection,'' in {\em IEEE Conference on Computer
  Vision and Pattern Recognition (CVPR)}, pp.~13025--13034, 2020.

\bibitem{fan2020bbs}
D.-P. Fan, Y.~Zhai, A.~Borji, J.~Yang, and L.~Shao, ``{BBS-Net}: {RGB-D}
  salient object detection with a bifurcated backbone strategy network,'' in
  {\em European Conference on Computer Vision (ECCV)}, pp.~275--292, 2020.

\bibitem{Fu2020JLDCF}
K.~Fu, D.-P. Fan, G.-P. Ji, and Q.~Zhao, ``{JL-DCF}: Joint learning and
  densely-cooperative fusion framework for {RGB-D} salient object detection,''
  in {\em IEEE Conference on Computer Vision and Pattern Recognition (CVPR)},
  pp.~3052--3062, 2020.

\bibitem{chen2018progressively}
H.~Chen and Y.~Li, ``Progressively complementarity-aware fusion network for
  {RGB-D} salient object detection,'' in {\em IEEE Conference on Computer
  Vision and Pattern Recognition (CVPR)}, pp.~3051--3060, 2018.

\bibitem{jing2020weakly}
J.~Zhang, X.~Yu, A.~Li, P.~Song, B.~Liu, and Y.~Dai, ``Weakly-supervised
  salient object detection via scribble annotations,'' in {\em IEEE Conference
  on Computer Vision and Pattern Recognition (CVPR)}, pp.~12543--12552, 2020.

\bibitem{scrn_sal}
Z.~Wu, L.~Su, and Q.~Huang, ``Stacked cross refinement network for edge-aware
  salient object detection,'' in {\em IEEE International Conference on Computer
  Vision (ICCV)}, pp.~7264--7273, 2019.

\bibitem{ucnet_sal}
J.~Zhang, D.-P. Fan, Y.~Dai, S.~Anwar, F.~S. Saleh, T.~Zhang, and N.~Barnes,
  ``Uc-net: Uncertainty inspired rgb-d saliency detection via conditional
  variational autoencoders,'' in {\em IEEE Conference on Computer Vision and
  Pattern Recognition (CVPR)}, pp.~8582 -- 8591, 2020.

\bibitem{kendall2017uncertainties}
A.~Kendall and Y.~Gal, ``What uncertainties do we need in bayesian deep
  learning for computer vision?,'' {\em arXiv preprint arXiv:1703.04977}, 2017.

\bibitem{vae_bayes_kumar}
D.~P. {Kingma} and M.~{Welling}, ``Auto-encoding variational bayes,'' in {\em
  International Conference on Learning Representations (ICLR)}, 2014.

\bibitem{GAN_nips}
I.~Goodfellow, J.~Pouget-Abadie, M.~Mirza, B.~Xu, D.~Warde-Farley, S.~Ozair,
  A.~Courville, and Y.~Bengio, ``Generative adversarial nets,'' in {\em
  Conference on Neural Information Processing Systems (NIPS)}, pp.~2672--2680,
  2014.

\bibitem{structure_output}
K.~Sohn, H.~Lee, and X.~Yan, ``Learning structured output representation using
  deep conditional generative models,'' in {\em Conference on Neural
  Information Processing Systems (NIPS)}, pp.~3483--3491, 2015.

\bibitem{Lagging_Inference_Networks}
J.~He, D.~Spokoyny, G.~Neubig, and T.~Berg-Kirkpatrick, ``Lagging inference
  networks and posterior collapse in variational autoencoders,'' in {\em
  International Conference on Learning Representations (ICLR)}, 2019.

\bibitem{gan_raw}
I.~Goodfellow, J.~Pouget-Abadie, M.~Mirza, B.~Xu, D.~Warde-Farley, S.~Ozair,
  A.~Courville, and Y.~Bengio, ``Generative adversarial nets,'' in {\em
  Advances in Neural Information Processing Systems}, vol.~27, pp.~2672--2680,
  Curran Associates, Inc., 2014.

\bibitem{abp}
T.~Han, Y.~Lu, S.-C. Zhu, and Y.~N. Wu, ``Alternating back-propagation for
  generator network,'' in {\em AAAI Conference on Artificial Intelligence
  (AAAI)}, p.~1976–1984, 2017.

\bibitem{xie2019learning}
J.~Xie, R.~Gao, Z.~Zheng, S.-C. Zhu, and Y.~N. Wu, ``Learning dynamic generator
  model by alternating back-propagation through time,'' in {\em AAAI Conference
  on Artificial Intelligence (AAAI)}, pp.~5498--5507, 2019.

\bibitem{liu2008monte}
J.~S. Liu, {\em Monte Carlo strategies in scientific computing}.
\newblock Springer Science \& Business Media, 2008.

\bibitem{neal2011mcmc}
R.~M. Neal, ``{MCMC} using hamiltonian dynamics,'' {\em Handbook of markov
  chain monte carlo}, vol.~2, no.~11, p.~2, 2011.

\bibitem{WellingT11}
M.~Welling and Y.~W. Teh, ``Bayesian learning via stochastic gradient langevin
  dynamics,'' in {\em International Conference on Machine Learning (ICML)},
  pp.~681--688, 2011.

\bibitem{DubeyRWPSX16}
K.~A. Dubey, S.~J. Reddi, S.~A. Williamson, B.~P{\'{o}}czos, A.~J. Smola, and
  E.~P. Xing, ``Variance reduction in stochastic gradient langevin dynamics,''
  in {\em Conference on Neural Information Processing Systems (NIPS)},
  pp.~1154--1162, 2016.

\bibitem{ebm_prior}
B.~Pang, T.~Han, E.~Nijkamp, S.~Zhu, and Y.~N. Wu, ``Learning latent space
  energy-based prior model,'' in {\em Conference on Neural Information
  Processing Systems (NeurIPS)}, 2020.

\bibitem{PangW21}
B.~Pang and Y.~N. Wu, ``Latent space energy-based model of symbol-vector
  coupling for text generation and classification,'' in {\em International
  Conference on Machine Learning (ICML)}, pp.~8359--8370, 2021.

\bibitem{Liu_2021_ICCV_VST}
N.~Liu, N.~Zhang, K.~Wan, L.~Shao, and J.~Han, ``Visual saliency transformer,''
  in {\em IEEE International Conference on Computer Vision (ICCV)},
  pp.~4722--4732, 2021.

\bibitem{cpd_sal}
Z.~Wu, L.~Su, and Q.~Huang, ``Cascaded partial decoder for fast and accurate
  salient object detection,'' in {\em IEEE Conference on Computer Vision and
  Pattern Recognition (CVPR)}, pp.~3902--3911, 2019.

\bibitem{nldf_sal}
Z.~Luo, A.~Mishra, A.~Achkar, J.~Eichel, S.~Li, and P.-M. Jodoin, ``Non-local
  deep features for salient object detection,'' in {\em IEEE Conference on
  Computer Vision and Pattern Recognition (CVPR)}, 2017.

\bibitem{wang2020progressive}
B.~Wang, Q.~Chen, M.~Zhou, Z.~Zhang, X.~Jin, and K.~Gai, ``Progressive feature
  polishing network for salient object detection.,'' in {\em AAAI Conference on
  Artificial Intelligence (AAAI)}, pp.~12128--12135, 2020.

\bibitem{zhang2020learning_eccv}
J.~Zhang, J.~Xie, and N.~Barnes, ``Learning noise-aware encoder-decoder from
  noisy labels by alternating back-propagation for saliency detection,'' in
  {\em European Conference on Computer Vision (ECCV)}, pp.~349--366, 2020.

\bibitem{aixuan_cod_sod21}
A.~Li, J.~Zhang, Y.~Lyu, B.~Liu, T.~Zhang, and Y.~Dai, ``Uncertainty-aware
  joint salient object and camouflaged object detection,'' in {\em IEEE
  Conference on Computer Vision and Pattern Recognition (CVPR)}, 2021.

\bibitem{distributional_uncertainty_cvpr2023}
X.~Tian, J.~Zhang, M.~Xiang, and Y.~Dai, ``Modeling the distributional
  uncertainty for salient object detection models,'' in {\em IEEE Conference on
  Computer Vision and Pattern Recognition (CVPR)}, 2023.

\bibitem{qin2019basnet}
X.~Qin, Z.~Zhang, C.~Huang, C.~Gao, M.~Dehghan, and M.~Jagersand, ``Basnet:
  Boundary-aware salient object detection,'' in {\em IEEE Conference on
  Computer Vision and Pattern Recognition (CVPR)}, pp.~7479--7489, 2019.

\bibitem{Pang_2020_CVPR}
Y.~Pang, X.~Zhao, L.~Zhang, and H.~Lu, ``Multi-scale interactive network for
  salient object detection,'' in {\em IEEE Conference on Computer Vision and
  Pattern Recognition (CVPR)}, pp.~9413--9422, 2020.

\bibitem{qu2017rgbd}
L.~Qu, S.~He, J.~Zhang, J.~Tian, Y.~Tang, and Q.~Yang, ``{RGBD} salient object
  detection via deep fusion,'' {\em IEEE Transactions on Image Processing
  (TIP)}, vol.~26, no.~5, pp.~2274--2285, 2017.

\bibitem{ji2021cal}
W.~Ji, J.~Li, S.~Yu, M.~Zhang, Y.~Piao, S.~Yao, Q.~Bi, K.~Ma, Y.~Zheng, H.~Lu,
  and L.~Cheng, ``Calibrated rgb-d salient object detection,'' in {\em IEEE
  Conference on Computer Vision and Pattern Recognition (CVPR)},
  pp.~9471--9481, 2021.

\bibitem{zhao2019Contrast}
J.-X. Zhao, Y.~Cao, D.-P. Fan, M.-M. Cheng, X.-Y. Li, and L.~Zhang, ``Contrast
  prior and fluid pyramid integration for {RGBD} salient object detection,'' in
  {\em IEEE Conference on Computer Vision and Pattern Recognition (CVPR)},
  pp.~3927--3936, 2019.

\bibitem{ssf_rgbd}
M.~Zhang, W.~Ren, Y.~Piao, Z.~Rong, and H.~Lu, ``Select, supplement and focus
  for {RGB-D} saliency detection,'' in {\em IEEE Conference on Computer Vision
  and Pattern Recognition (CVPR)}, pp.~3472--3481, 2020.

\bibitem{ji2020accurate}
W.~Ji, J.~Li, M.~Zhang, Y.~Piao, and H.~Lu, ``Accurate {RGB-D} salient object
  detection via collaborative learning,'' in {\em European Conference on
  Computer Vision (ECCV)}, 2020.

\bibitem{piao2019depth}
Y.~Piao, W.~Ji, J.~Li, M.~Zhang, and H.~Lu, ``Depth-induced multi-scale
  recurrent attention network for saliency detection,'' in {\em IEEE
  International Conference on Computer Vision (ICCV)}, pp.~7254--7263, 2019.

\bibitem{zhang2020bilateral}
Z.~Zhang, Z.~Lin, J.~Xu, W.~Jin, S.~Lu, and D.~Fan, ``Bilateral attention
  network for {RGB-D} salient object detection,'' {\em IEEE Transactions on
  Image Processing (TIP)}, vol.~30, pp.~1949--1961, 2021.

\bibitem{Sun_2021_CVPR_DSA2F}
P.~Sun, W.~Zhang, H.~Wang, S.~Li, and X.~Li, ``Deep rgb-d saliency detection
  with depth-sensitive attention and automatic multi-modal fusion,'' in {\em
  IEEE Conference on Computer Vision and Pattern Recognition (CVPR)},
  pp.~1407--1417, 2021.

\bibitem{Zhang_2021_ICCV_RGBD}
J.~Zhang, D.-P. Fan, Y.~Dai, X.~Yu, Y.~Zhong, N.~Barnes, and L.~Shao, ``Rgb-d
  saliency detection via cascaded mutual information minimization,'' in {\em
  IEEE International Conference on Computer Vision (ICCV)}, pp.~4338--4347,
  2021.

\bibitem{transformer_nips}
A.~Vaswani, N.~Shazeer, N.~Parmar, J.~Uszkoreit, L.~Jones, A.~N. Gomez,
  L.~Kaiser, and I.~Polosukhin, ``Attention is all you need,'' in {\em
  Conference on Neural Information Processing Systems (NIPS)}, pp.~5998--6008,
  2017.

\bibitem{dosovitskiy_ViT_ICLR_2021}
A.~Dosovitskiy, L.~Beyer, A.~Kolesnikov, D.~Weissenborn, X.~Zhai,
  T.~Unterthiner, M.~Dehghani, M.~Minderer, G.~Heigold, S.~Gelly, J.~Uszkoreit,
  and N.~Houlsby, ``An image is worth 16x16 words: Transformers for image
  recognition at scale,'' in {\em International Conference on Learning
  Representations (ICLR)}, 2021.

\bibitem{liu2021swin}
Z.~Liu, Y.~Lin, Y.~Cao, H.~Hu, Y.~Wei, Z.~Zhang, S.~Lin, and B.~Guo, ``Swin
  transformer: Hierarchical vision transformer using shifted windows,'' in {\em
  IEEE International Conference on Computer Vision (ICCV)}, 2021.

\bibitem{carion_DETR_ECCV_2020}
N.~Carion, F.~Massa, G.~Synnaeve, N.~Usunier, A.~Kirillov, and S.~Zagoruyko,
  ``End-to-end object detection with transformers,'' in {\em European
  Conference on Computer Vision (ECCV)}, pp.~213--229, 2020.

\bibitem{wang_PVT_2021}
W.~Wang, E.~Xie, X.~Li, D.-P. Fan, K.~Song, D.~Liang, T.~Lu, P.~Luo, and
  L.~Shao, ``Pyramid vision transformer: A versatile backbone for dense
  prediction without convolutions,'' in {\em IEEE International Conference on
  Computer Vision (ICCV)}, pp.~568--578, 2021.

\bibitem{mvt_multi_view_transformer}
S.~Chen, T.~Yu, and P.~Li, ``Mvt: Multi-view vision transformer for 3d object
  recognition,'' in {\em British Machine Vision Conference (BMVC)}, 2021.

\bibitem{zheng_SETR_2020}
S.~Zheng, J.~Lu, H.~Zhao, X.~Zhu, Z.~Luo, Y.~Wang, Y.~Fu, J.~Feng, T.~Xiang,
  P.~H. Torr, and L.~Zhang, ``Rethinking semantic segmentation from a
  sequence-to-sequence perspective with transformers,'' {\em IEEE Conference on
  Computer Vision and Pattern Recognition (CVPR)}, pp.~6881--6890, 2021.

\bibitem{dpt_transformer}
R.~Ranftl, A.~Bochkovskiy, and V.~Koltun, ``Vision transformers for dense
  prediction,'' in {\em IEEE International Conference on Computer Vision
  (ICCV)}, pp.~12179--12188, 2021.

\bibitem{xu_TransCenterTracking_2021}
Y.~Xu, Y.~Ban, G.~Delorme, C.~Gan, D.~Rus, and X.~Alameda{-}Pineda,
  ``Transcenter: Transformers with dense queries for multiple-object
  tracking,'' {\em arXiv preprint arXiv:2103.15145}, 2021.

\bibitem{yan_SpatialTemporalTransformerTrackingv2_2021}
B.~Yan, H.~Peng, J.~Fu, D.~Wang, and H.~Lu, ``Learning spatio-temporal
  transformer for visual tracking,'' in {\em IEEE International Conference on
  Computer Vision (ICCV)}, pp.~10448--10457, 2021.

\bibitem{mao_TFPose_2021}
W.~Mao, Y.~Ge, C.~Shen, Z.~Tian, X.~Wang, and Z.~Wang, ``Tfpose: Direct human
  pose estimation with transformers,'' {\em arXiv preprint arXiv:2103.15320},
  2021.

\bibitem{stoffl_InstancePose_2021}
L.~Stoffl, M.~Vidal, and A.~Mathis, ``End-to-end trainable multi-instance pose
  estimation with transformers,'' {\em arXiv preprint arXiv:2103.12115}, 2021.

\bibitem{PHiSeg2019}
C.~F. Baumgartner, K.~C. Tezcan, K.~Chaitanya, A.~M. H{\"{o}}tker, U.~J.
  Muehlematter, K.~Schawkat, A.~S. Becker, O.~Donati, and E.~Konukoglu,
  ``Phiseg: Capturing uncertainty in medical image segmentation,'' in {\em The
  24th International Conference on Medical Image Computing and Computer
  Assisted Intervention (MICCAI)}, pp.~119--127, 2019.

\bibitem{probabilistic_unet}
S.~Kohl, B.~Romera-Paredes, C.~Meyer, J.~De~Fauw, J.~R. Ledsam, K.~Maier-Hein,
  S.~M.~A. Eslami, D.~Jimenez~Rezende, and O.~Ronneberger, ``A probabilistic
  u-net for segmentation of ambiguous images,'' in {\em Conference on Neural
  Information Processing Systems (NeurIPS)}, pp.~6965--6975, 2018.

\bibitem{SuperVAE_AAAI19}
B.~Li, Z.~Sun, and Y.~Guo, ``Supervae: Superpixelwise variational autoencoder
  for salient object detection,'' in {\em AAAI Conference on Artificial
  Intelligence (AAAI)}, pp.~8569--8576, 2019.

\bibitem{jing2020uncertainty}
J.~Zhang, D.-P. Fan, Y.~Dai, S.~Anwar, F.~S. Saleh, S.~Aliakbarian, and
  N.~Barnes, ``Uncertainty inspired rgb-d saliency detection,'' {\em IEEE
  Transactions on Pattern Analysis and Machine Intelligence (TPAMI)}, 2021.

\bibitem{groenendijk2020benefit}
R.~Groenendijk, S.~Karaoglu, T.~Gevers, and T.~Mensink, ``On the benefit of
  adversarial training for monocular depth estimation,'' {\em Computer Vision
  and Image Understanding}, vol.~190, p.~102848, 2020.

\bibitem{gan_maskerrcnn}
Q.~H. Le, K.~Youcef-Toumi, D.~Tsetserukou, and A.~Jahanian, ``Gan mask
  r-cnn:instance semantic segmentation benefits from generative adversarial
  networks,'' {\em arXiv preprint arXiv:2010.13757}, 2020.

\bibitem{gan_semi_seg}
N.~Souly, C.~Spampinato, and M.~Shah, ``Semi supervised semantic segmentation
  using generative adversarial network,'' in {\em IEEE International Conference
  on Computer Vision (ICCV)}, pp.~5689--5697, 2017.

\bibitem{hung2018adversarial}
W.-C. Hung, Y.-H. Tsai, Y.-T. Liou, Y.-Y. Lin, and M.-H. Yang, ``Adversarial
  learning for semi-supervised semantic segmentation,'' in {\em British Machine
  Vision Conference (BMVC)}, 2018.

\bibitem{zhang2021energy}
J.~Zhang, J.~Xie, Z.~Zheng, and N.~Barnes, ``Energy-based generative
  cooperative saliency prediction,'' {\em arXiv preprint arXiv:2106.13389},
  2021.

\bibitem{XieLGW18}
J.~Xie, Y.~Lu, R.~Gao, and Y.~N. Wu, ``Cooperative learning of energy-based
  model and latent variable model via {MCMC} teaching,'' in {\em AAAI
  Conference on Artificial Intelligence (AAAI)}, pp.~4292--4301, 2018.

\bibitem{xie2018cooperative}
J.~Xie, Y.~Lu, R.~Gao, S.-C. Zhu, and Y.~N. Wu, ``Cooperative training of
  descriptor and generator networks,'' {\em IEEE Transactions on Pattern
  Analysis and Machine Intelligence (TPAMI)}, vol.~42, no.~1, pp.~27--45, 2018.

\bibitem{xie2021cooperative}
J.~Xie, Z.~Zheng, X.~Fang, S.-C. Zhu, and Y.~N. Wu, ``Cooperative training of
  fast thinking initializer and slow thinking solver for conditional
  learning,'' {\em IEEE Transactions on Pattern Analysis and Machine
  Intelligence (TPAMI)}, 2021.

\bibitem{xie2016theory}
J.~Xie, Y.~Lu, S.-C. Zhu, and Y.~Wu, ``A theory of generative convnet,'' in
  {\em International Conference on Machine Learning (ICML)}, pp.~2635--2644,
  2016.

\bibitem{nijkamp2019learning}
E.~Nijkamp, M.~Hill, S.~Zhu, and Y.~N. Wu, ``Learning non-convergent
  non-persistent short-run {MCMC} toward energy-based model,'' in {\em
  Conference on Neural Information Processing Systems (NeurIPS)},
  pp.~5233--5243, 2019.

\bibitem{XieHZW15}
J.~Xie, W.~Hu, S.~Zhu, and Y.~N. Wu, ``Learning sparse {FRAME} models for
  natural image patterns,'' {\em International Journal of Computer Vision
  (IJCV)}, vol.~114, no.~2-3, pp.~91--112, 2015.

\bibitem{xie2016inducing}
J.~Xie, Y.~Lu, S.-C. Zhu, and Y.~N. Wu, ``Inducing wavelets into random fields
  via generative boosting,'' {\em Applied and Computational Harmonic Analysis},
  vol.~41, no.~1, pp.~4--25, 2016.

\bibitem{ZhengXL21}
Z.~Zheng, J.~Xie, and P.~Li, ``Patchwise generative convnet: Training
  energy-based models from a single natural image for internal learning,'' in
  {\em IEEE Conference on Computer Vision and Pattern Recognition (CVPR)},
  pp.~2961--2970, 2021.

\bibitem{ZhaoXL21}
Y.~Zhao, J.~Xie, and P.~Li, ``Learning energy-based generative models via
  coarse-to-fine expanding and sampling,'' in {\em International Conference on
  Learning Representations (ICLR)}, 2021.

\bibitem{GaoLZZW18}
R.~Gao, Y.~Lu, J.~Zhou, S.~Zhu, and Y.~N. Wu, ``Learning generative convnets
  via multi-grid modeling and sampling,'' in {\em IEEE Conference on Computer
  Vision and Pattern Recognition (CVPR)}, pp.~9155--9164, 2018.

\bibitem{DuM19}
Y.~Du and I.~Mordatch, ``Implicit generation and modeling with energy based
  models,'' in {\em Conference on Neural Information Processing Systems
  (NeurIPS)}, pp.~3603--3613, 2019.

\bibitem{GaoSPWK21}
R.~Gao, Y.~Song, B.~Poole, Y.~N. Wu, and D.~P. Kingma, ``Learning energy-based
  models by diffusion recovery likelihood,'' in {\em International Conference
  on Learning Representations (ICLR)}, 2021.

\bibitem{XieZL21}
J.~Xie, Z.~Zheng, and P.~Li, ``Learning energy-based model with variational
  auto-encoder as amortized sampler,'' in {\em AAAI Conference on Artificial
  Intelligence (AAAI)}, pp.~10441--10451, 2021.

\bibitem{XieZLL22}
J.~Xie, Y.~Zhu, J.~Li, and P.~Li, ``A tale of two flows: Cooperative learning
  of langevin flow and normalizing flow toward energy-based model,'' in {\em
  International Conference on Learning Representations (ICLR)}, 2022.

\bibitem{XieZW17}
J.~Xie, S.~Zhu, and Y.~N. Wu, ``Synthesizing dynamic patterns by
  spatial-temporal generative convnet,'' in {\em IEEE Conference on Computer
  Vision and Pattern Recognition (CVPR)}, pp.~1061--1069, 2017.

\bibitem{XieZW21}
J.~Xie, S.~Zhu, and Y.~N. Wu, ``Learning energy-based spatial-temporal
  generative convnets for dynamic patterns,'' {\em IEEE Transactions on Pattern
  Analysis and Machine Intelligence (TPAMI)}, vol.~43, no.~2, pp.~516--531,
  2021.

\bibitem{XieZGWZW18}
J.~Xie, Z.~Zheng, R.~Gao, W.~Wang, S.~Zhu, and Y.~N. Wu, ``Learning descriptor
  networks for 3d shape synthesis and analysis,'' in {\em IEEE Conference on
  Computer Vision and Pattern Recognition (CVPR)}, pp.~8629--8638, 2018.

\bibitem{xie2020generative}
J.~Xie, Z.~Zheng, R.~Gao, W.~Wang, S.-C. Zhu, and Y.~N. Wu, ``Generative
  voxelnet: Learning energy-based models for 3d shape synthesis and analysis,''
  {\em IEEE Transactions on Pattern Analysis and Machine Intelligence (TPAMI)},
  2020.

\bibitem{XieXZZW21}
J.~Xie, Y.~Xu, Z.~Zheng, S.~Zhu, and Y.~N. Wu, ``Generative pointnet: Deep
  energy-based learning on unordered point sets for 3d generation,
  reconstruction and classification,'' in {\em IEEE Conference on Computer
  Vision and Pattern Recognition (CVPR)}, pp.~14976--14985, 2021.

\bibitem{xu2022energy}
Y.~Xu, J.~Xie, T.~Zhao, C.~Baker, Y.~Zhao, and Y.~N. Wu, ``Energy-based
  continuous inverse optimal control,'' {\em IEEE transactions on neural
  networks and learning systems}, 2022.

\bibitem{aistats_lebm}
Y.~Zhu, J.~Xie, and P.~Li, ``Likelihood-based generative radiance field with
  latent space energy-based model for 3d-aware disentangled image
  representation,'' in {\em International Conference on Artificial Intelligence
  and Statistics (AISTATS)}, 2023.

\bibitem{PangZ0W21}
B.~Pang, T.~Zhao, X.~Xie, and Y.~N. Wu, ``Trajectory prediction with latent
  belief energy-based model,'' in {\em IEEE Conference on Computer Vision and
  Pattern Recognition (CVPR)}, pp.~11814--11824, 2021.

\bibitem{AnXL21}
D.~An, J.~Xie, and P.~Li, ``Learning deep latent variable models by short-run
  {MCMC} inference with optimal transport correction,'' in {\em {IEEE}
  Conference on Computer Vision and Pattern Recognition (CVPR)},
  pp.~15415--15424, 2021.

\bibitem{kingma2014adam}
D.~P. Kingma and J.~Ba, ``Adam: {A} method for stochastic optimization,'' in
  {\em International Conference on Learning Representations (ICLR)}, 2015.

\bibitem{nijkamp2020learning}
E.~Nijkamp, B.~Pang, T.~Han, L.~Zhou, S.-C. Zhu, and Y.~N. Wu, ``Learning
  multi-layer latent variable model via variational optimization of short run
  mcmc for approximate inference,'' in {\em European Conference on Computer
  Vision (ECCV)}, pp.~361--378, Springer, 2020.

\bibitem{robbins1951stochastic}
H.~Robbins and S.~Monro, ``A stochastic approximation method,'' {\em The annals
  of mathematical statistics}, pp.~400--407, 1951.

\bibitem{AroraRZ18}
S.~Arora, A.~Risteski, and Y.~Zhang, ``Do gans learn the distribution? some
  theory and empirics,'' in {\em International Conference on Learning
  Representations (ICLR)}, 2018.

\bibitem{PathakKDDE16}
D.~Pathak, P.~Kr{\"{a}}henb{\"{u}}hl, J.~Donahue, T.~Darrell, and A.~A. Efros,
  ``Context encoders: Feature learning by inpainting,'' in {\em IEEE Conference
  on Computer Vision and Pattern Recognition (CVPR)}, 2016.

\bibitem{IsolaZZE17}
P.~Isola, J.~Zhu, T.~Zhou, and A.~A. Efros, ``Image-to-image translation with
  conditional adversarial networks,'' in {\em {IEEE} Conference on Computer
  Vision and Pattern Recognition (CVPR)}, pp.~5967--5976, 2017.

\bibitem{denseaspp}
M.~Yang, K.~Yu, C.~Zhang, Z.~Li, and K.~Yang, ``Denseaspp for semantic
  segmentation in street scenes,'' in {\em IEEE Conference on Computer Vision
  and Pattern Recognition (CVPR)}, pp.~3684--3692, 2018.

\bibitem{rca_eccv}
Y.~Zhang, K.~Li, K.~Li, L.~Wang, B.~Zhong, and Y.~Fu, ``Image super-resolution
  using very deep residual channel attention networks,'' in {\em European
  Conference on Computer Vision (ECCV)}, pp.~286--301, 2018.

\bibitem{hendrycks2016gaussian}
D.~Hendrycks and K.~Gimpel, ``Gaussian error linear units (gelus),'' {\em arXiv
  preprint arXiv:1606.08415}, 2016.

\bibitem{imagesaliency}
L.~Wang, H.~Lu, Y.~Wang, M.~Feng, D.~Wang, B.~Yin, and X.~Ruan, ``Learning to
  detect salient objects with image-level supervision,'' in {\em IEEE
  Conference on Computer Vision and Pattern Recognition (CVPR)}, pp.~136--145,
  2017.

\bibitem{yan2013hierarchical}
Q.~Yan, L.~Xu, J.~Shi, and J.~Jia, ``Hierarchical saliency detection,'' in {\em
  IEEE Conference on Computer Vision and Pattern Recognition (CVPR)},
  pp.~1155--1162, 2013.

\bibitem{Manifold-Ranking:CVPR-2013}
C.~Yang, L.~Zhang, H.~Lu, X.~Ruan, and M.-H. Yang, ``Saliency detection via
  graph-based manifold ranking,'' in {\em IEEE Conference on Computer Vision
  and Pattern Recognition (CVPR)}, pp.~3166--3173, 2013.

\bibitem{li2015visual}
G.~Li and Y.~Yu, ``Visual saliency based on multiscale deep features,'' in {\em
  IEEE Conference on Computer Vision and Pattern Recognition (CVPR)},
  pp.~5455--5463, 2015.

\bibitem{pascal_s_dataset}
Y.~Li, X.~Hou, C.~Koch, J.~M. Rehg, and A.~L. Yuille, ``The secrets of salient
  object segmentation,'' in {\em IEEE Conference on Computer Vision and Pattern
  Recognition (CVPR)}, pp.~280--287, 2014.

\bibitem{sod_dataset}
V.~Movahedi and J.~H. Elder, ``Design and perceptual validation of performance
  measures for salient object segmentation,'' in {\em IEEE Conference on
  Computer Vision and Pattern Recognition (CVPR) Workshop}, pp.~49--56, 2010.

\bibitem{Liu19PoolNet}
J.-J. Liu, Q.~Hou, M.-M. Cheng, J.~Feng, and J.~Jiang, ``A simple pooling-based
  design for real-time salient object detection,'' in {\em IEEE Conference on
  Computer Vision and Pattern Recognition (CVPR)}, 2019.

\bibitem{zhao2019EGNet}
J.-X. Zhao, J.-J. Liu, D.-P. Fan, Y.~Cao, J.~Yang, and M.-M. Cheng,
  ``Egnet:edge guidance network for salient object detection,'' in {\em IEEE
  International Conference on Computer Vision (ICCV)}, pp.~8778--8787, 2019.

\bibitem{zhou2020interactive}
H.~Zhou, X.~Xie, J.-H. Lai, Z.~Chen, and L.~Yang, ``Interactive two-stream
  decoder for accurate and fast saliency detection,'' in {\em IEEE Conference
  on Computer Vision and Pattern Recognition (CVPR)}, pp.~9141--9150, 2020.

\bibitem{xu2021locate}
B.~Xu, H.~Liang, R.~Liang, and P.~Chen, ``Locate globally, segment locally: A
  progressive architecture with knowledge review network for salient object
  detection,'' in {\em AAAI Conference on Artificial Intelligence (AAAI)},
  pp.~3004--3012, 2021.

\bibitem{Miao_2021_ACM_MM}
M.~{Zhang}, T.~{Liu}, Y.~{Piao}, S.~{Yao}, and H.~{Lu}, ``Auto-msfnet: Search
  multi-scale fusion network for salient object detection,'' in {\em ACM
  Multimedia Conference 2021}, 2021.

\bibitem{NJU2000}
R.~Ju, Y.~Liu, T.~Ren, L.~Ge, and G.~Wu, ``Depth-aware salient object detection
  using anisotropic center-surround difference,'' {\em Signal Processing: Image
  Communication}, vol.~38, pp.~115 -- 126, 2015.

\bibitem{peng2014rgbd}
H.~Peng, B.~Li, W.~Xiong, W.~Hu, and R.~Ji, ``{RGBD} salient object detection:
  A benchmark and algorithms,'' in {\em European Conference on Computer Vision
  (ECCV)}, pp.~92--109, 2014.

\bibitem{li2014saliency}
N.~Li, J.~Ye, Y.~Ji, H.~Ling, and J.~Yu, ``Saliency detection on light field,''
  in {\em IEEE Conference on Computer Vision and Pattern Recognition (CVPR)},
  pp.~2806--2813, 2014.

\bibitem{cheng2014depth}
Y.~Cheng, H.~Fu, X.~Wei, J.~Xiao, and X.~Cao, ``Depth enhanced saliency
  detection method,'' in {\em Proceedings of International Conference on
  Internet Multimedia Computing and Service}, pp.~23--27, 2014.

\bibitem{niu2012leveraging}
Y.~Niu, Y.~Geng, X.~Li, and F.~Liu, ``Leveraging stereopsis for saliency
  analysis,'' in {\em IEEE Conference on Computer Vision and Pattern
  Recognition (CVPR)}, pp.~454--461, 2012.

\bibitem{sip_dataset}
D.-P. Fan, Z.~Lin, Z.~Zhang, M.~Zhu, and M.-M. Cheng, ``{Rethinking RGB-D
  Salient Object Detection: Models, Datasets, and Large-Scale Benchmarks},''
  {\em IEEE Transactions on Neural Networks and Learning Systems (TNNLS)},
  2020.

\bibitem{fan2018enhanced}
D.-P. Fan, C.~Gong, Y.~Cao, B.~Ren, M.-M. Cheng, and A.~Borji,
  ``Enhanced-alignment measure for binary foreground map evaluation,'' in {\em
  International Joint Conference on Artificial Intelligence (IJCAI)},
  pp.~698--704, 2018.

\bibitem{fan2017structure}
D.-P. Fan, M.-M. Cheng, Y.~Liu, T.~Li, and A.~Borji, ``Structure-measure: A new
  way to evaluate foreground maps,'' in {\em IEEE International Conference on
  Computer Vision (ICCV)}, pp.~4548--4557, 2017.

\bibitem{imagenet_1k}
J.~Deng, W.~Dong, R.~Socher, L.-J. Li, K.~Li, and L.~Fei-Fei, ``Imagenet: A
  large-scale hierarchical image database,'' in {\em IEEE Conference on
  Computer Vision and Pattern Recognition (CVPR)}, pp.~248--255, 2009.

\end{thebibliography}

\end{document}